\documentclass[preprint,10pt,twocolumn,3p]{elsarticle}

\usepackage{amssymb}
\usepackage{times}
\usepackage{color}
\usepackage{subfigure}
\usepackage{epsfig}
\usepackage{graphicx}
\usepackage{lipsum}
\usepackage{amsmath}
\usepackage{multirow}
\usepackage{algorithm}
\usepackage{algpseudocode}
\usepackage{url}
\usepackage[pagebackref=false,breaklinks=true,letterpaper=true,colorlinks,urlcolor=blue,citecolor=blue,linkcolor=blue,bookmarks=false]{hyperref}

\journal{Computer Vision and Image Understanding}

\newcommand{\re}[1]{\textcolor{black}{#1}}
\begin{document}

\begin{frontmatter}



\title{Stylizing Face Images via Multiple Exemplars}


\author{Yibing~Song$^1$,
        Linchao~Bao$^2$,
        Shengfeng~He$^3$\\
        Qingxiong~Yang$^4$
        and~Ming-Hsuan~Yang$^5$
        }

\address{$^1$Department of Computer Science, City University of Hong Kong\\
$^2$Tencent AI Lab\\
$^3$South China University of Technology\\
$^4$School of Information Science and Technology, University of Science and Technology of China\\
$^5$Electrical Engineering and Computer Science, University of California at Merced}

\begin{abstract}
\re{We address the problem of transferring the style of a headshot photo to face images.
Existing methods using a single exemplar lead to inaccurate results when the exemplar does not contain sufficient stylized facial components for a given photo.
In this work, we propose an algorithm to stylize face images using multiple exemplars containing different subjects in the same style.
Patch correspondences between an input photo and multiple exemplars are established using a Markov Random Field (MRF), which enables accurate local energy transfer via
Laplacian stacks.
As image patches from multiple exemplars are used, the boundaries of facial components on the target image are inevitably inconsistent.
The artifacts are removed by a post-processing step using an edge-preserving filter.
Experimental results show that the proposed algorithm consistently produces visually pleasing results.}

\end{abstract}

\begin{keyword}
Style transfer \sep Image Processing



\end{keyword}

\end{frontmatter}


\section{Introduction}
\re{Transferring photo styles of professional headshot portraits to ordinary ones is of great importance in photo editing.}
\re{Traditionally, it requires professional photographers to perform painstaking
post-editing using specially designed photo editing systems.}
\re{Recently, automatic methods are proposed to ease this problem \cite{PhotoShop,Sunkavalli-siggraph10-Harmonization,Shih-siggraph14-StyleTransfer}.}
\re{These methods transfer the styles of photos produced by professional photographers to ordinary photos using exemplar-based learning algorithms.}

\begin{figure*}[t]
\begin{center}
\begin{tabular}{ccccc}
\includegraphics[width=.2\linewidth]{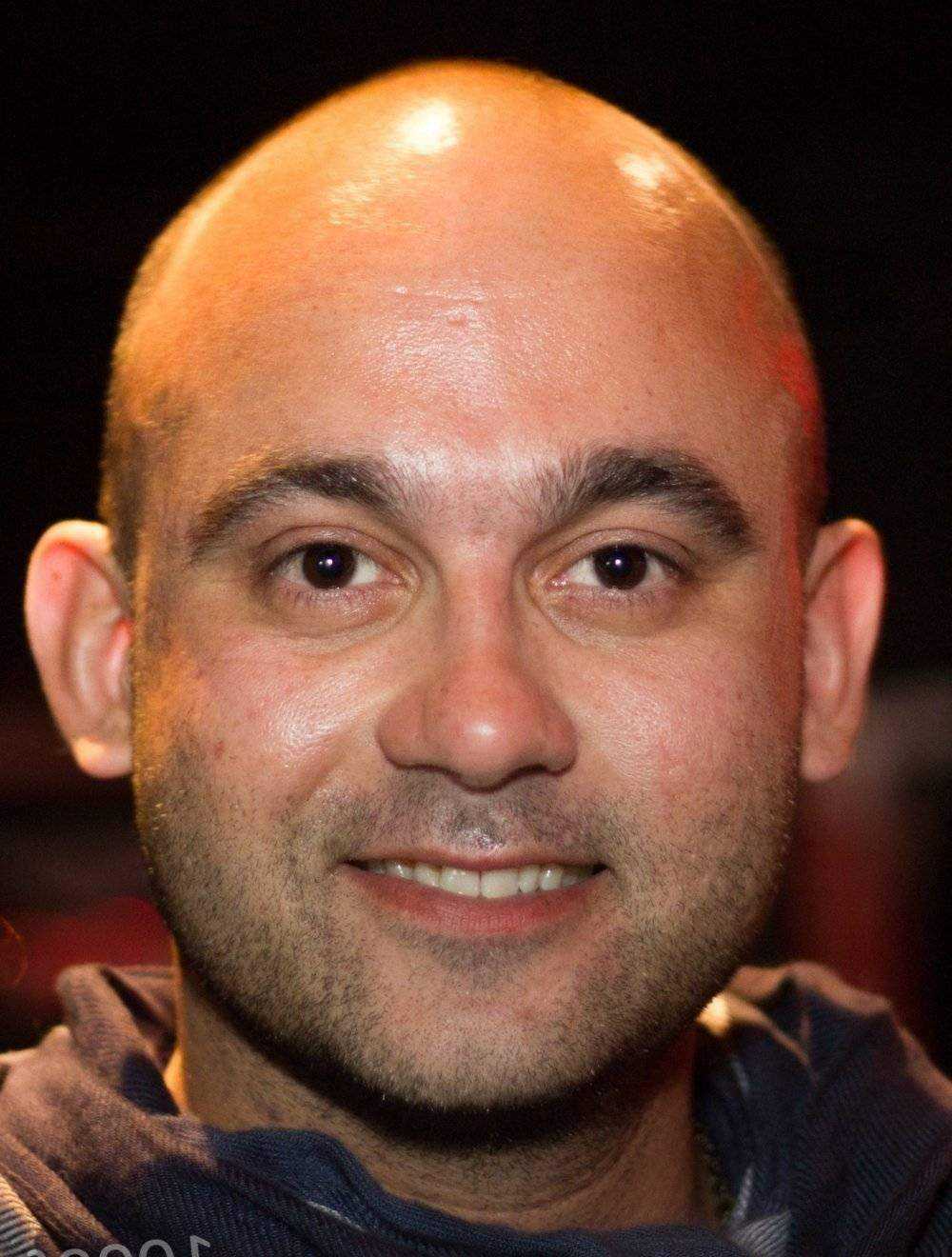}&
\includegraphics[width=.2\linewidth]{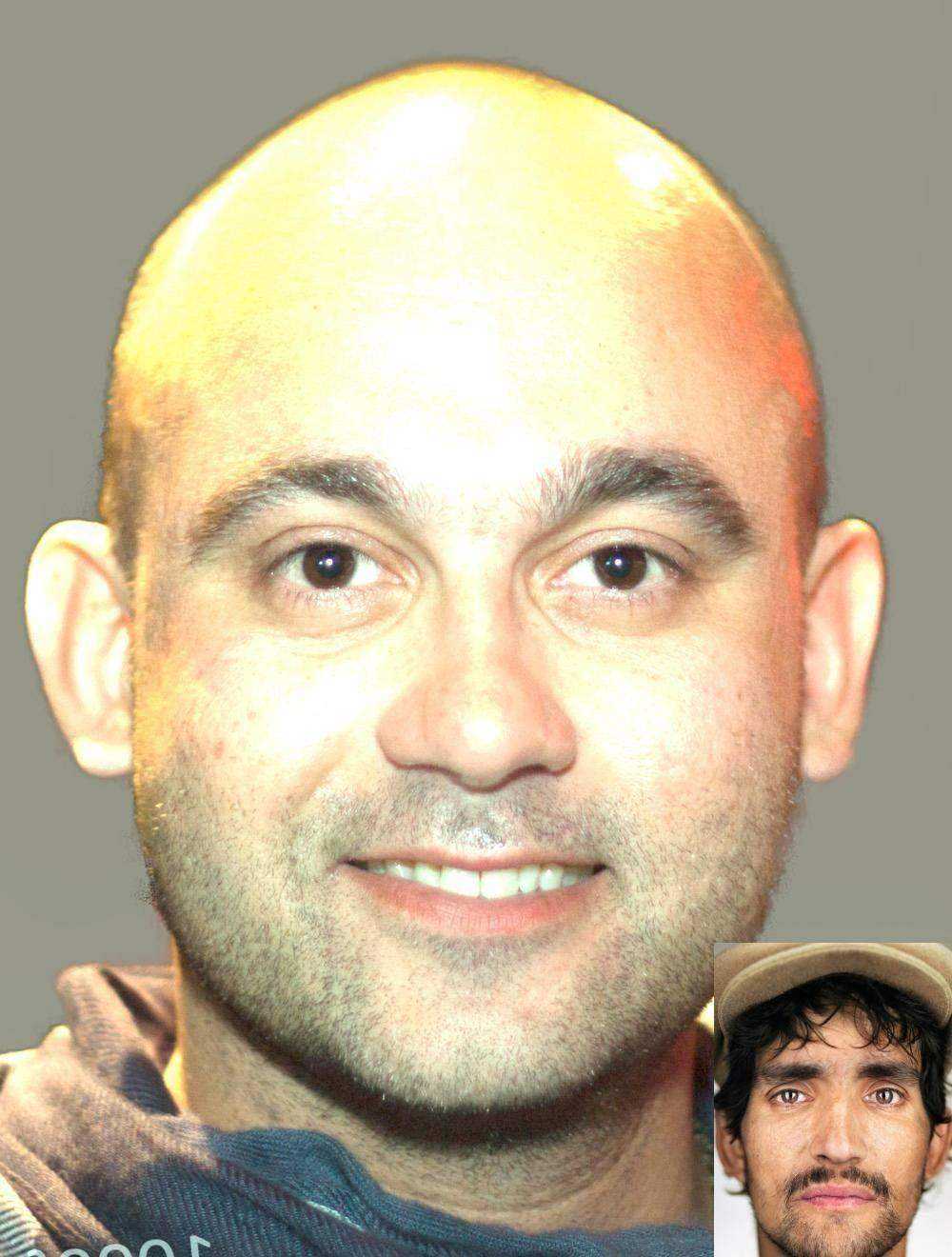}&
\includegraphics[width=.2\linewidth]{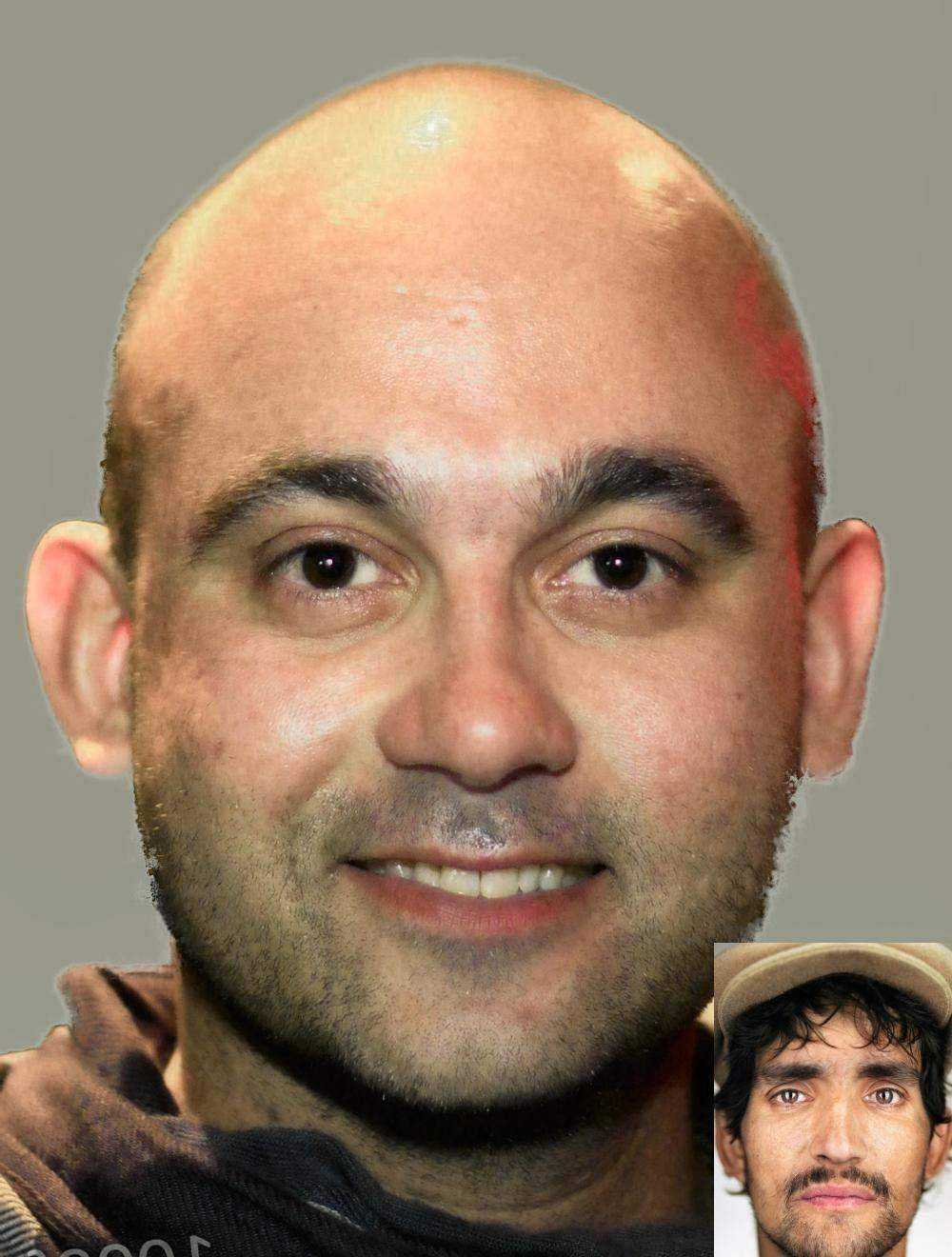}&
\includegraphics[width=.2\linewidth]{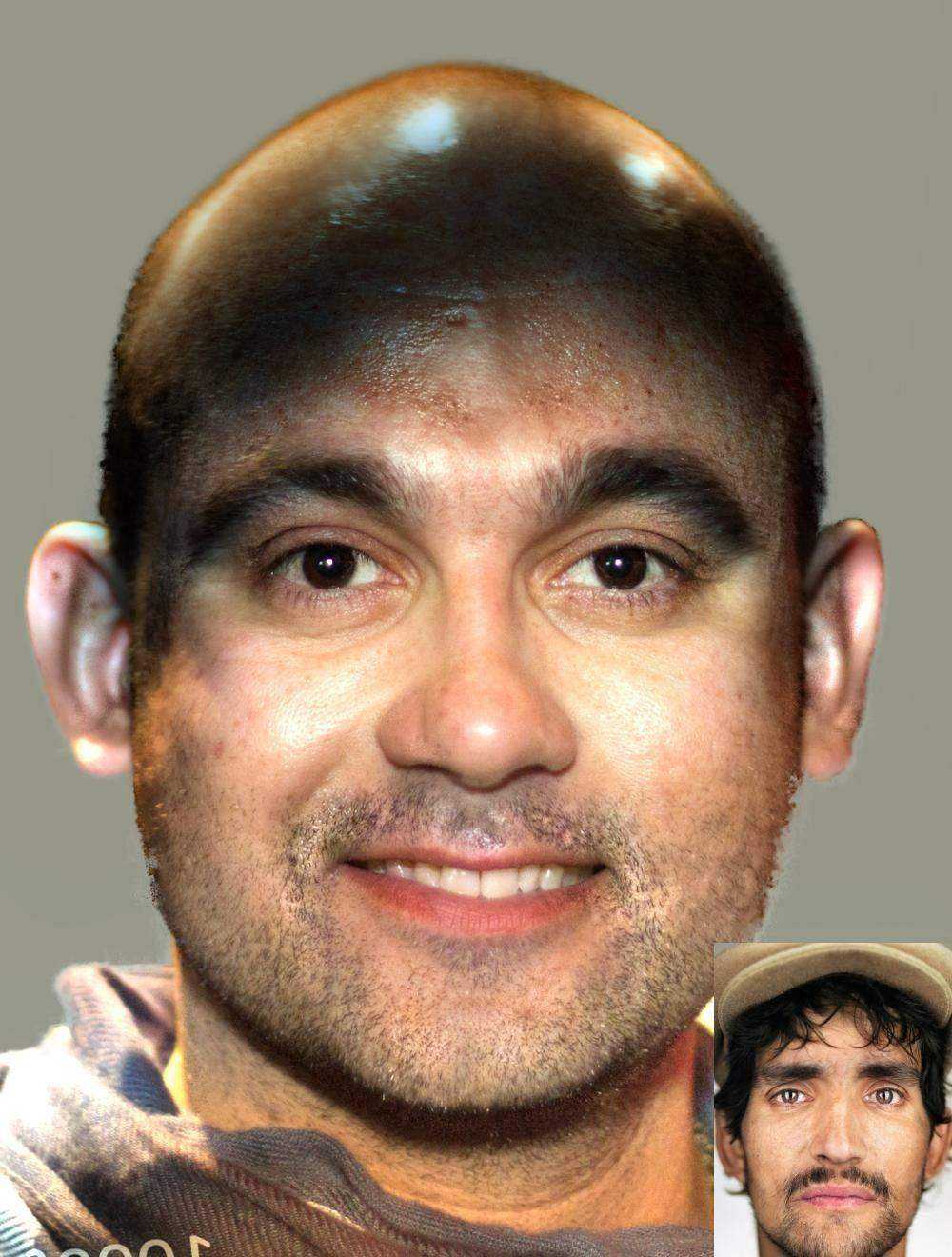}&
\includegraphics[width=.2\linewidth]{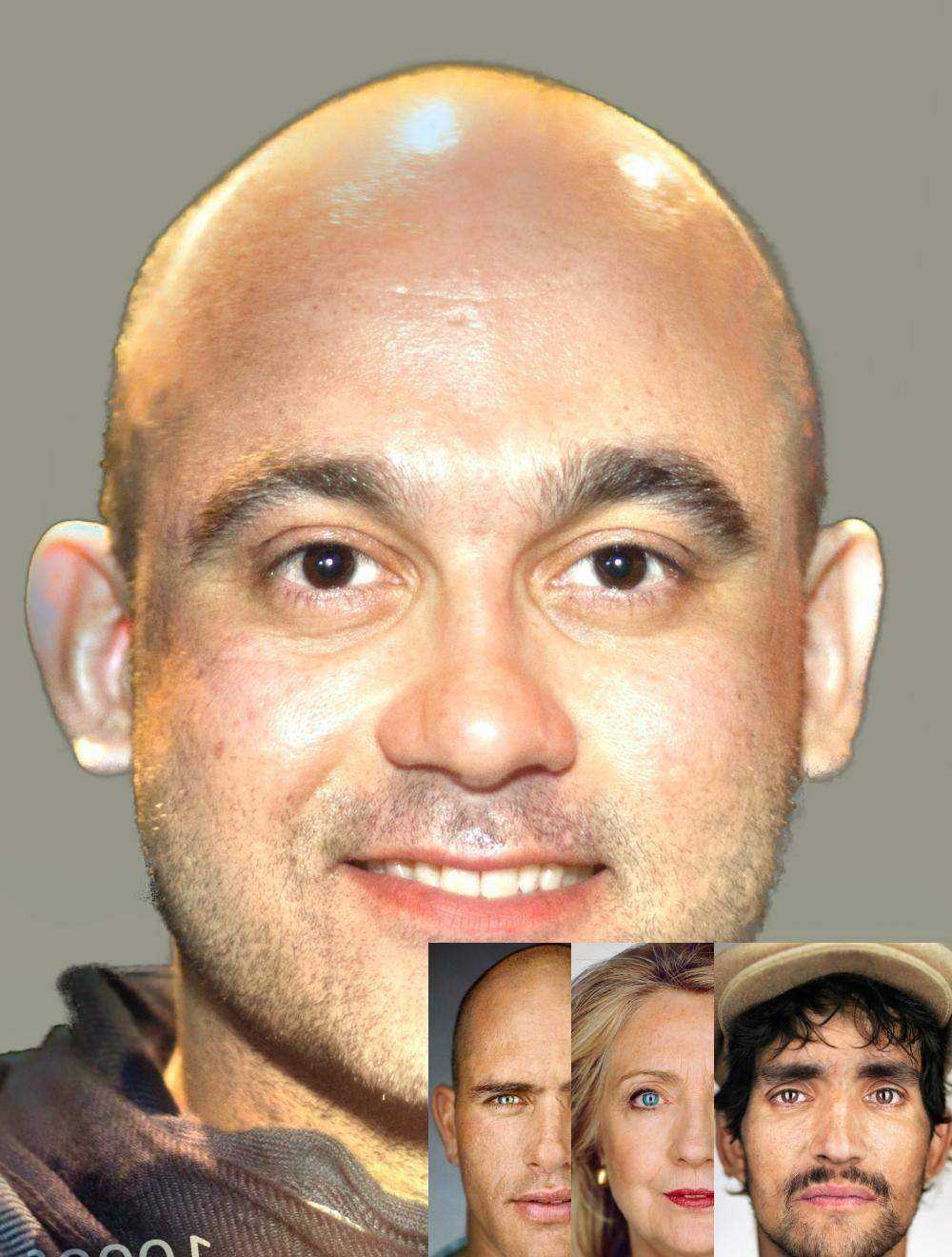}\\
(a) Input photo& (b) PhotoShop \cite{PhotoShop} &(c) Holistic \cite{Sunkavalli-siggraph10-Harmonization}&(d) Local \cite{Shih-siggraph14-StyleTransfer}&(e) Proposed\\
\end{tabular}
\end{center}

\caption{Face style transfer from single exemplar is widely adopted in
commercial products (e.g., Adobe PhotoShop match color function) and
recent work.
Given a collection of exemplars of the same style, these methods
select only one exemplar manually or automatically.
Existing methods are less effective when the selected exemplar differs
significantly from the input photo in terms of facial components.
\re{Such differences bring in
unnatural (e.g., (b)) or less stylistic (e.g., (c)) effects on the results from the holistic methods}.
In contrast, local method can effectively transfer
similar local details (e.g., nose and mouth in (d)) but \re{limits its performance
on the dissimilar regions (e.g., forehead)}.
Instead of selecting one single exemplar, the proposed algorithm
finds most similar facial components in the whole collection to address this
issue.
}

\label{fig:intro1}
\end{figure*}

\re{Although significant advancements have been made in recent years,
existing exemplar-based methods involve only a single exemplar for holistic style transfer.}
\re{They produce erroneous results if the exemplar is not able to provide sufficient stylized facial components for the given photo.}
\re{A straightforward solution is to select the best exemplar among a collection in the same style \cite{Shih-siggraph14-StyleTransfer}.}
\re{However, as the subject in the input photo is different from those in
the exemplar set, it is difficult to find a single exemplar
where all the facial components are similar to those in the input photo.}
\re{The mismatches between the input photo and selected
exemplar lead to incompatibility issues, which largely degrade the stylization quality.}
Figure \ref{fig:intro1} shows different methods using a single exemplar as reference.
Since the hair structures of the subjects in the input image and the
selected exemplar are different,
the methods based on holistic appearance are less effective to transfer
the skin tone and the ambient light to the stylized output.
Figure \ref{fig:intro1}(b) and (c) show that the
stylized images generated by the holistic methods
\cite{PhotoShop,Sunkavalli-siggraph10-Harmonization} are either
unnatural or \re{less stylistic}.
In contrast, the local method \cite{Shih-siggraph14-StyleTransfer}
can effectively stylize the input photo around similar facial components
(e.g., the nose and mouth shown in Figure \ref{fig:intro1}(d)).
However, some undesired effects are likely to be produced in the regions
where the components are different (e.g., forehead).
\re{To alleviate the problems of finding proper components for stylization, we select local regions from multiple exemplars instead of relying on a single one.}
\re{As such, we can consistently find correct and similar components from
all the exemplars even though they belong to different subjects}.

\renewcommand{\tabcolsep}{.1pt}

In this paper we propose a face stylization algorithm using
multiple exemplars.
\re{Instead of limiting to a single exemplar for each input photo,
we search the whole collection of exemplars with the same style to find the most similar component represented in each local patch of the input photo.}
\re{Given a photo, we first align all the exemplars using the local affine transformation and SIFT flow methods \cite{liu-pami2011-SiftFlow}}.
\re{Then we locally establish the patch correspondences between the input photo and multiple exemplars through a Markov random field.}
\re{Next, we construct a Laplacian pyramid for every image and remap the local contrast at multiple scales}.
\re{Finally, we remove the artifacts caused by inconsistent remapping from different exemplars using
a  edge-preserving filter}.
\re{As similar components can be consistently selected from the exemplar collection,
the proposed algorithm can effectively perform style transfer to an input photo.}
Qualitative and quantitative experimental results on
a benchmark dataset demonstrate the
effectiveness of the proposed algorithm with different artistic
styles.

The contributions of this work are summarized as follows:
\begin{itemize}
  \item \re{We propose a style transfer algorithm in which a Markov random field is used to incorporate patches from multiple exemplars.
 The proposed method enables the use of all stylization information from different exemplars.}
  \item \re{We propose an artifact removal method based on an edge-preserving filter.
 It removes the artifacts introduced by inconsistent boundaries of local patches stylized from different exemplars.}
  \item In addition to visual comparison conducted by existing methods, we perform quantitative evaluations using both objective and subjective metrics to demonstrate the effectiveness of the proposed method.
\end{itemize}

\section{Related Work}
Image style transfer methods can be broadly categorized into
holistic and local approaches.

{\flushleft \textbf{Holistic Approaches}:} These methods typically learn a mapping
function using one exemplar to adjust
the tone and lighting of the input photo.
In \cite{Pitie-iccv05-ColorTransfer}, a transformation function is
estimated over the entire image to map one distribution into another for color transfer.
A multiple layer style transfer method is proposed in \cite{bae-sig06-tone}
where an input image is decomposed into base and detail layers \re{where the style is
transferred independently}.
Further improvement is made in \cite{Sunkavalli-siggraph10-Harmonization}
where a multi-scale approach is presented to reduce artifacts
through an image pyramid.
In \cite{pitie-cviu07-grading}, a color grading approach is developed by using
color distribution transfer.
A graph regularization for color processing is proposed in \cite{lezoray-cviu07-graph}.
To reduce time complexity, an efficient method is proposed in
\cite{Hacohen-siggraph11-ImageEnhancement} based on the generalized
patchmatch algorithm \cite{Barnes-eccv10-PatchMatch}.
It uses a holistic non-linear parametric color model
to address dense correspondence problem.
We note these algorithms are effective in transferring image styles holistically
at the expense of capturing fine details, which are well transferred
using the proposed method.

{\flushleft \textbf{Local Approaches}:} These methods transfer the color and tone based on the distributions
on the exemplars.
In \cite{Tai-cvpr05-EM},
a local method is proposed for regional color
transfer between two natural images by probabilistic segmentation, and a scheme based on
expectation maximization is proposed to impose spatial and color smoothness.
An exemplar-based style transfer method is proposed
in \cite{Shih-siggraphAsia13-DayNight} where local affine
color transformation model is developed to render
natural images during different time of the day.
In addition to color or tone transfer,
numerous face photo decomposition methods based on edge-preserving filters
\cite{Farbman-siggraph08-WLS,yang-ijcv14-bf,kaiming-pami2013-GuidedFilter}
are developed for makeup transfer \cite{Guo-cvpr09-FaceMakeup}
and relighting \cite{Xiaowu-cvpr11-FaceRelighting}.
From an identity-specific collection of face images, an algorithm is developed to
enhance low-quality photos based on high-quality ones by exploiting holistic and
face-specific regions (e.g., deblurring, light transfer, and super resolution)
\cite{Joshi-TOG10-PhotoEnhance}.
The training and input photos used in \cite{Joshi-TOG10-PhotoEnhance} are from
the same subject, and their goal is for image enhancement.

{\flushleft \textbf{Face Style Transfer Approach}:}
A local method that transfers the face style of an exemplar to
the input face image is proposed in \cite{Shih-siggraph14-StyleTransfer}.
It first generates dense correspondence
between an input photo and one selected exemplar.
Then it decomposes each image into a Laplacian stack before transferring
the local energy in each image frequency subband within each layer.
Finally all the stacks are aggregated to generate the output image.
Since the style represented by the local energy is precisely transferred in multiple
layers, it has the advantage to handle detailed facial components.
%
\re{Compared to the holistic methods, local approaches can better capture
the region details and thus facilitate face stylization}.
However, if the components appeared in the exemplars and the input photos are significantly different,
the resulting images are likely to contain undesired effects.
%
In this work, we use multiple exemplars to solve this problem.

\section{Algorithm}

\begin{figure*}[t]
\begin{center}
\begin{tabular}{c}
\includegraphics[width=.9\linewidth]{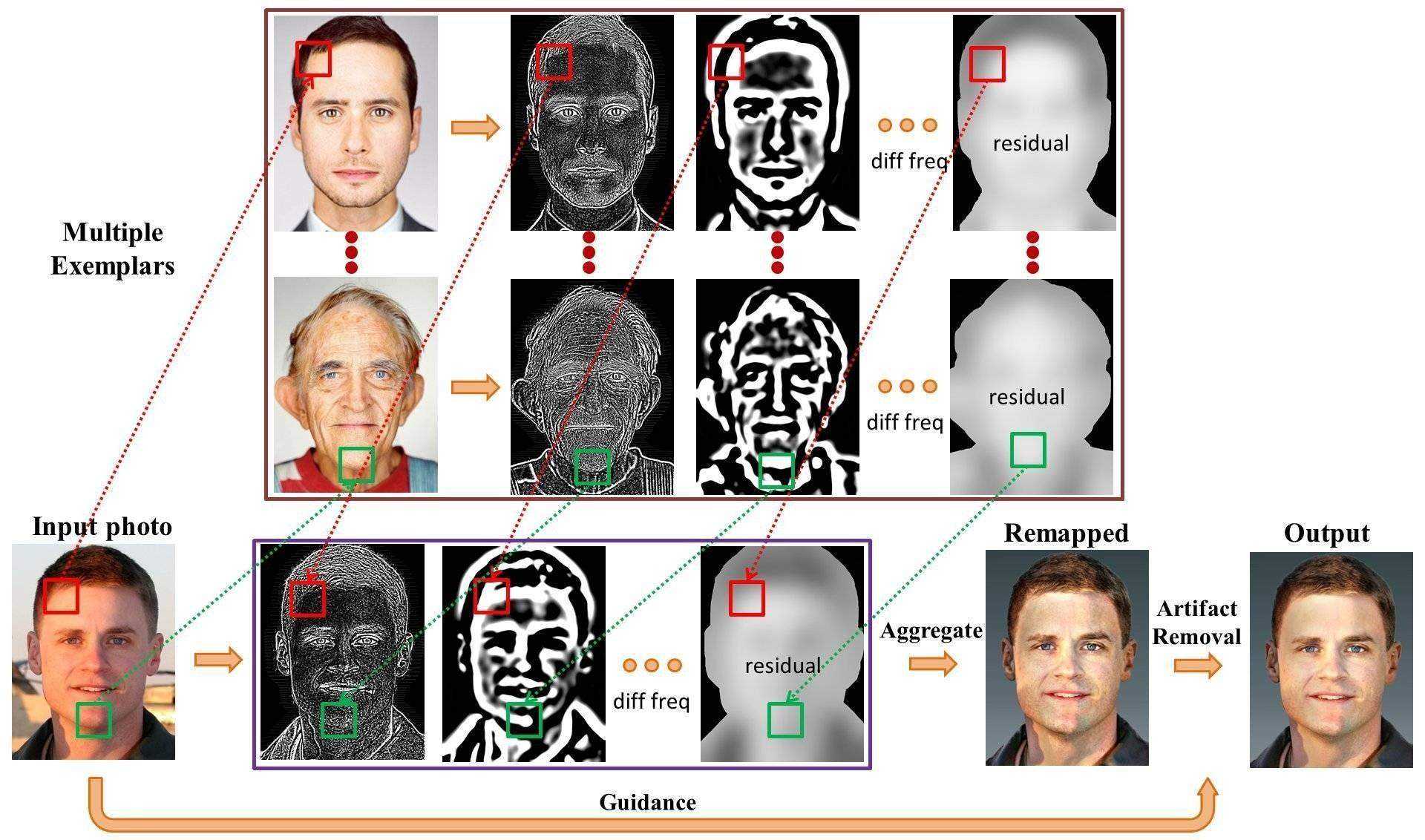}\\
\end{tabular}
\end{center}
\caption{\re{Face stylization via multiple exemplars}.
Using several exemplars from one collection we can consistently identify
similar facial components for each input photo.
Through local remapping in the Laplacian stacks we can effectively transfer
the local contrast.
However, style transfer will be inconsistent around the boundaries due to
the involvement of multiple exemplars and artifacts may occur.
\re{These artifacts are removed using the proposed edge-preserving filtering method
with the guidance of the input image}.}
\label{fig:Intuition}
\end{figure*}

The motivation of this work is illustrated with an example in Figure
\ref{fig:Intuition}.
Both the input image and exemplars are in the same resolution
and divided into overlapping patches.
Given a collection of exemplars in one
style, we aim to transfer the local details and contrast to an input photo while
maintaining its textures and structures.
We describe the details of the proposed algorithm in the following sections.

\subsection{Face Alignment and Local Identification}\label{sec:lm}

We align each exemplar to the input photo in the same way as
illustrated in \cite{Shih-siggraph14-StyleTransfer}.
First we obtain facial landmarks of each image using the fast
landmark detection method \cite{Vahid-cvpr14-Landmark}.
Through landmark correspondence, we apply a local affine transformation
to generate a dense correspondence field which
warps each exemplar into the input photo.
\re{We warp each exemplar accordingly and further align each warped exemplar
using the SIFT flow method \cite{liu-pami2011-SiftFlow}}.
\re{It refines the dense correspondence field locally to achieve pixel wise precision}.
After alignment we uniformly divide both exemplar
and the input image into overlapping patches.
The patch size and the center pixel locations are the same for each
input and exemplar patches.

\re{We construct a MRF model to incorporate all the exemplars for local
patch selection}.
\re{The MRF formulation considers both patch similarity and local smoothness constraints}.
We denote $N$ as the number of patches extracted from one image, and
$T_p$ as one patch centered at pixel $p$ in the input photo.
In addition, we denote $E_p$ and $E_q$
as the selected exemplar patches centered on $p$ and its neighboring pixel $q$.
The joint probability of patches from an input photo and selected exemplars can
be written as:
\begin{equation}
\emph{P}(T_1,\cdots,T_N,E_1,\cdots,E_N)=\prod_p\Phi(T_p,E_p)\prod_{p,q}\Psi(E_p,E_q)
\label{eq:joint_p}
\end{equation}
where $E_p$ has a discrete representation taking values
from the number of exemplars.
We denote $E^k_p$ as the patch centered
on $p$ in the $k$-th exemplar.
We compute the similarity $\Phi_p(T_p,E^k_p)$ between
$T_p$ and $E^k_p$ by
\begin{equation}
\Phi_p(T_p,E^k_p)=\exp(-{\frac{D_p^2}{2\sigma_d^2}})
\label{eq:data}
\end{equation}
where $D_p$ is the distance between an input patch $T_p$ and the corresponding
exemplar patch $E^k_p$.
We define patch distance in terms of
normalized cross correlation and absolute difference by
\begin{equation}
D_p=\alpha\cdot(1-D_{ncc})+(1-\alpha)\cdot D_{abs}
\label{eq:cost}
\end{equation}
where $\alpha$ is a weighting factor, $D_{abs}$ is the tone similarity
and $D_{ncc}$ is the structural similarity.
\re{We set $\alpha$ to be 0.8 in all the experiments since we emphasize on the structure
similarity during local patch selection}.
\re{Meanwhile, we also set a small weight (i.e, $1-\alpha$) on the tone similarity when the structures among
exemplar patches are similar.}
The value of each image pixel is normalized to $[0,1]$.

The compatibility function $\Psi_p(E^k_p,E^j_q)$ measures the local
smoothness between two exemplar patches centered at pixel $p$ and its
neighboring pixel $q$, respectively.
We define it as
\begin{eqnarray}
\hspace{-0.5cm}
\Psi_p(E^k_p,E^j_q)&=&\exp(-{\frac{C}{2\sigma_c^2}})\label{eq:smooth1}\\
C&=&\frac{1}{n}\sum_{o\in\Omega}\|E^k_p(o)-E^j_q(o)\|^2\label{eq:smooth2}
\end{eqnarray}
where $n$ is the number of pixels in $\Omega$ which is the overlapping
region between $E^k_p$ and $E^j_q$.
We use the minimum mean-squared error (MMSE) to estimate
the optimal candidate patch with
\begin{eqnarray}
\hspace{-0.5cm}
\hat{E_p}&=&\sum_{E_p^k}E_p^k\cdot\Phi(T_p,E^k_p)\prod_jM_k^j(E_p^k)\\
M_k^j&=&\sum_{E_q^j}\Psi(E_p^k,E_q^j)\Phi(T_q,E^j_q)\prod_{l\neq j}\hat{M_j^l}(E_q^j)\label{eq:message}
\end{eqnarray}
where $\hat{M_k^j}(E_p^k)$ is the message $M_k^j(E_p^k)$ computed from
the previous iteration.
\re{The probabilities of the patch similarity and local smoothness are updated in each iteration of the
belief propagation \cite{Freeman-ijcv00-bp,Yedidia-2003-bp} with the MRF model}.
\re{After the belief propagation process, we select the optimal patches locally which contain the maximum probabilities.}

\renewcommand{\tabcolsep}{0.5pt}
\begin{figure}[tp]
\begin{center}
\begin{tabular}{c}
\includegraphics[width=.96\linewidth]{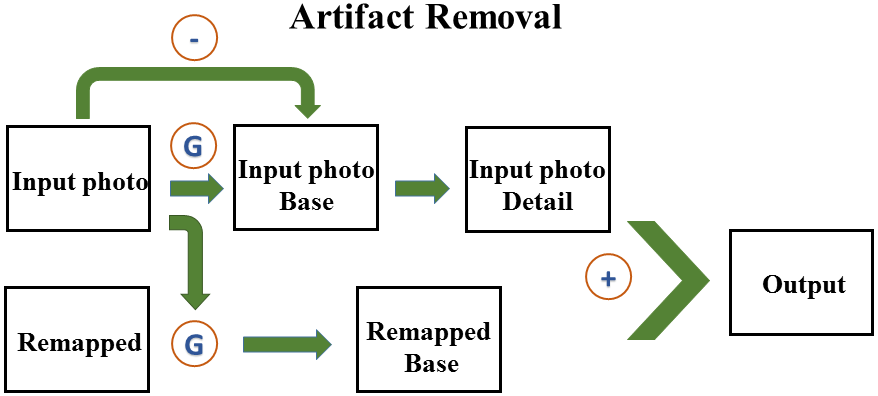}\\
\end{tabular}
\begin{tabular}{cc}
\includegraphics[width=0.48\linewidth]{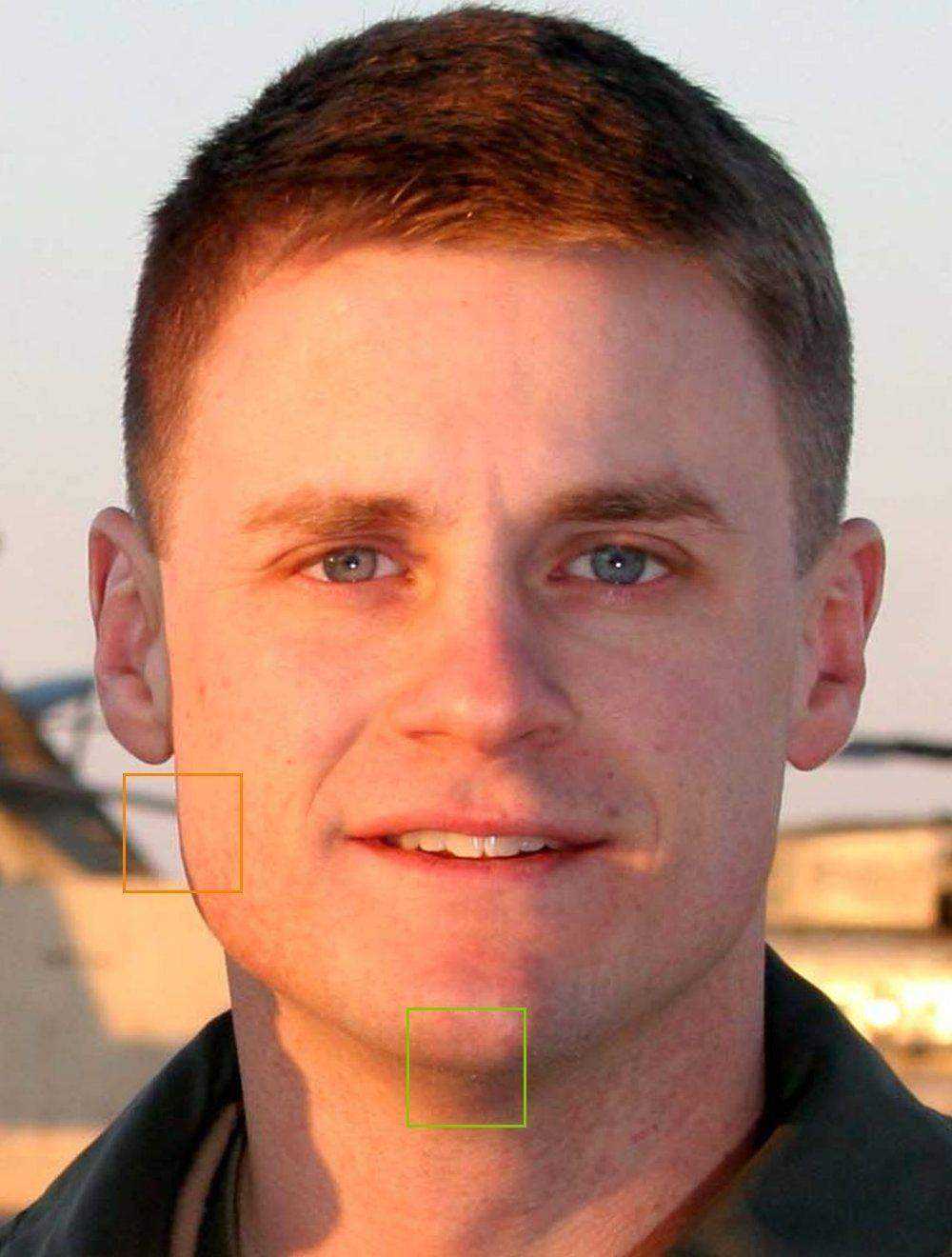}&
\includegraphics[width=0.48\linewidth]{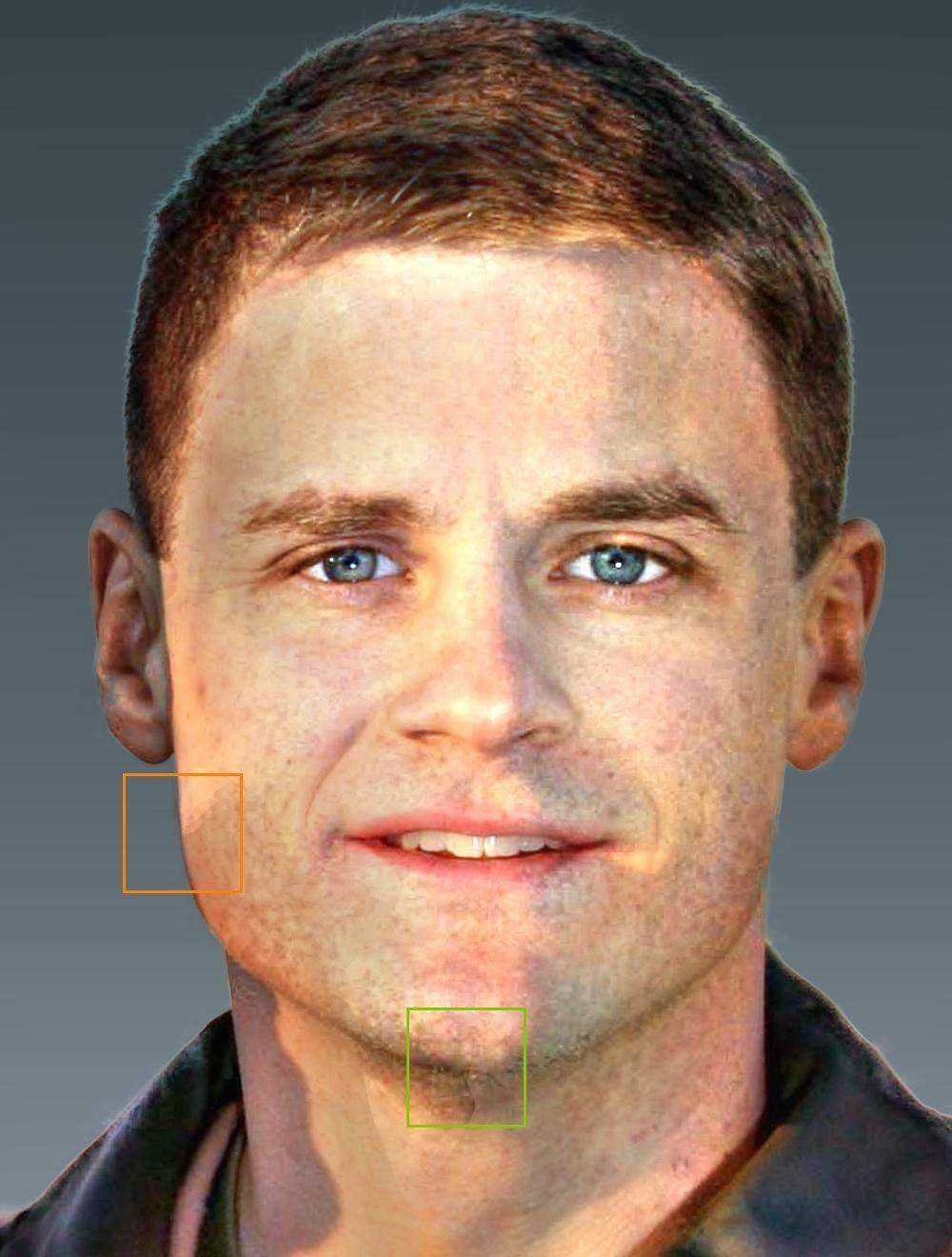}\\
\includegraphics[width=0.23\linewidth]{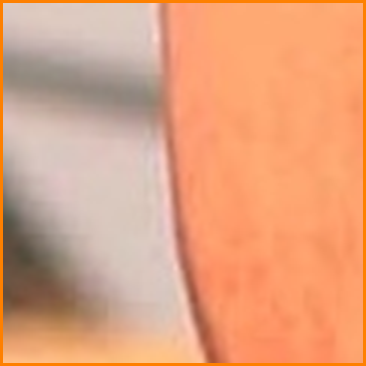}
\includegraphics[width=0.23\linewidth]{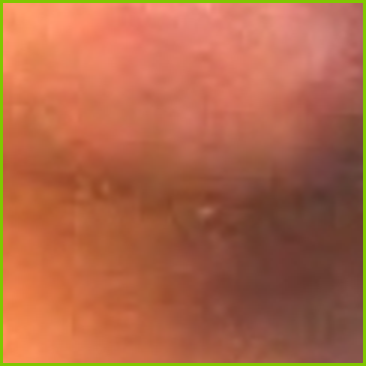}&
\includegraphics[width=0.23\linewidth]{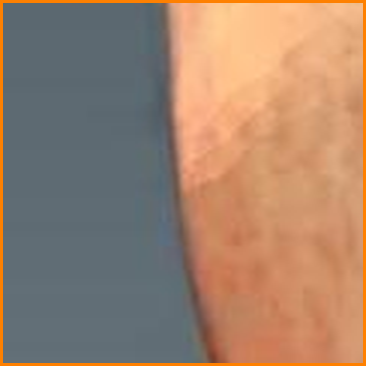}
\includegraphics[width=0.23\linewidth]{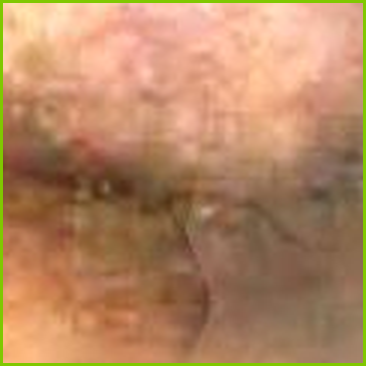}\\
(a) Input photo&(b) Remapped\\
\includegraphics[width=0.48\linewidth]{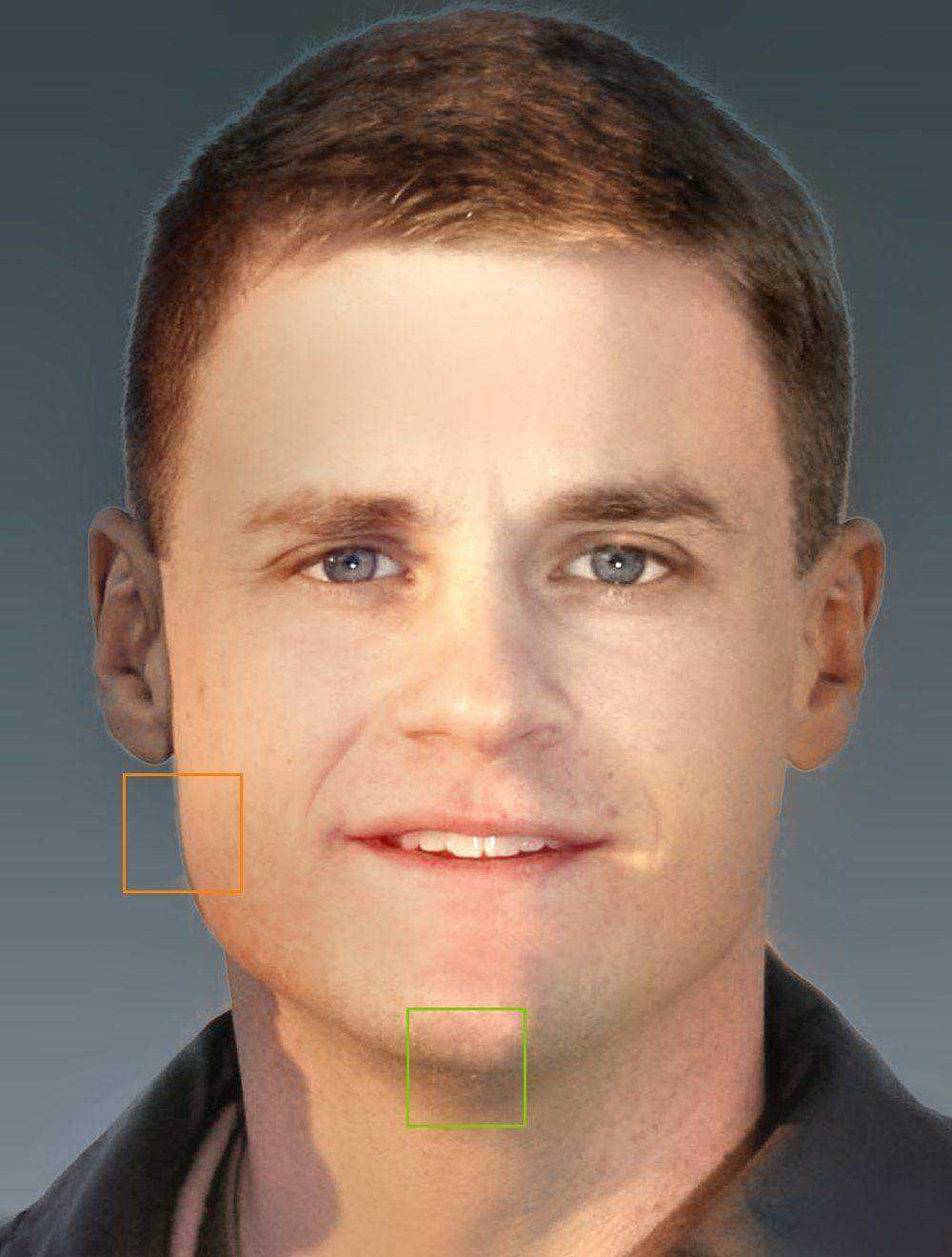}&
\includegraphics[width=0.48\linewidth]{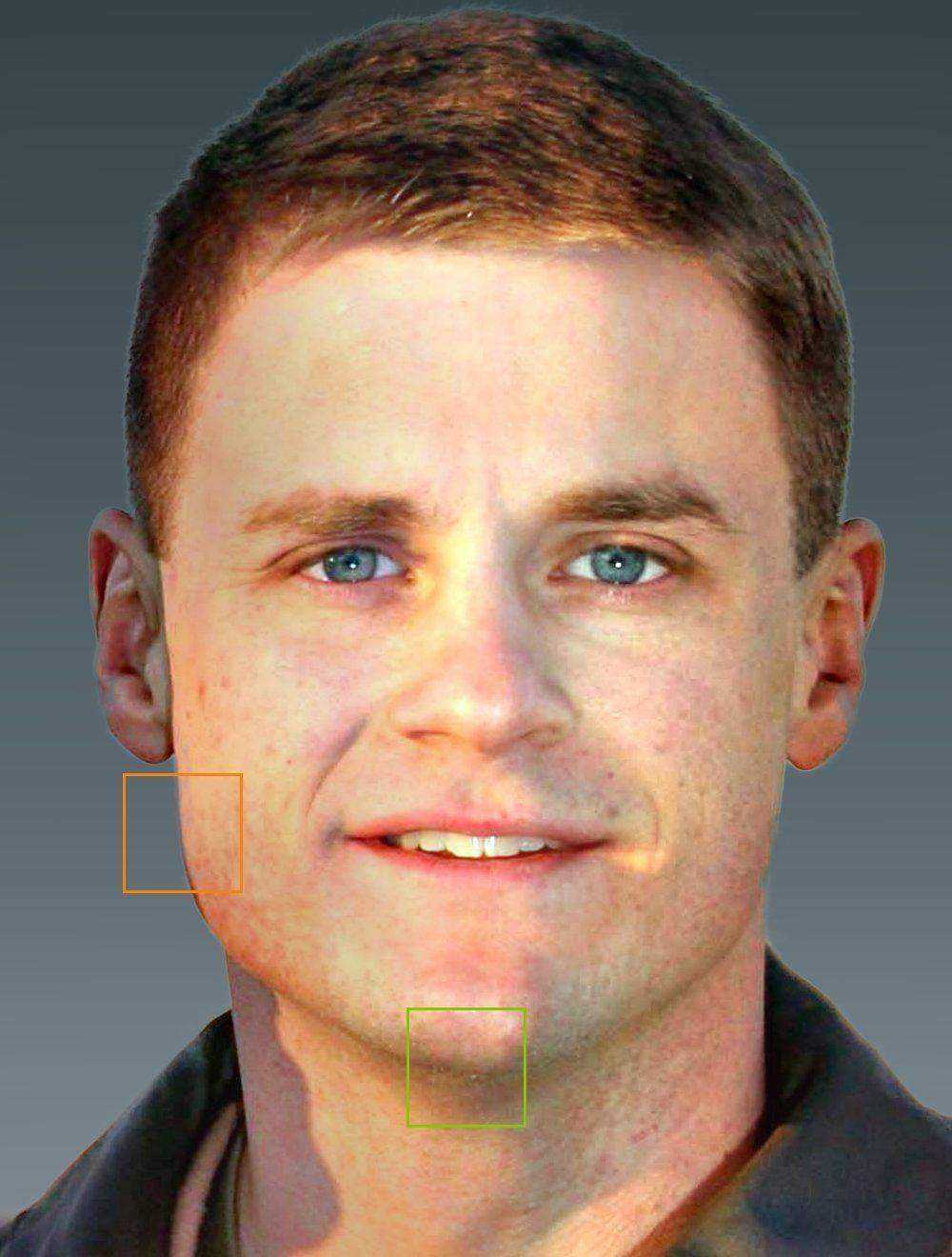}\\
\includegraphics[width=0.23\linewidth]{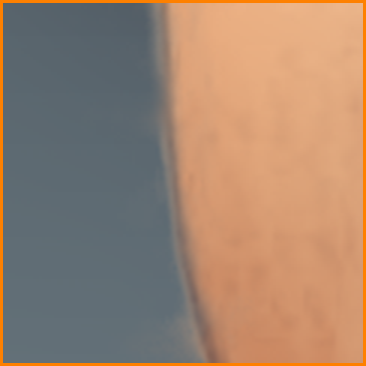}
\includegraphics[width=0.23\linewidth]{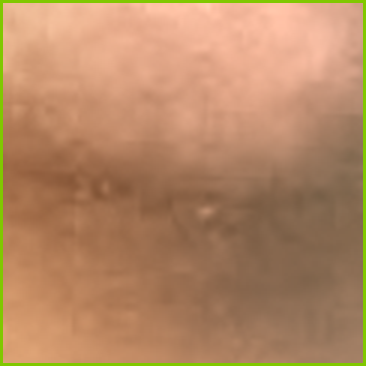}&
\includegraphics[width=0.23\linewidth]{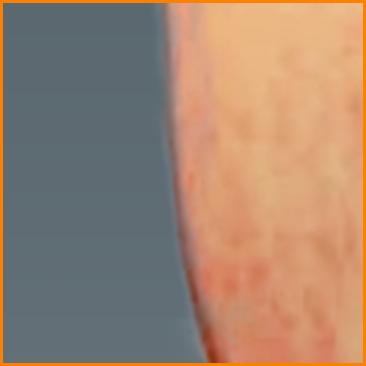}
\includegraphics[width=0.23\linewidth]{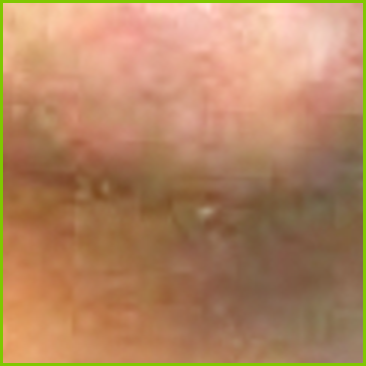}\\
(c) Guided filtered&(d) Output\\
\end{tabular}
\end{center}
\caption{Artifact removal.
The artifact occurred on the remapped result generated in Section \ref{sec:lr} is shown in (b). {We use guided filter to smooth such artifact (resulted image shown in (c))} and add back the details of
an input photo to generate the output in (d).}
\label{fig:refine}
\end{figure}

\renewcommand{\tabcolsep}{0.1pt}

\subsection{Local Remapping}
\label{sec:lr}
We decompose the input photo and every
exemplar separately into a Laplacian stack formulation.
A Laplacian stack consists of multiple layers among which the last one is the
residual and the remaining ones are the subtracted result of two Gaussian filtered
images with increasing radius.
For each layer, a local energy map $S$ is generated by locally averaging
the squared layer values.
\re{These local energy maps from the exemplars represent the style to be transferred
to the input photo.}
The goal of the local contrast transfer is to update all the layers in the Laplacian stacks
of the input image such that the energy distributions are similar to those in the exemplars.
%
We transfer local contrast at each pixel
location from multiple exemplars using the local patch selection method described
in Section \ref{sec:lm}.

We denote $L_p^l$ and $S_p^l$ as the values of pixel $p$ at
the $l$-th Laplacian layer and energy map, respectively.
The local remapping function at pixel $p$ can be written as:
\begin{equation}
L_p^l(R)=L_p^l(T)\times\sqrt{\frac{S_p^l(E)}{S_p^l(T)+\epsilon}}
\label{eq:mapping}
\end{equation}
where $R$ is the remapped image patch, $E$ is the patch in the exemplar photo selected at
pixel $p$, and $\epsilon$ is a small number to avoid division by
zero.
We locally remap the input photo in all the layers except the residual which
only contains low frequency components.
When we generate the residual layer of an output image, we use the values
from the residual layer of the identified exemplars.
\re{After this step with local energy maps, we accumulate all the layers in the Laplacian stack of the input photo.
Since a Laplacian stack is constructed based on the subtracted results of a Gaussian filtered image at different scales, the accumulation of
all the transferred layers is used to generate the stylized output.}

\begin{figure}[t]
\begin{center}
\begin{tabular}{cc}
\includegraphics[width=.48\linewidth]{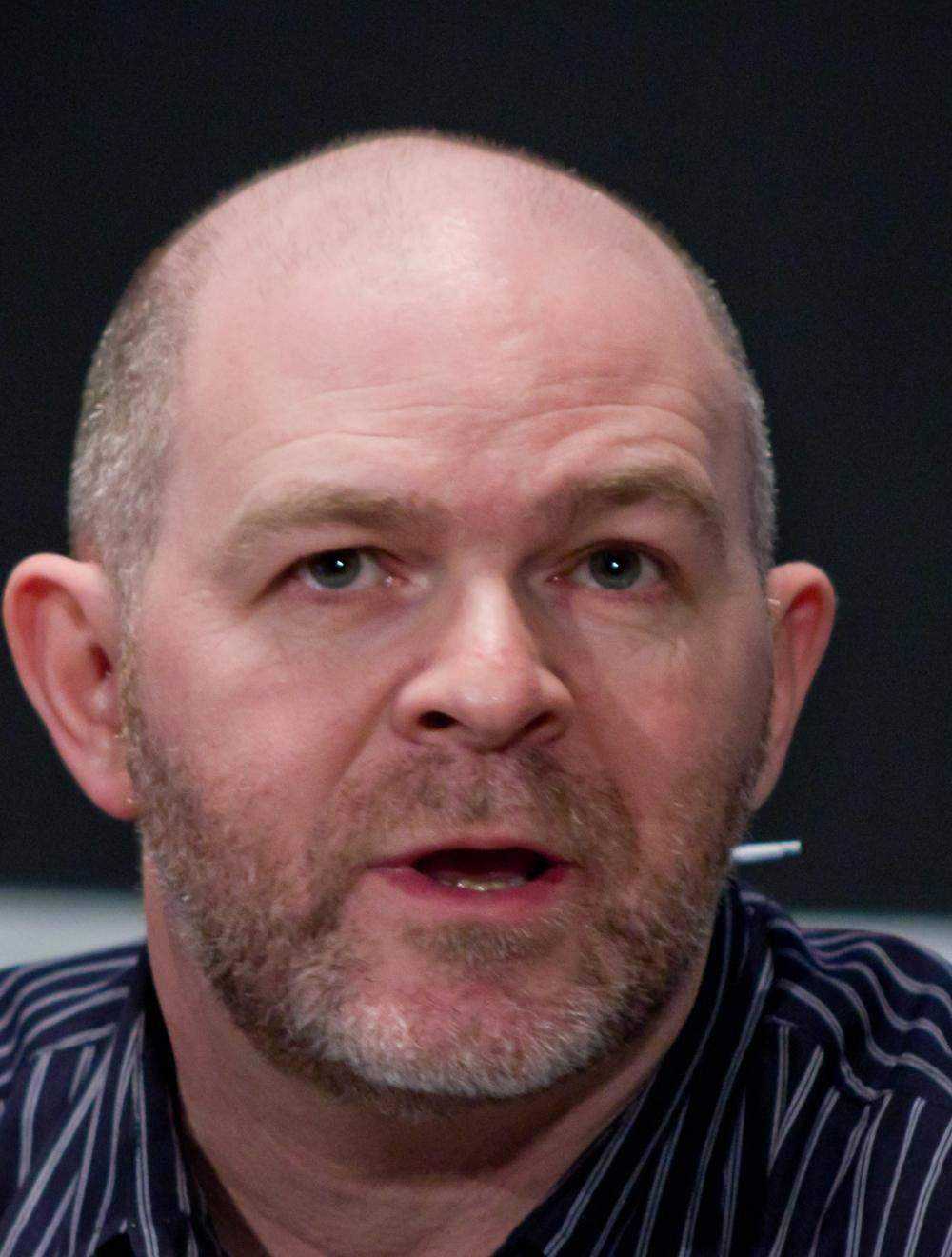}&
\includegraphics[width=.48\linewidth]{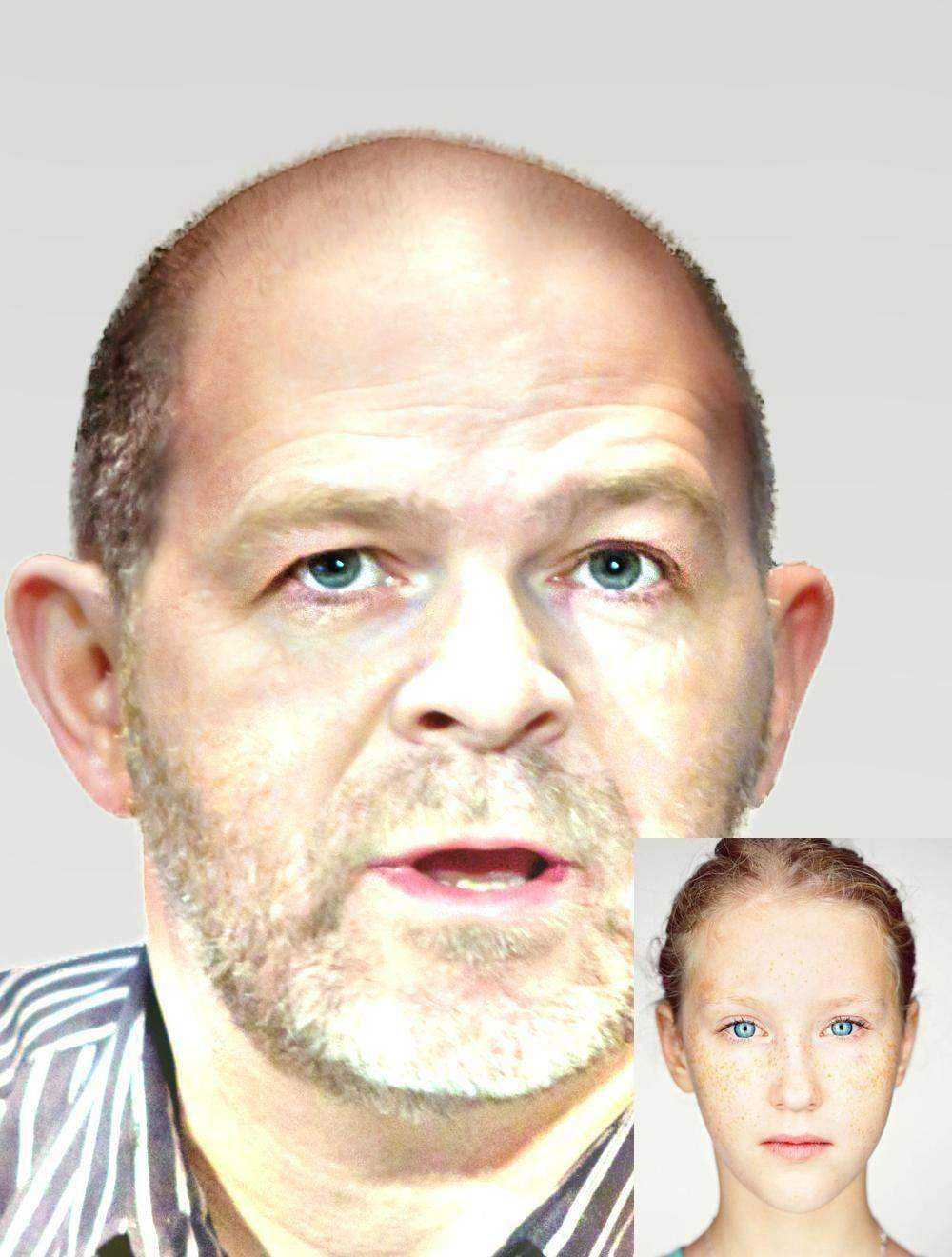}\\
(a) Input photo&(b) Local \cite{Shih-siggraph14-StyleTransfer}\\
\includegraphics[width=.48\linewidth]{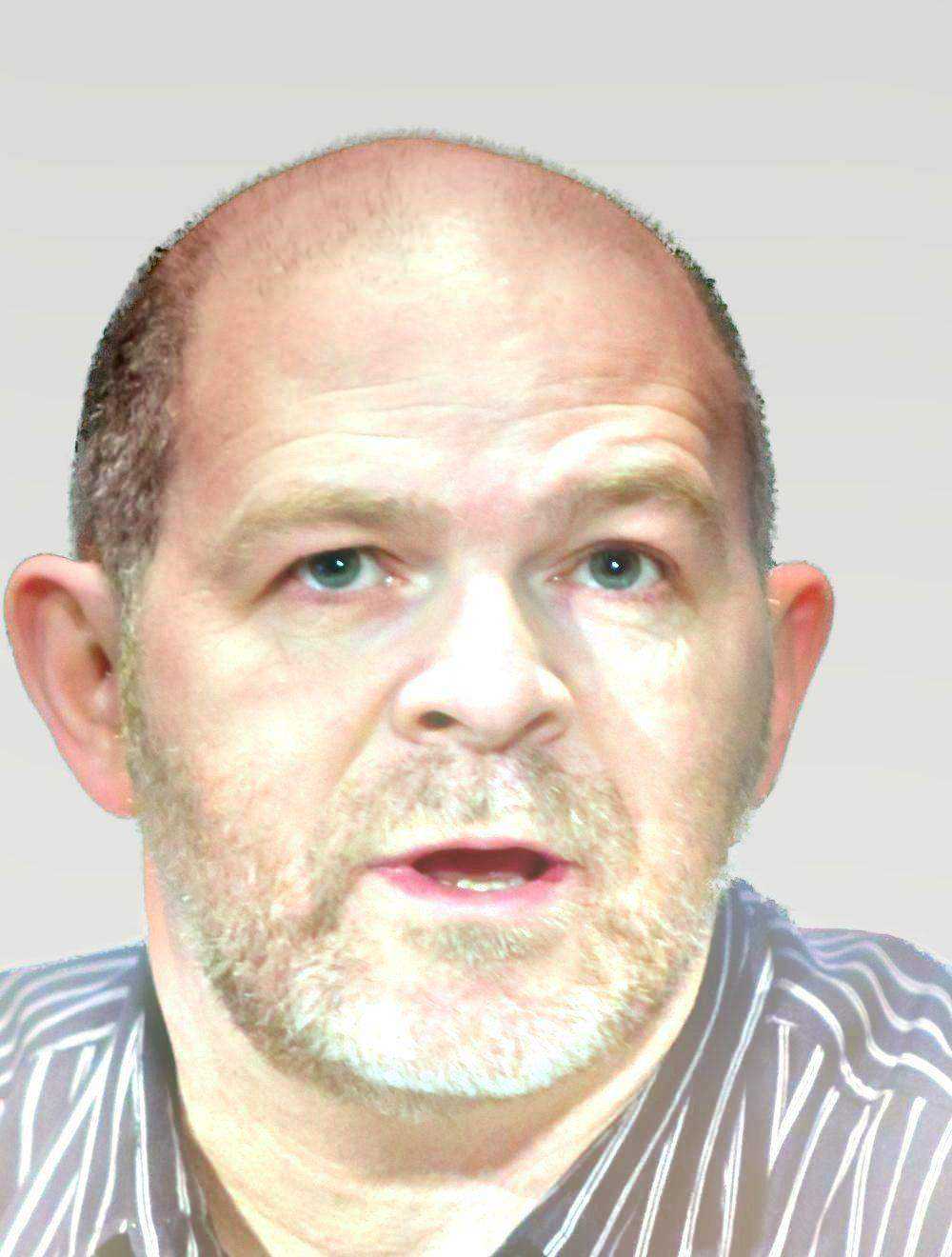}&
\includegraphics[width=.48\linewidth]{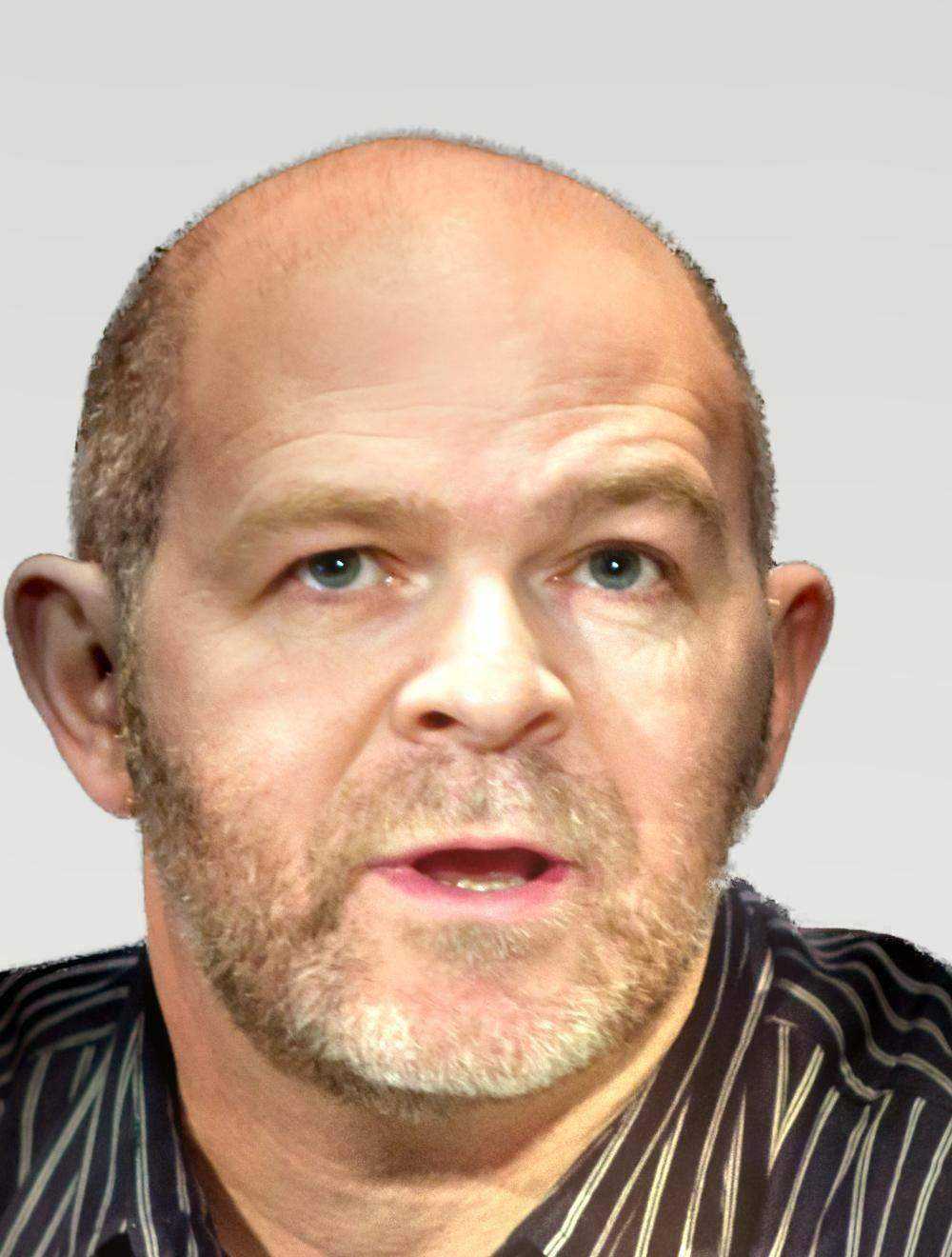}\\
(c) Post processing on (b)&(d) Proposed\\
\end{tabular}
\end{center}
\caption{Relationship between exemplar matching and post processing. (a) is the input photo and (b) is the result of \cite{Shih-siggraph14-StyleTransfer}.
The difference on chin (beard) between the exemplar and input photo produces artifacts on the transferred result. It can not be effectively removed through post processing step shown in (c). Through multiple exemplar matching the proposed method can perform correct local transfer and artifacts are effectively suppressed as shown in (d).}
\label{fig:analysis}
\end{figure}

\subsection{Artifact Removal}
\label{sec:pp}
\re{We aggregate each layer in the Laplacian stack to generate the remapped output image}.
\re{As local patches from multiple examples are selected between neighboring pixels, each remapped output image is likely
to contain artifacts around the facial component boundaries}.
Figure \ref{fig:refine}(b) shows one example that contains
artifacts due to inconsistent local patches.
As such, we use an edge-preserving filter
\cite{Petschnigg-siggraph04-JBF,Eisemann-siggraph04-JBF} to remove
artifacts and retain facial details.
We use the input photo as guidance to filter the remapped result.
\re{The artifacts are removed using an edge-preserving filter
at the expense of missing local details}.
\re{Nevertheless, these details are recovered through creating a similar blurry scenario that we use the input photo as guidance to filter itself}.
\re{The differences between the filtered result and the input photo are the missing details on the remapped result}.
\re{We transfer the details back to the filtered result to minimize over-smoothing effects}.
Consequently,
the holistic tone and local contrast can be well maintained in the final output
while artifacts are effectively removed.

Figure \ref{fig:refine} shows the main steps of the artifact removal process.
Given an input photo,  we use the matting method
\cite{levin-pami2008-matting} to substitute its original background with a
predefined background.
We then use the guided filter \cite{kaiming-pami2013-GuidedFilter}
to smooth the remapped result with
the input photo as guidance, as shown in Figure~\ref{fig:refine}(c).
The radius of the guided filter is set relatively large to remove the
artifacts on the remapped result.
The downside of filtering using a large radius is that the filtered images
are likely to be over-smoothed.
However, we can alleviate this problem with the
help of the input photo.
First we use the guided filter to smooth the input photo
using itself as guidance.
The filter radius is set the same as the previous filtering
process on the remapped result.
The missing details can then be obtained by subtracting the filtered result using the input photo.
Finally, we add back the missing details to the smoothed remapped image
and generate the final result shown in Figure \ref{fig:refine}(d).

\begin{algorithm}[t]
\caption{Proposed Face Style Transfer Algorithm}
\begin{algorithmic}[1]
\For{each exemplar $E$}
    \State Compute Laplacian stack $L$ and local energy $S$;
    \State Generate dense correspondence with an input photo $T$;
    \State Warp $S$ according to dense correspondence.
\EndFor
\State Select exemplar patch using  Markov random field;
\State Compute Laplacian stack $L$ and local energy $S$ for $T$;
\For{each layer $L$ of $T$}
    \For{each pixel $p$ in $L$}
        \If {not residual}
            \State local contrast transfer using Eq. \eqref{eq:mapping};
        \Else
            \State select from exemplar residuals;
        \EndIf
    \EndFor
\EndFor
\State Aggregate output stack to obtain remapped result $R$;
\State Guided filtering $R$ using $T$ as guidance to obtain $R^T$;
\State Guided filtering $T$ using $T$ as guidance to obtain $T^T$;
\State Output is obtained through $R^T+T-T^T$.
\end{algorithmic}
\label{algo:code}
\end{algorithm}

\subsection{Discussion}
\label{sec:pp}
\begin{figure*}[!ht]
\begin{center}
\begin{tabular}{cccc}
\includegraphics[width=.245\linewidth]{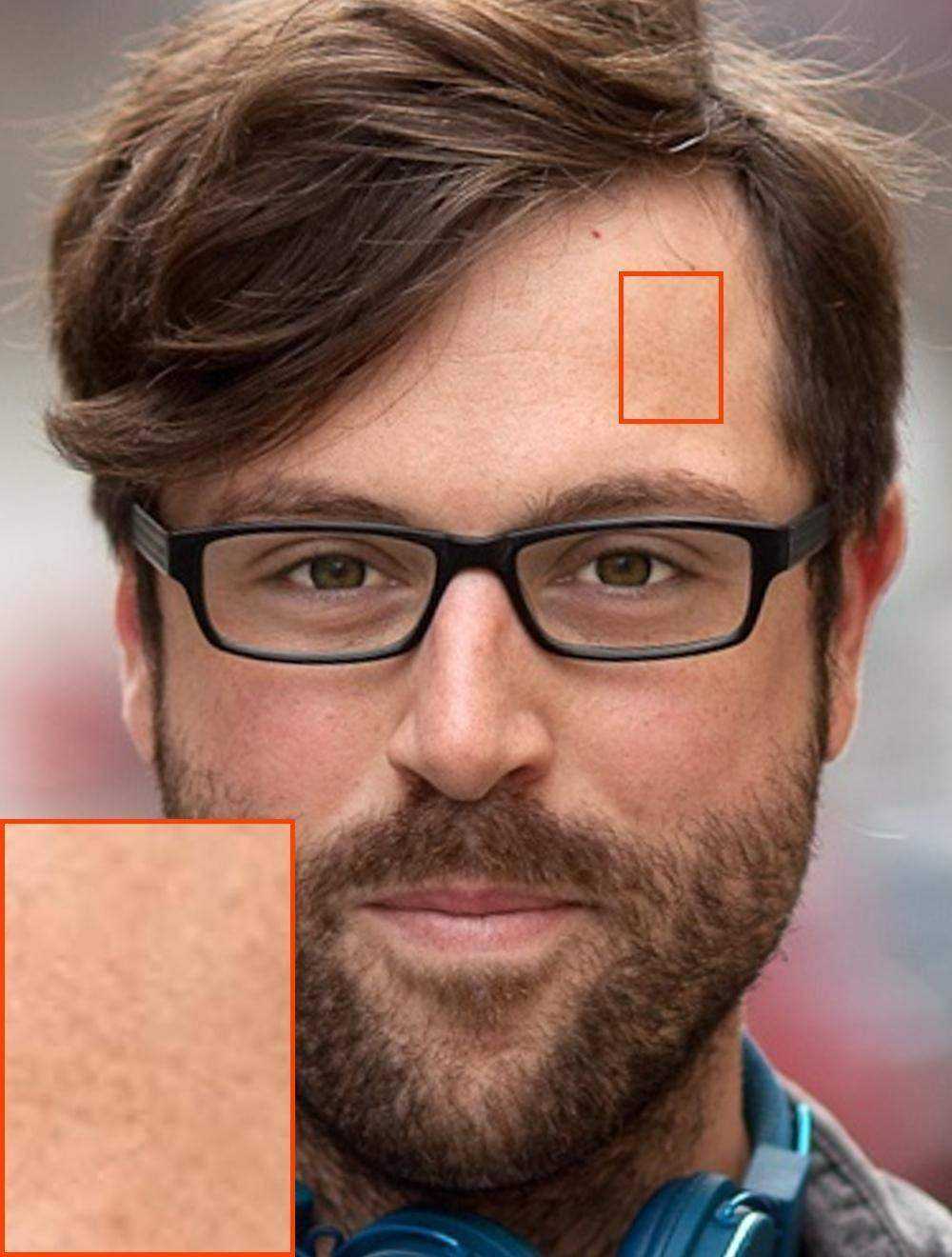}&
\includegraphics[width=.245\linewidth]{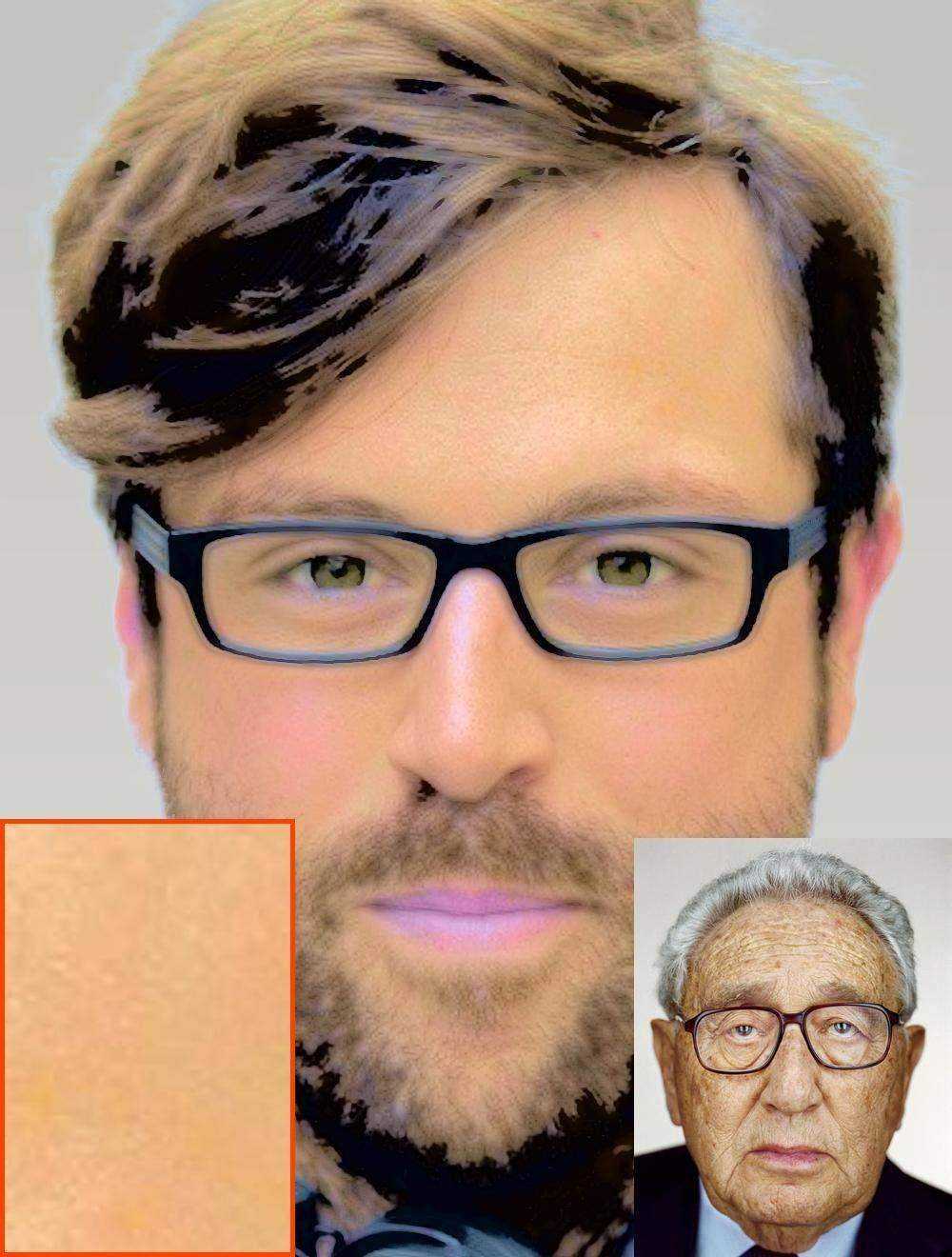}&
\includegraphics[width=.245\linewidth]{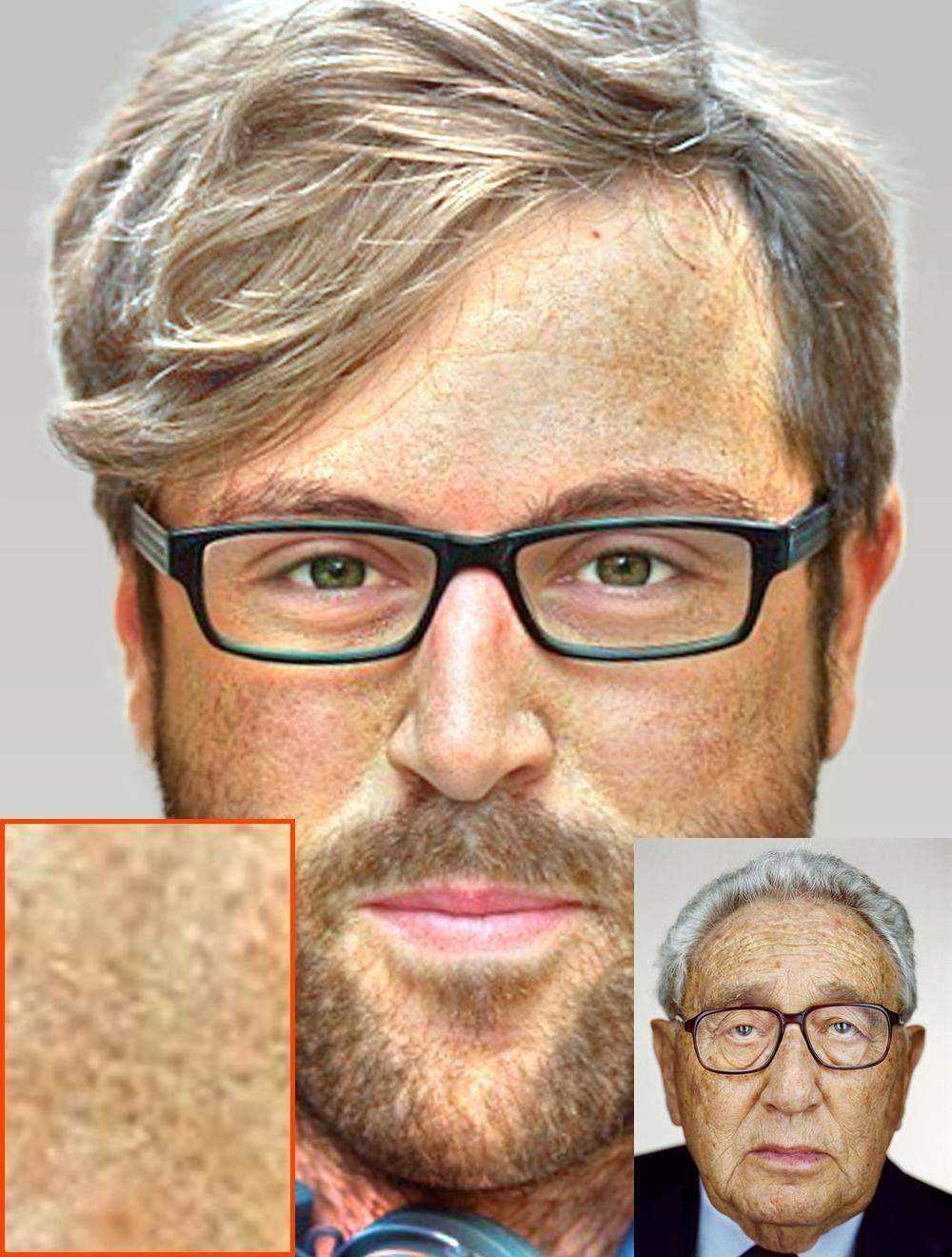}&
\includegraphics[width=.245\linewidth]{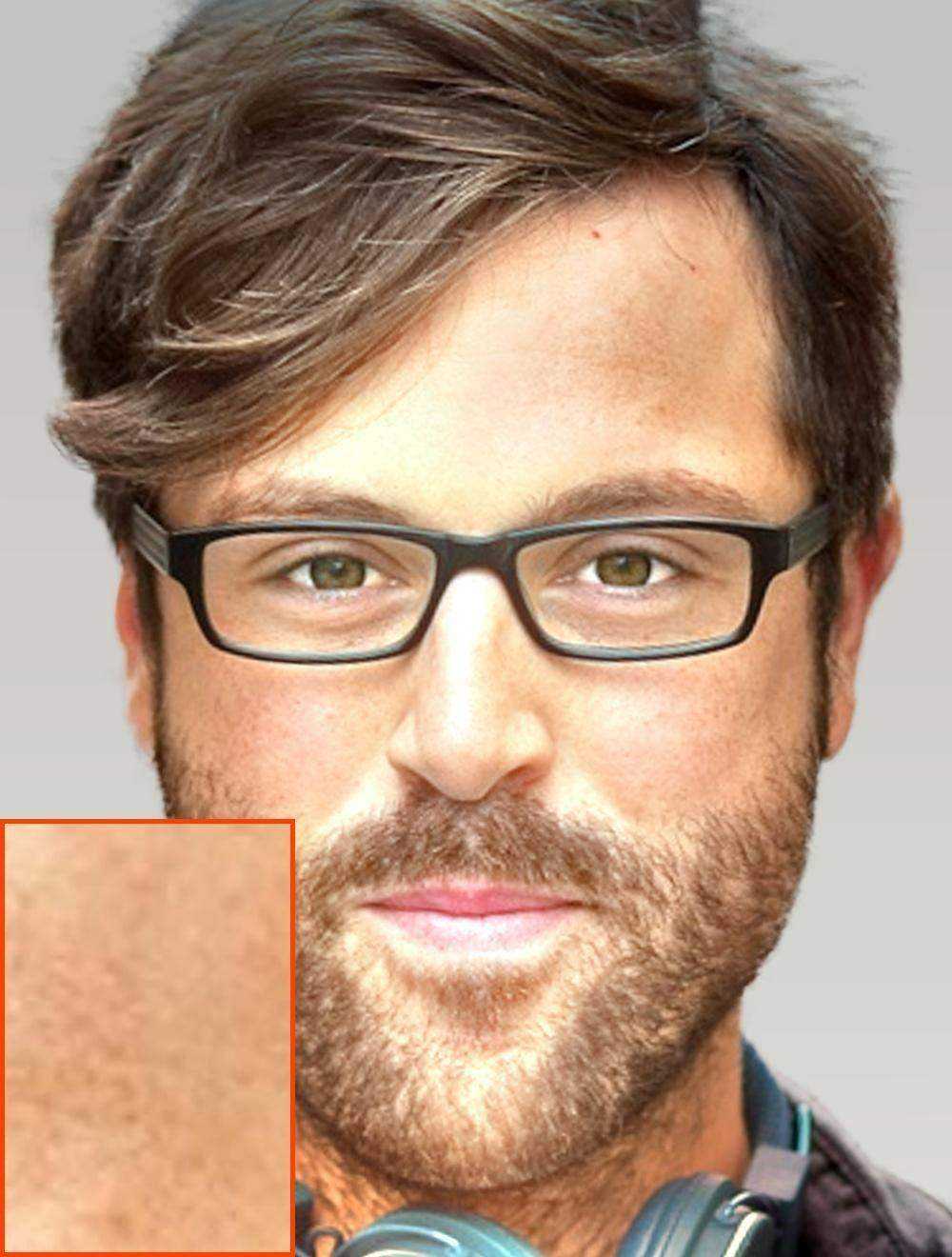}\\
\includegraphics[width=.245\linewidth]{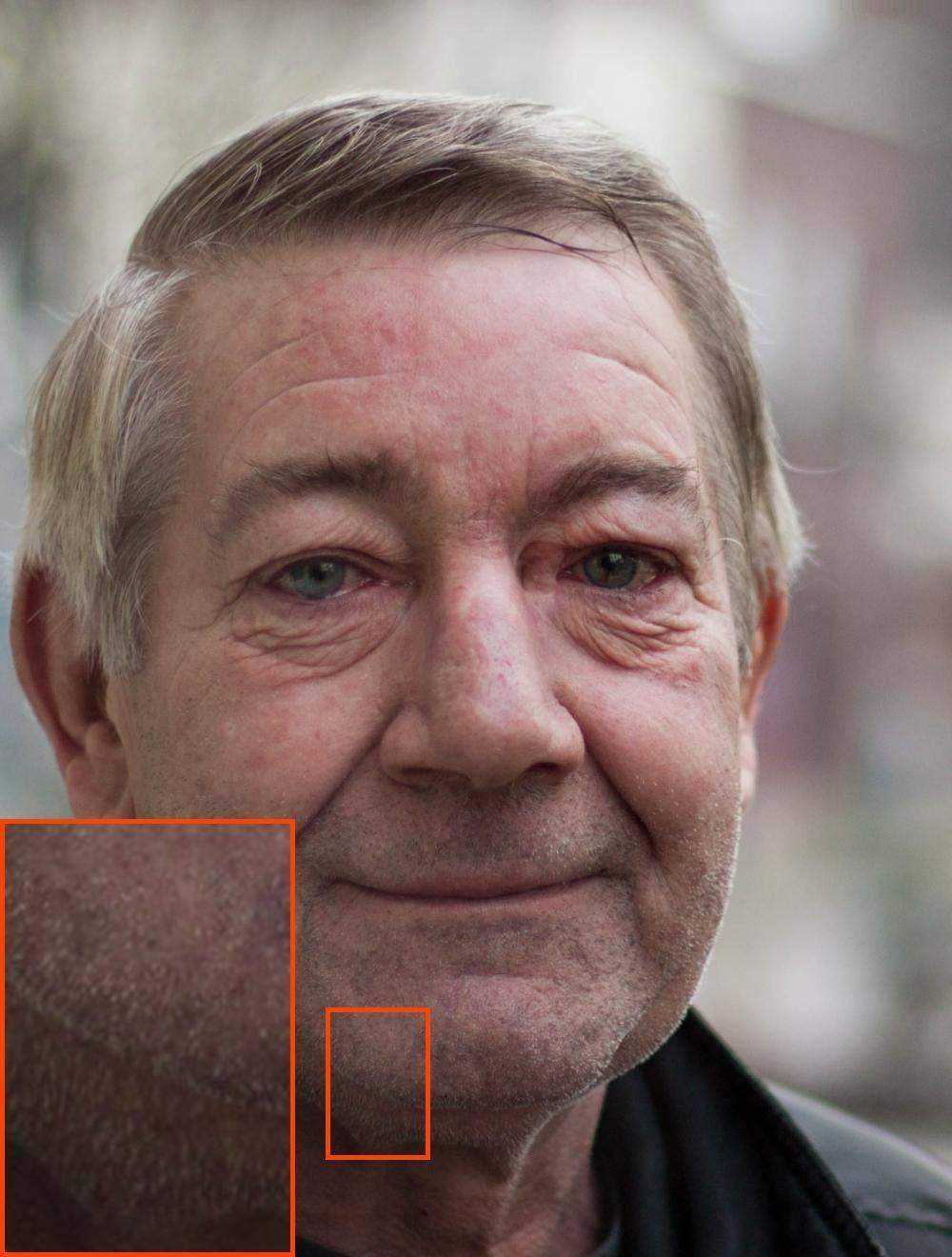}&
\includegraphics[width=.245\linewidth]{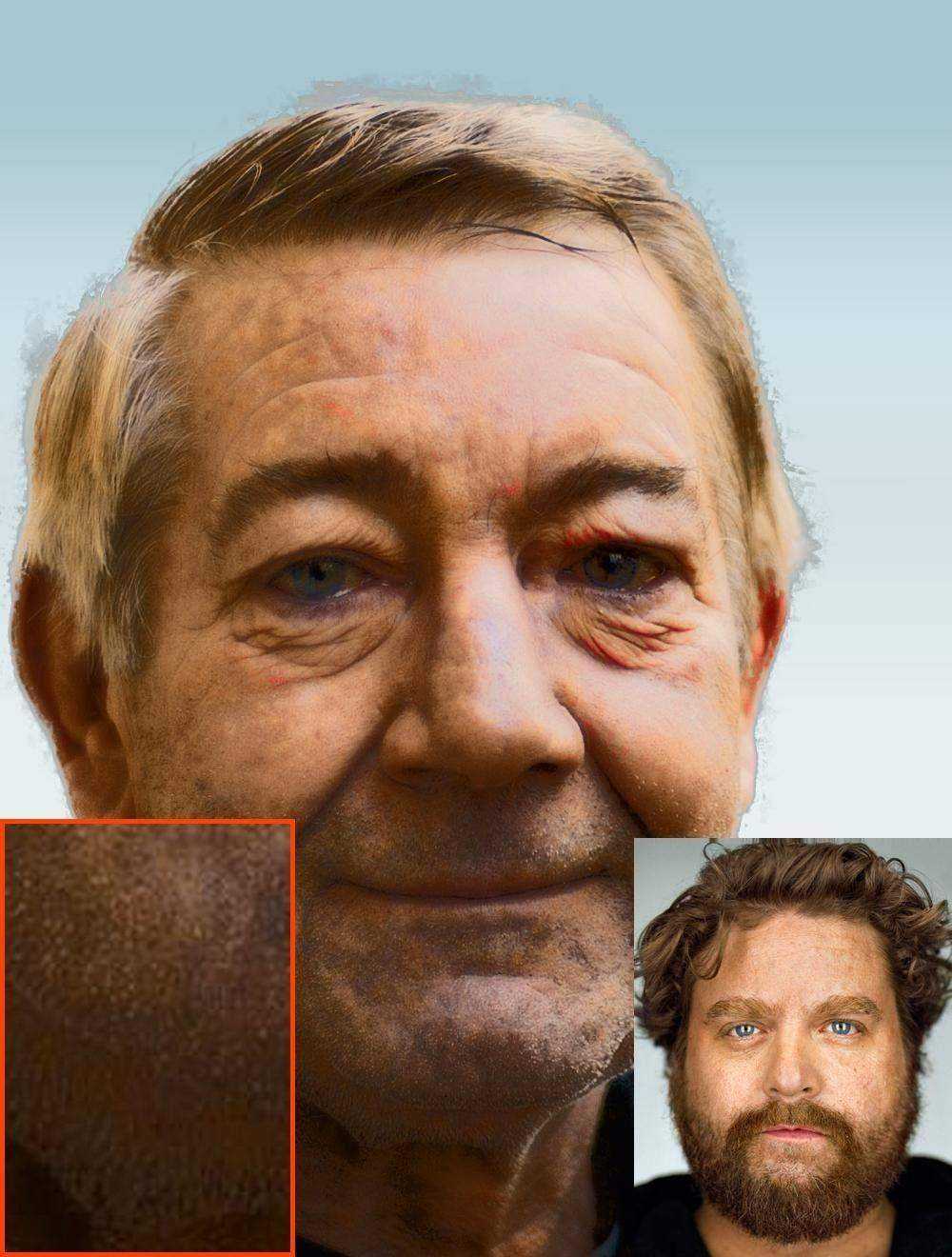}&
\includegraphics[width=.245\linewidth]{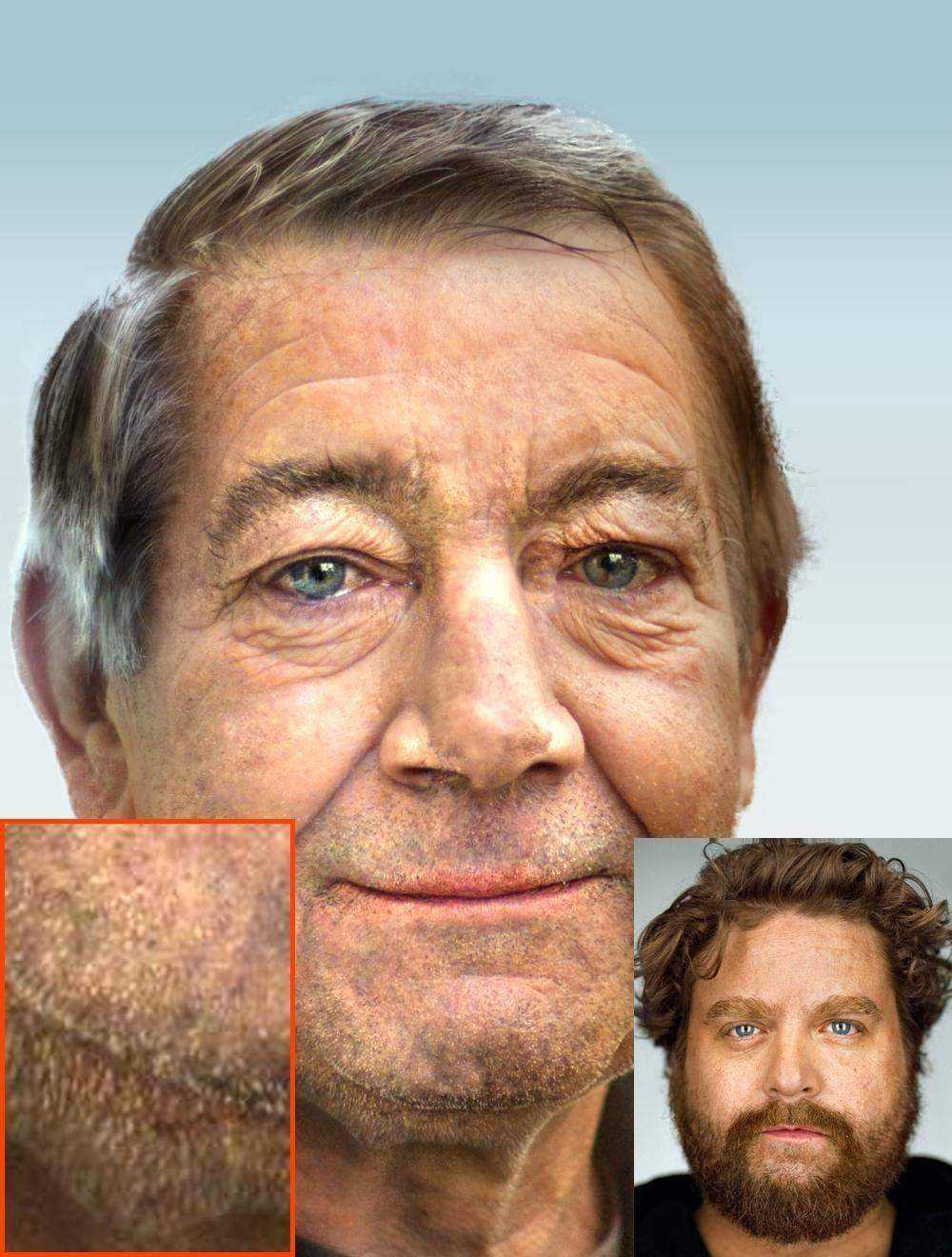}&
\includegraphics[width=.245\linewidth]{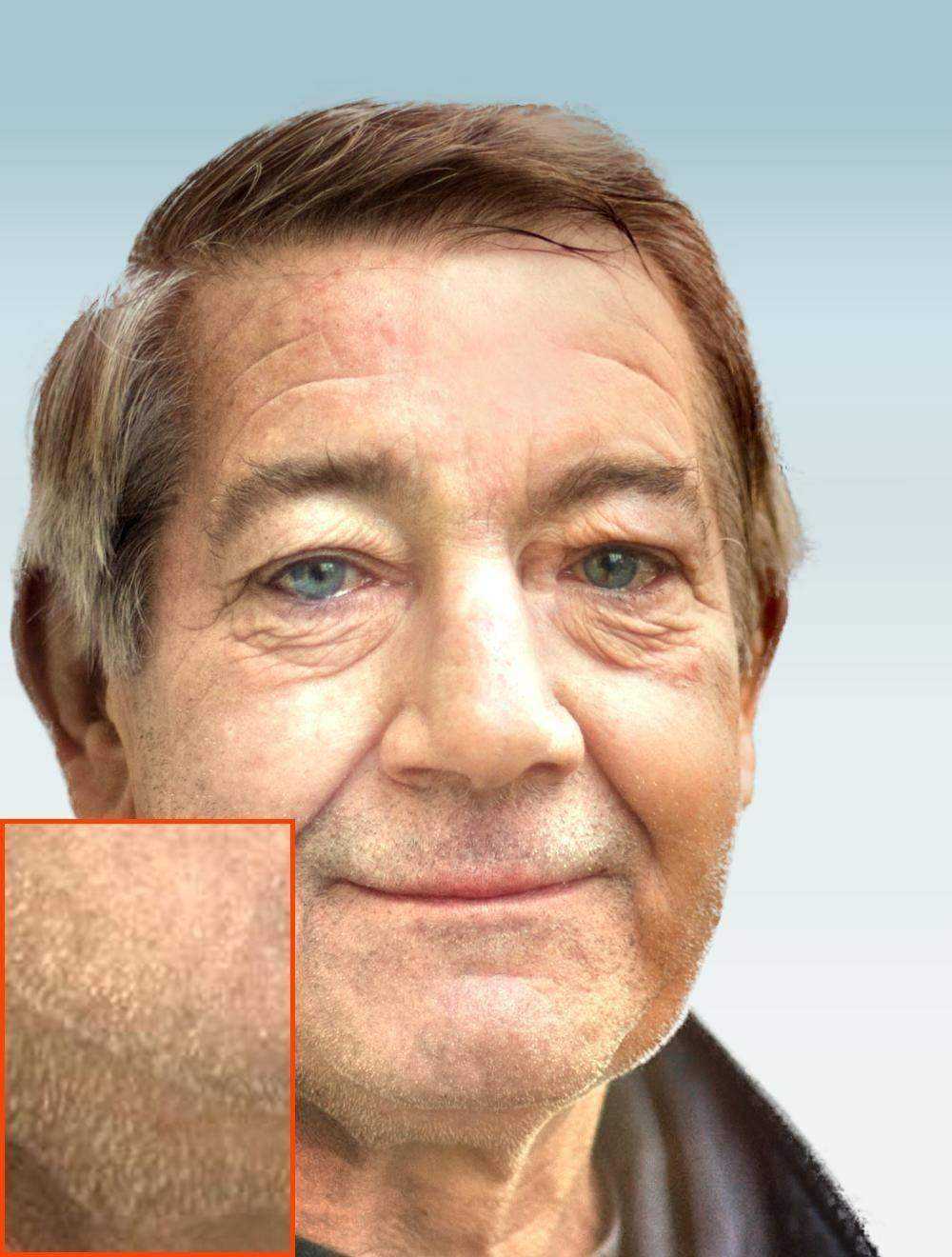}\\
\includegraphics[width=.245\linewidth]{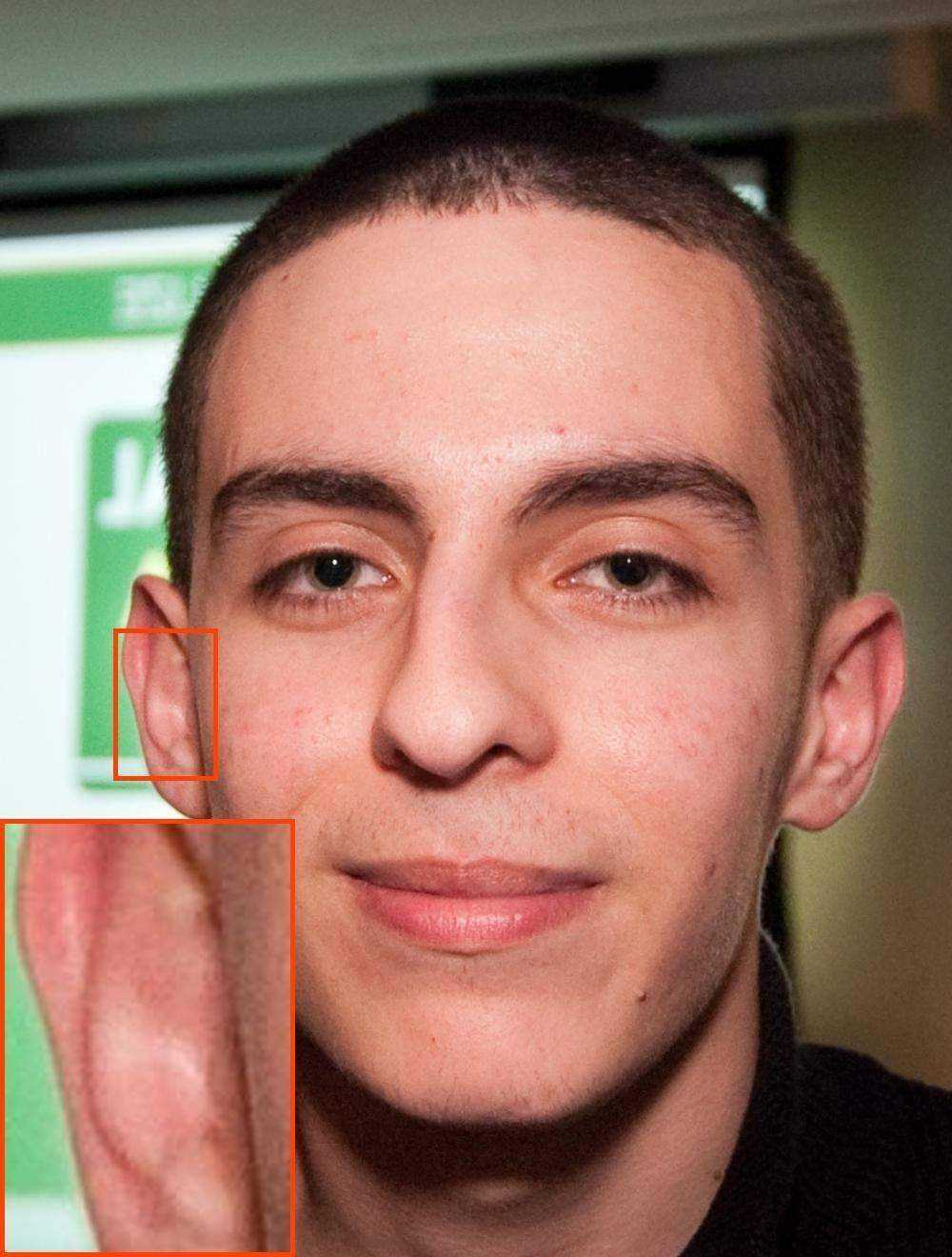}&
\includegraphics[width=.245\linewidth]{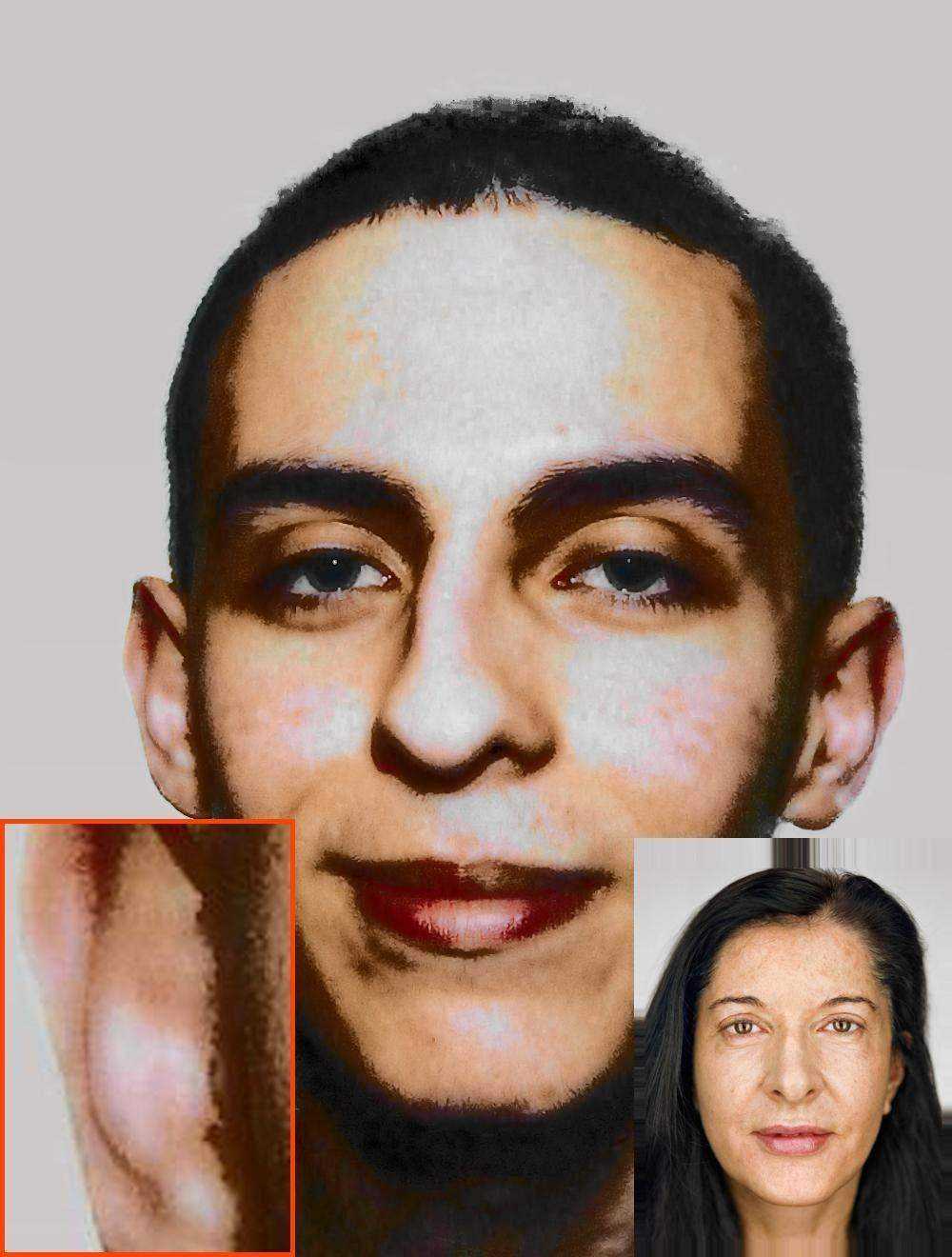}&
\includegraphics[width=.245\linewidth]{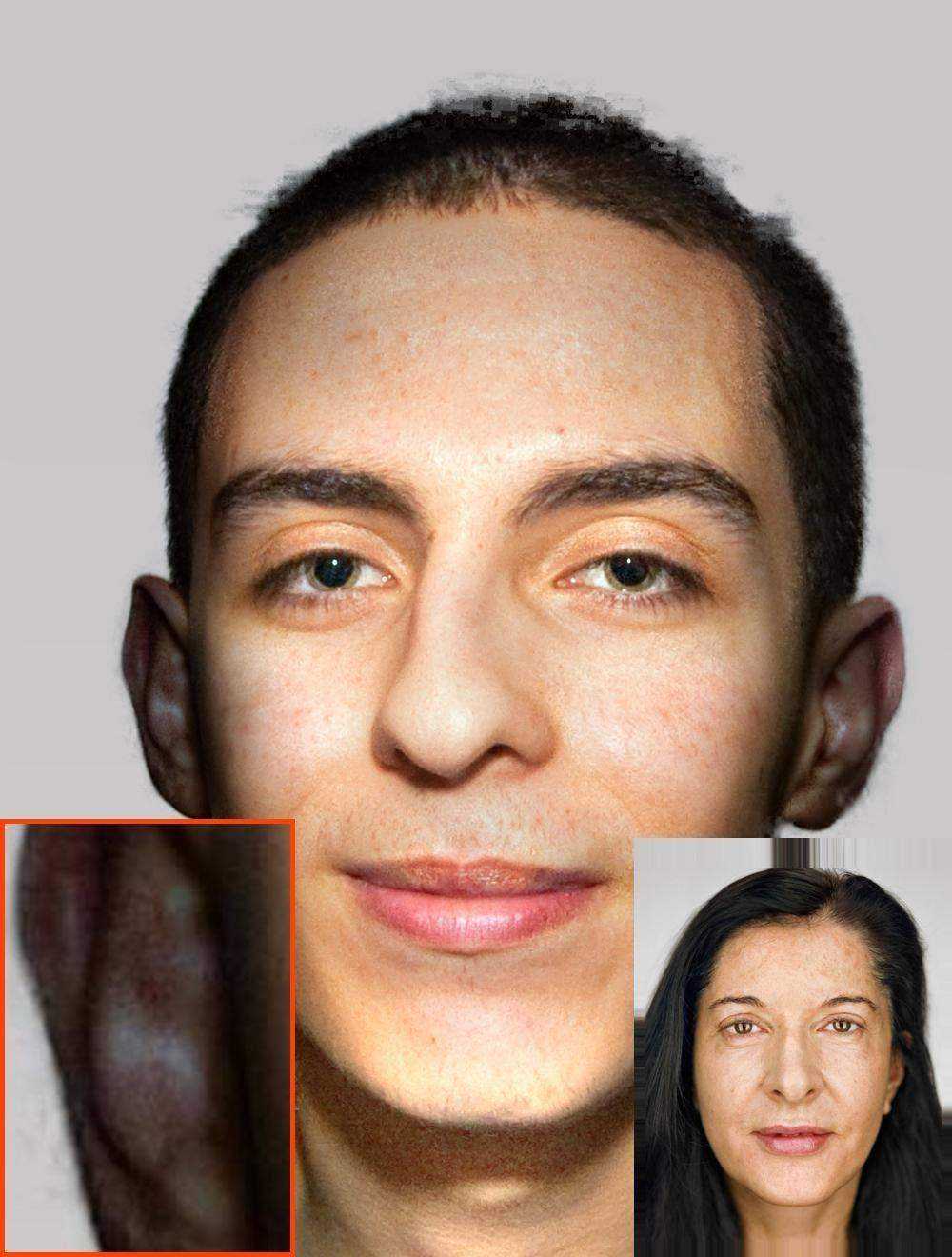}&
\includegraphics[width=.245\linewidth]{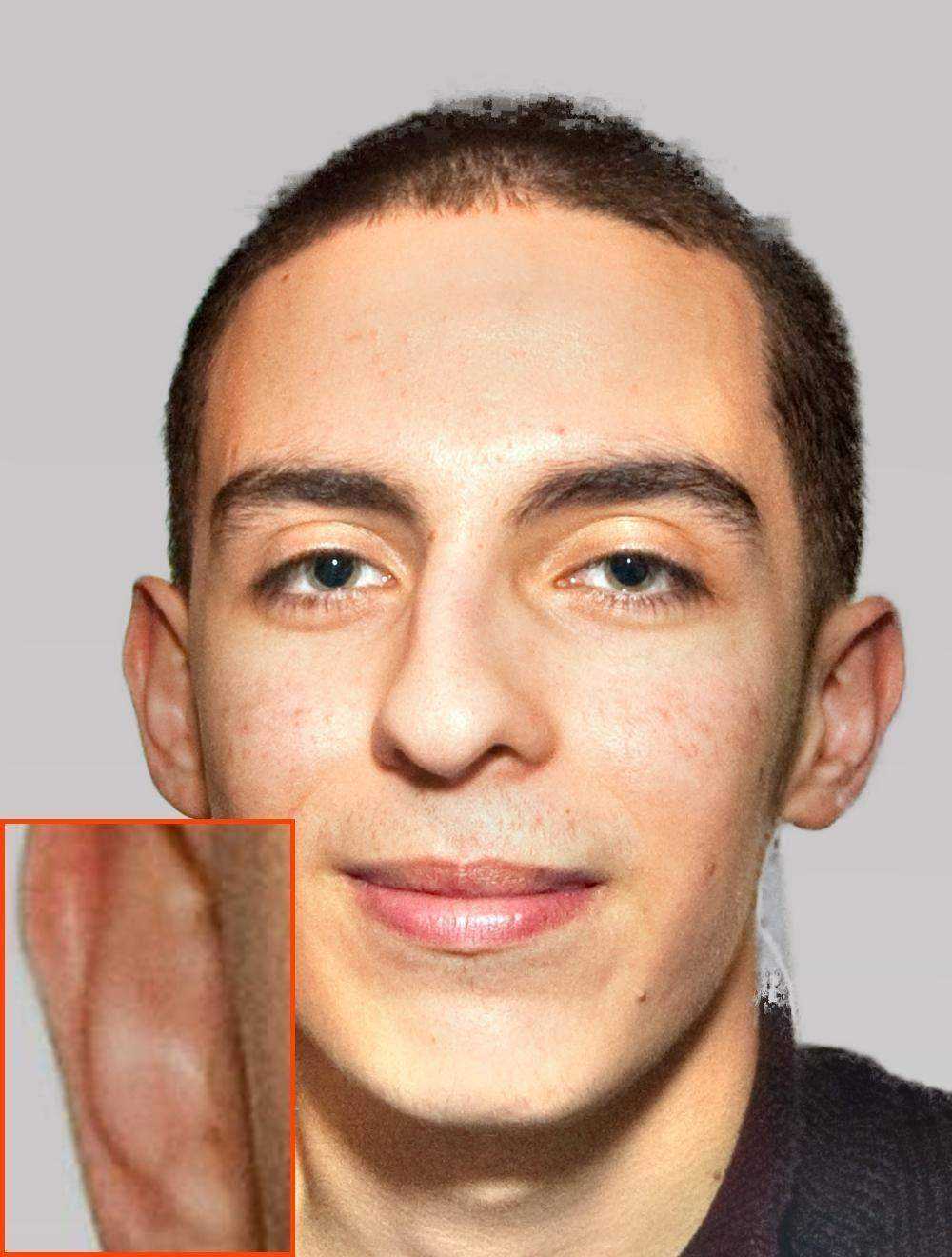}\\
(a) Input photo&(b) Holistic \cite{Sunkavalli-siggraph10-Harmonization}&(c) Local \cite{Shih-siggraph14-StyleTransfer}&(d) Proposed\\
\end{tabular}
\end{center}
\caption{Qualitative evaluation on the Martin dataset. The proposed
  method performs favorably against holistic and local methods.
These images can be better visualized with zoom-in to analyze the details.}
\label{fig:exp2}
\end{figure*}

\re{We note that the main contribution to the high-quality stylized images is the selection of propoer local patches from multiple exemplars rather than removal of artifacts}.
We show one example in Figure \ref{fig:analysis} where the stylized image is obtained by the state-of-the-art method
\cite{Shih-siggraph14-StyleTransfer} and post-processed by the artifact removal process discusvsed above.
Without correct exemplars selection, the artifacts in the stylized image can not be removed.
On the other hand, the proposed algorithm transfers low frequency components from multiple exemplars
while preserving high frequency contents of the input photo.
Figure \ref{fig:refine}(c) and (d) show one example where the guided filter is used to suppress inconsistent artifacts
(due to MRF regularization) and maintain high frequency details in the input photo.
In contrast, the state-of-the-art methods may fail to transfer high frequency details from exemplars.
Another example is shown in Figure \ref{fig:exp2}(c) where
the undesired textures such as wrinkles or beard are wrongly transferred to the output image.

The main steps of proposed style transfer algorithm are summarized in Algorithm \ref{algo:code}.

\begin{figure}[t]
\begin{center}
\begin{tabular}{ccc}
\includegraphics[width=.33\linewidth]{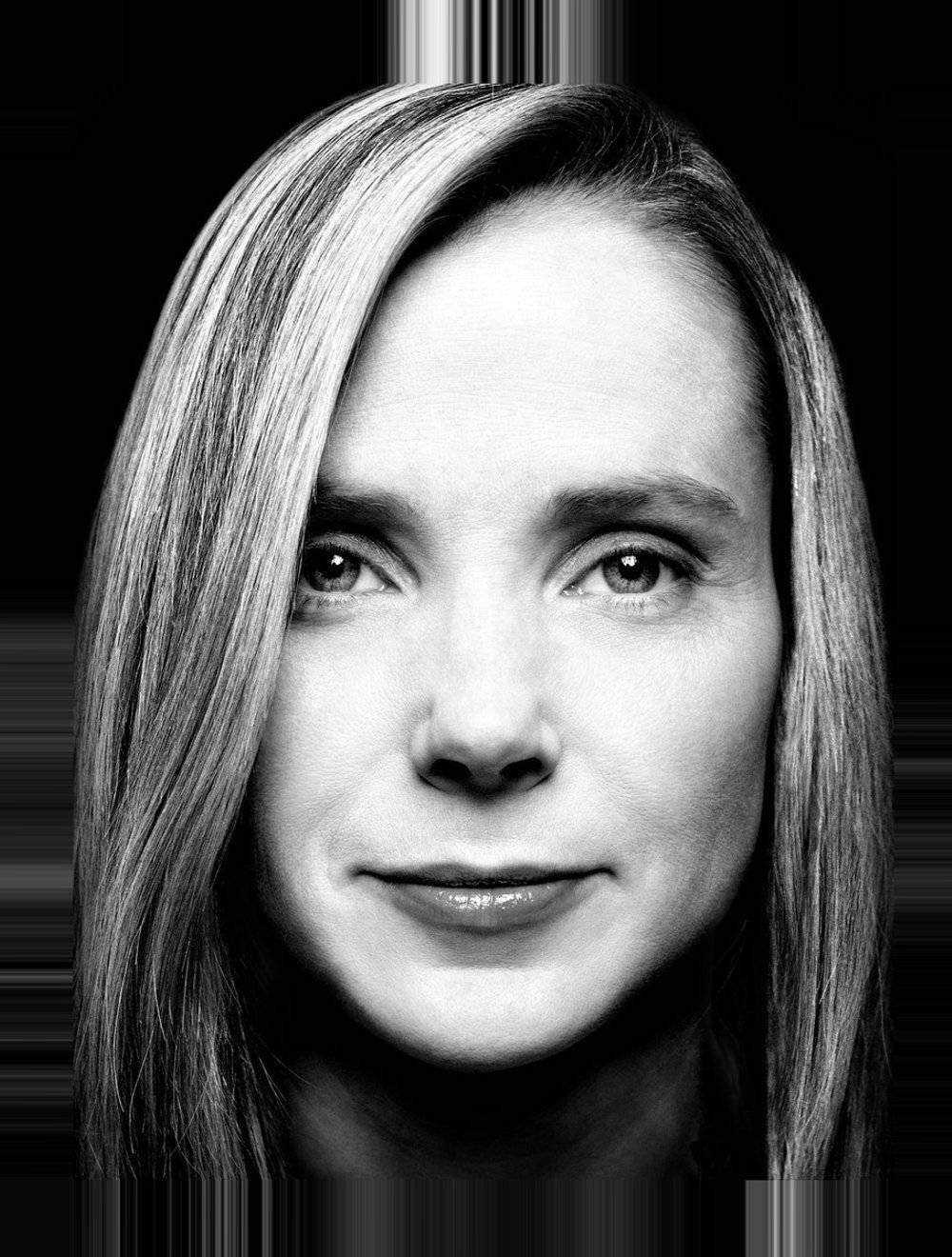}&
\includegraphics[width=.33\linewidth]{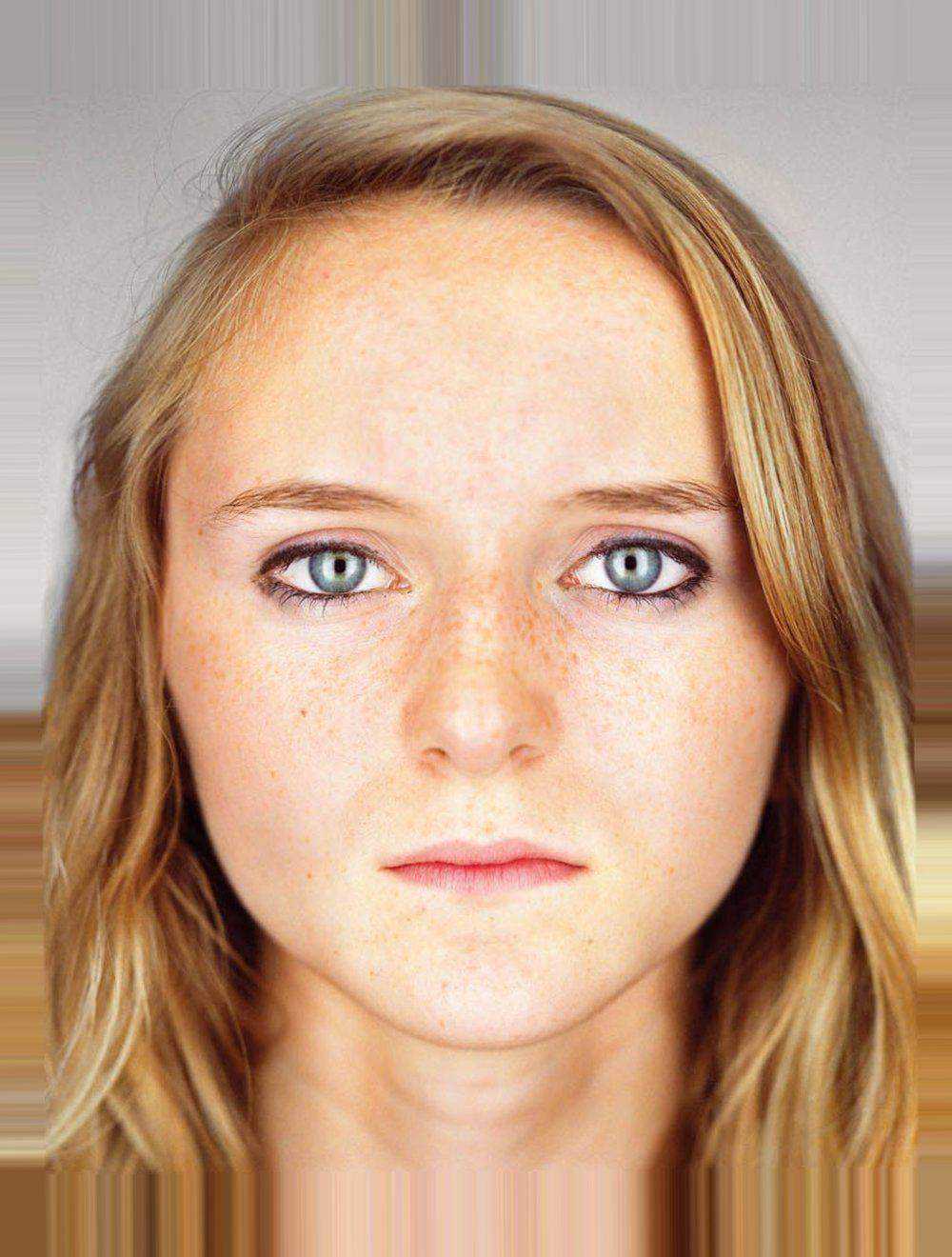}&
\includegraphics[width=.33\linewidth]{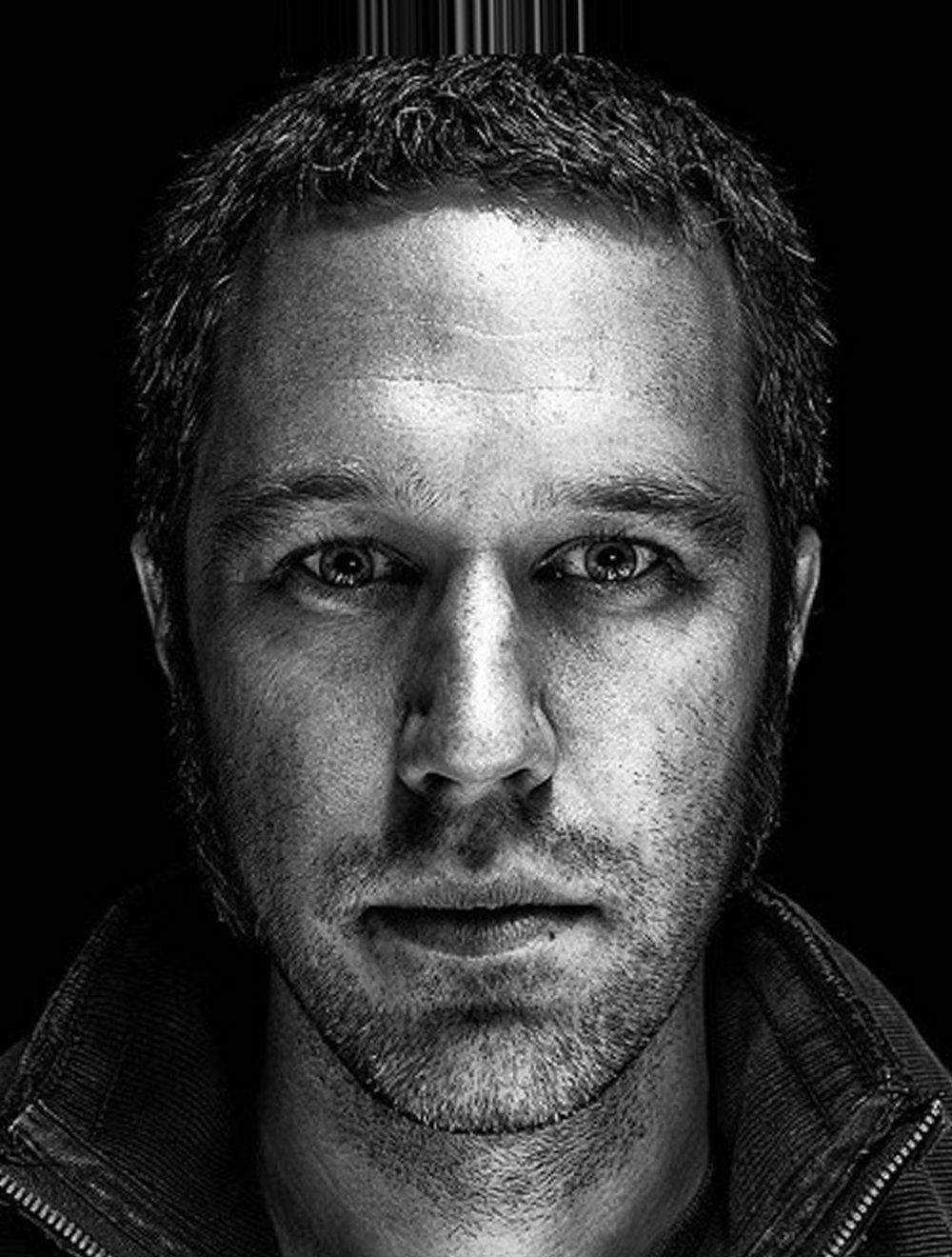}\\
\includegraphics[width=.33\linewidth]{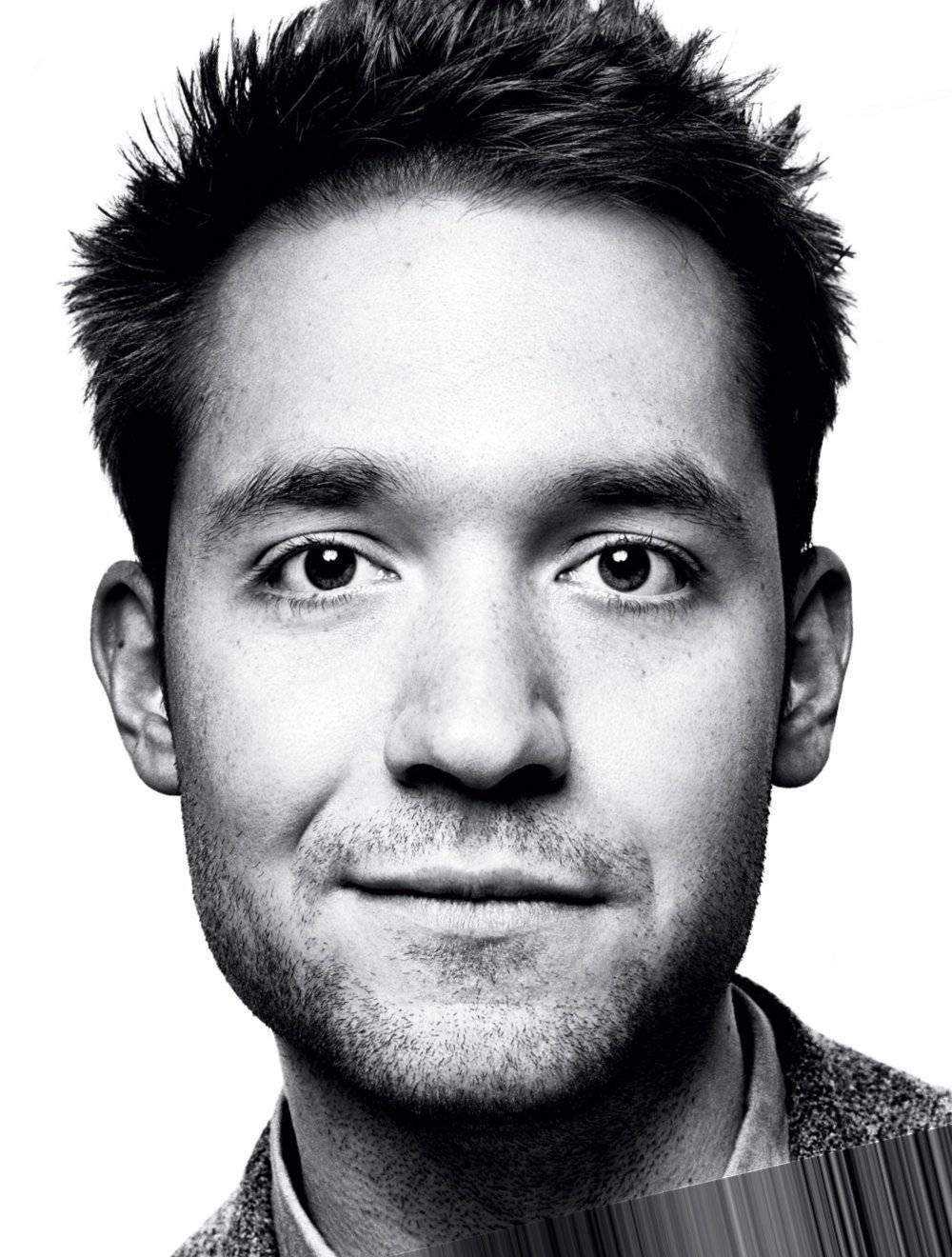}&
\includegraphics[width=.33\linewidth]{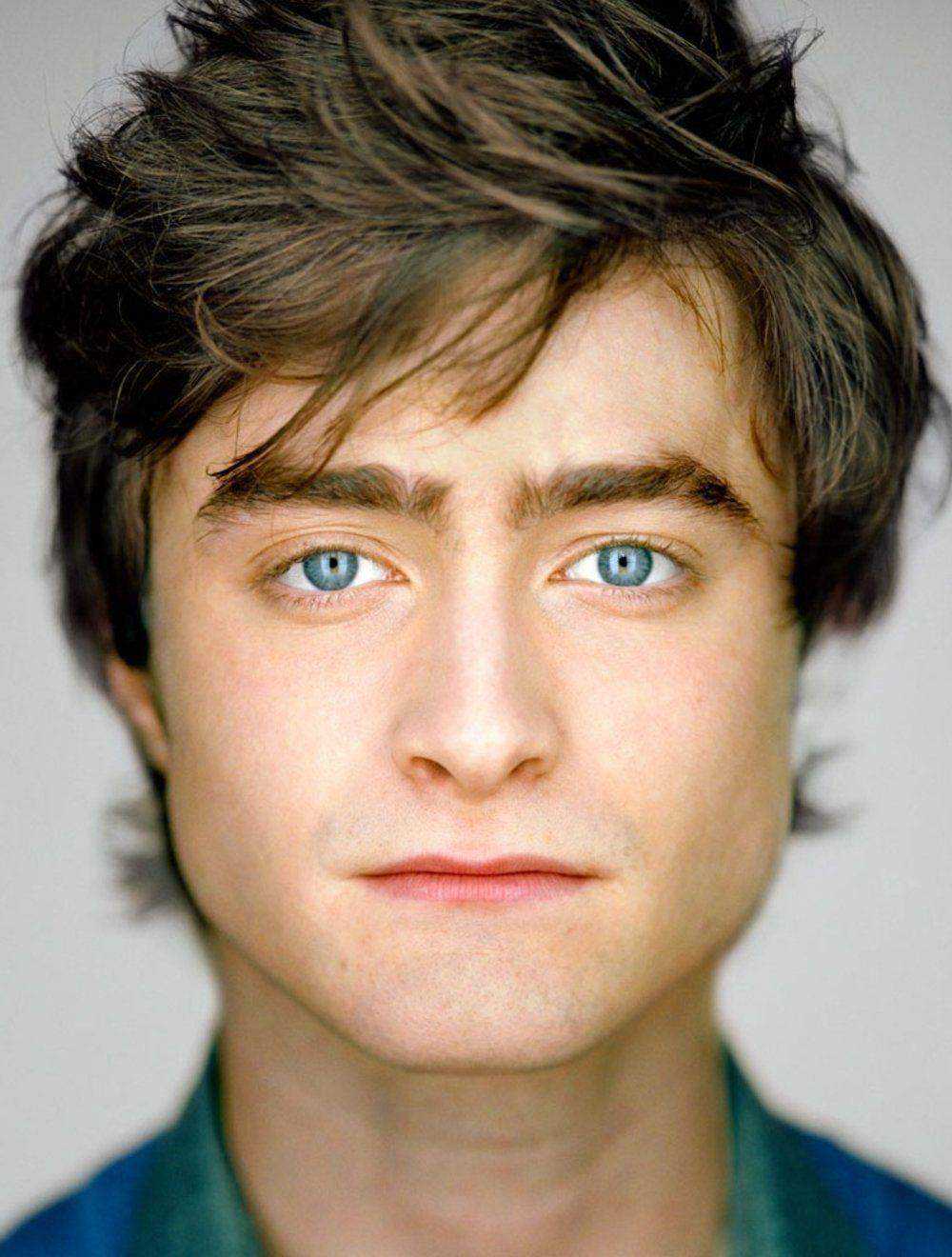}&
\includegraphics[width=.33\linewidth]{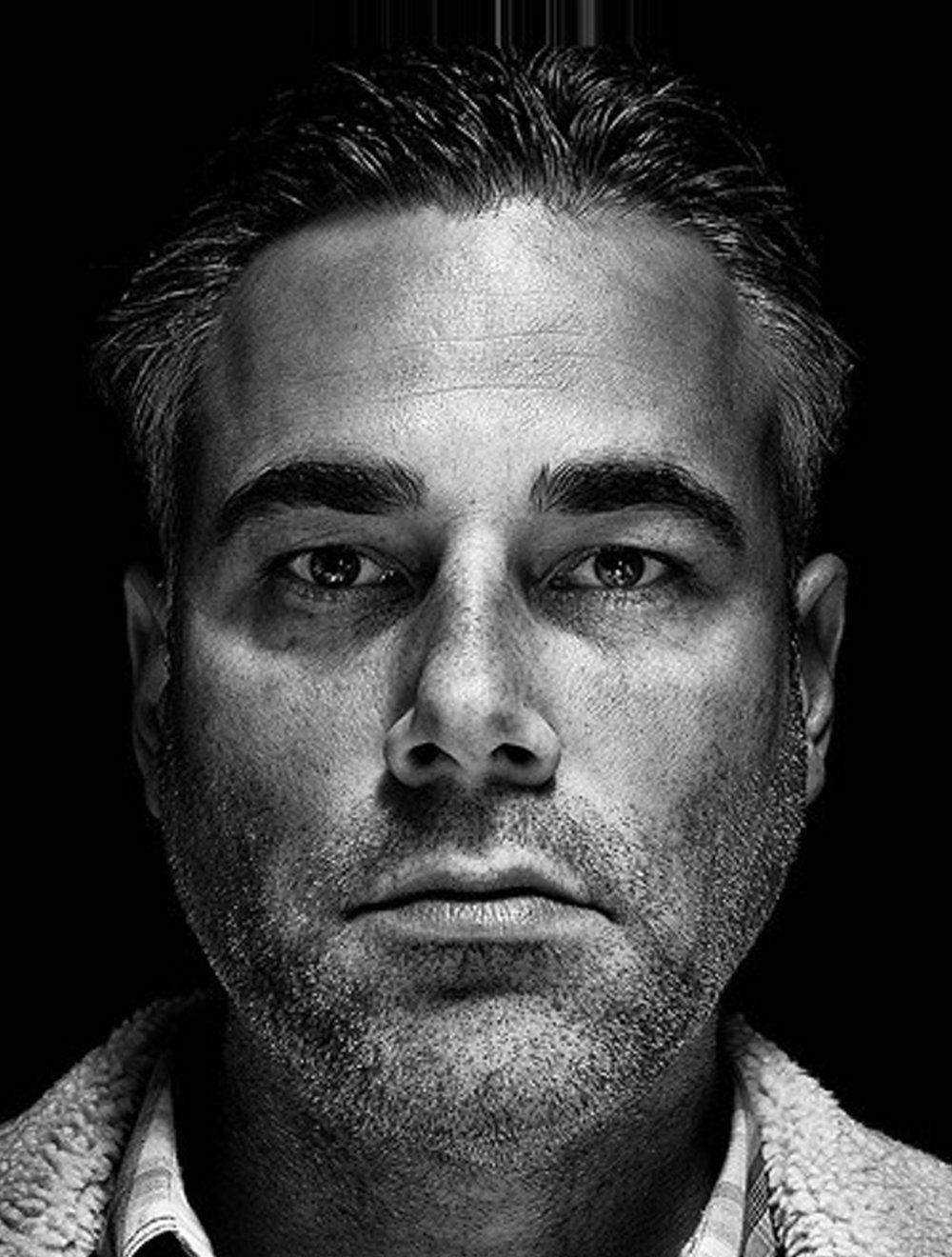}\\
(a) Platon& (b) Martin& (c) Kelco\\
\end{tabular}
\end{center}
\caption{Exemplars from the Platon, Martin and Kelco collections.
The face photos are captured with distinct styles.}
\label{fig:style}
\end{figure}

\begin{figure*}[!ht]
\begin{center}
\begin{tabular}{cccc}
\includegraphics[width=.245\linewidth]{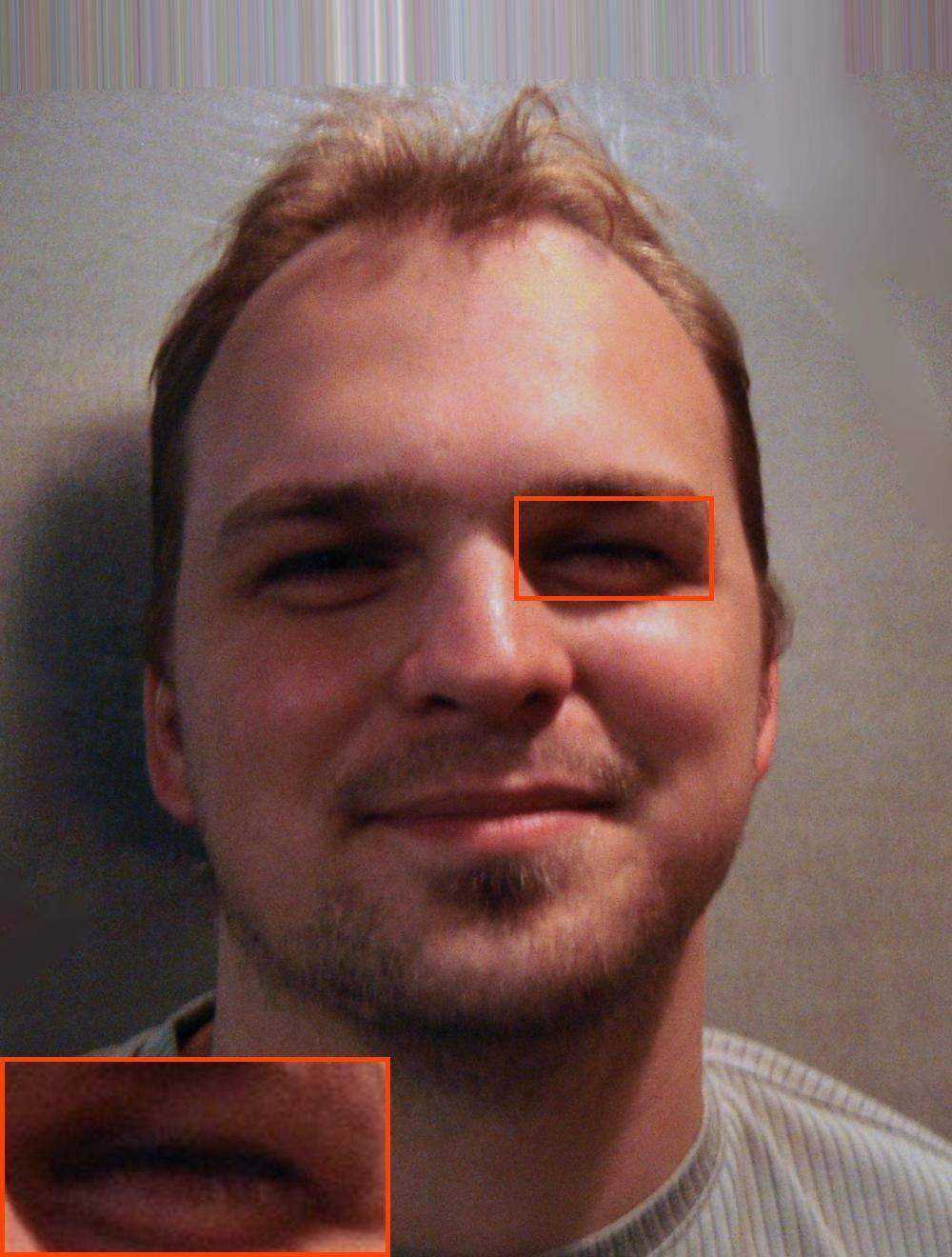}&
\includegraphics[width=.245\linewidth]{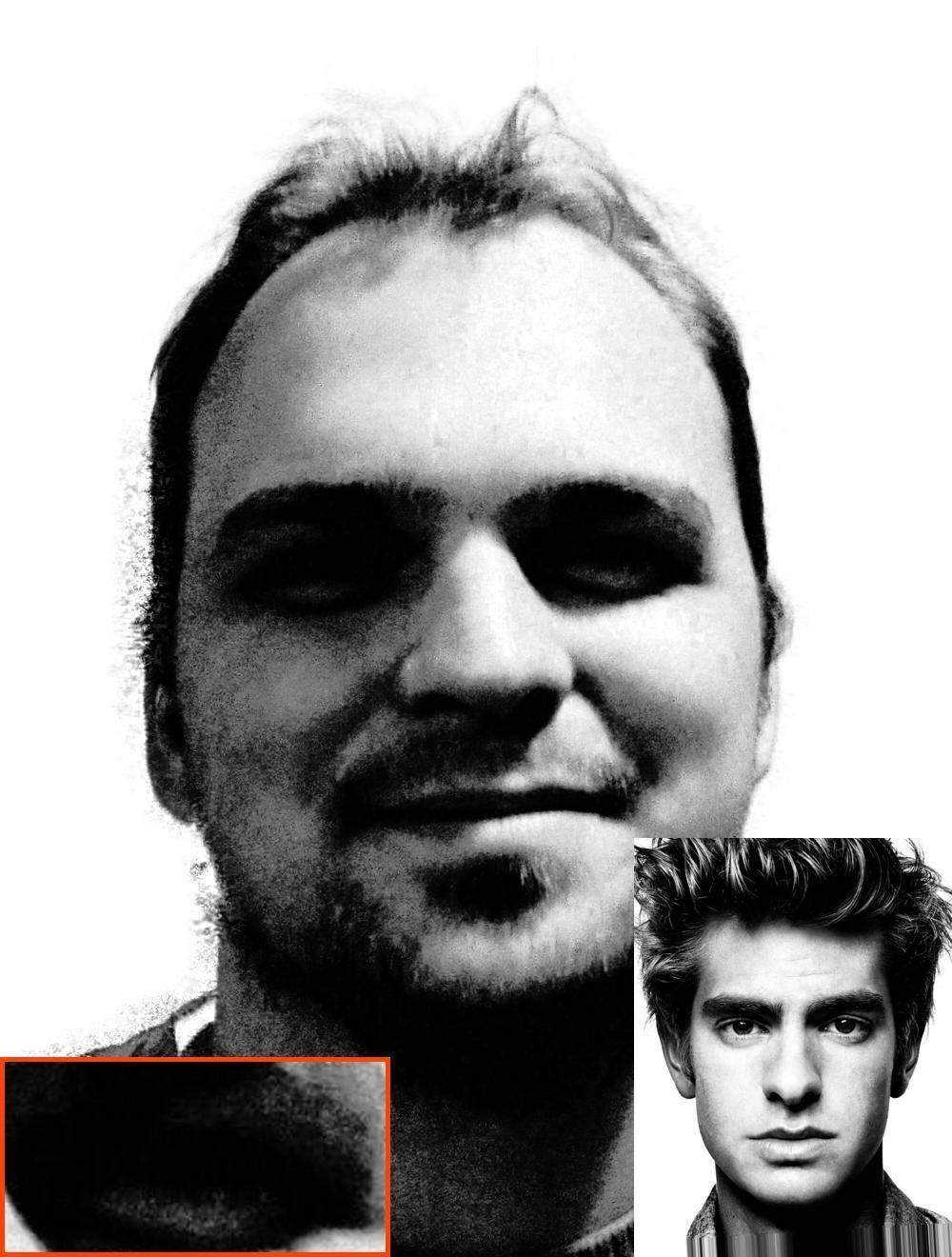}&
\includegraphics[width=.245\linewidth]{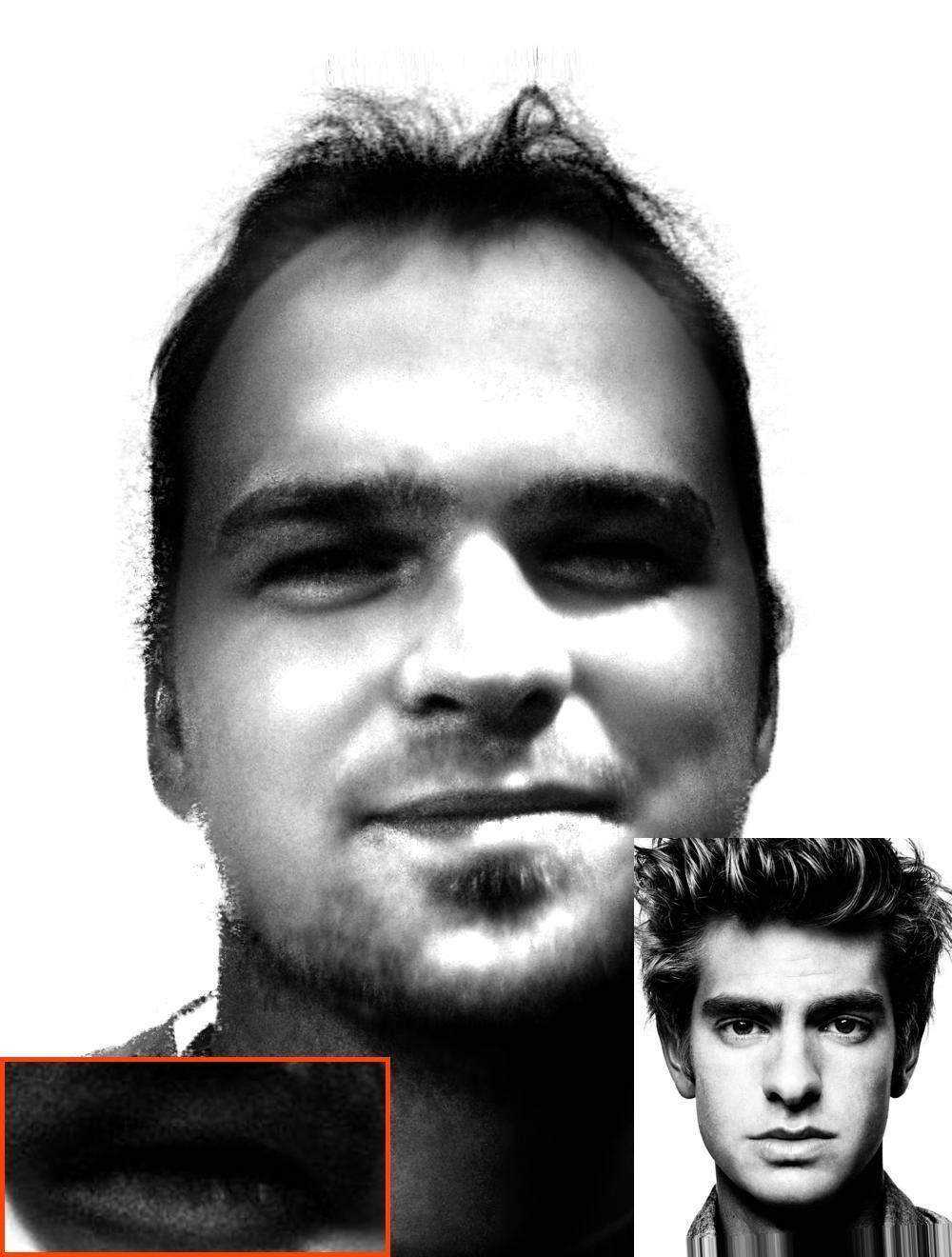}&
\includegraphics[width=.245\linewidth]{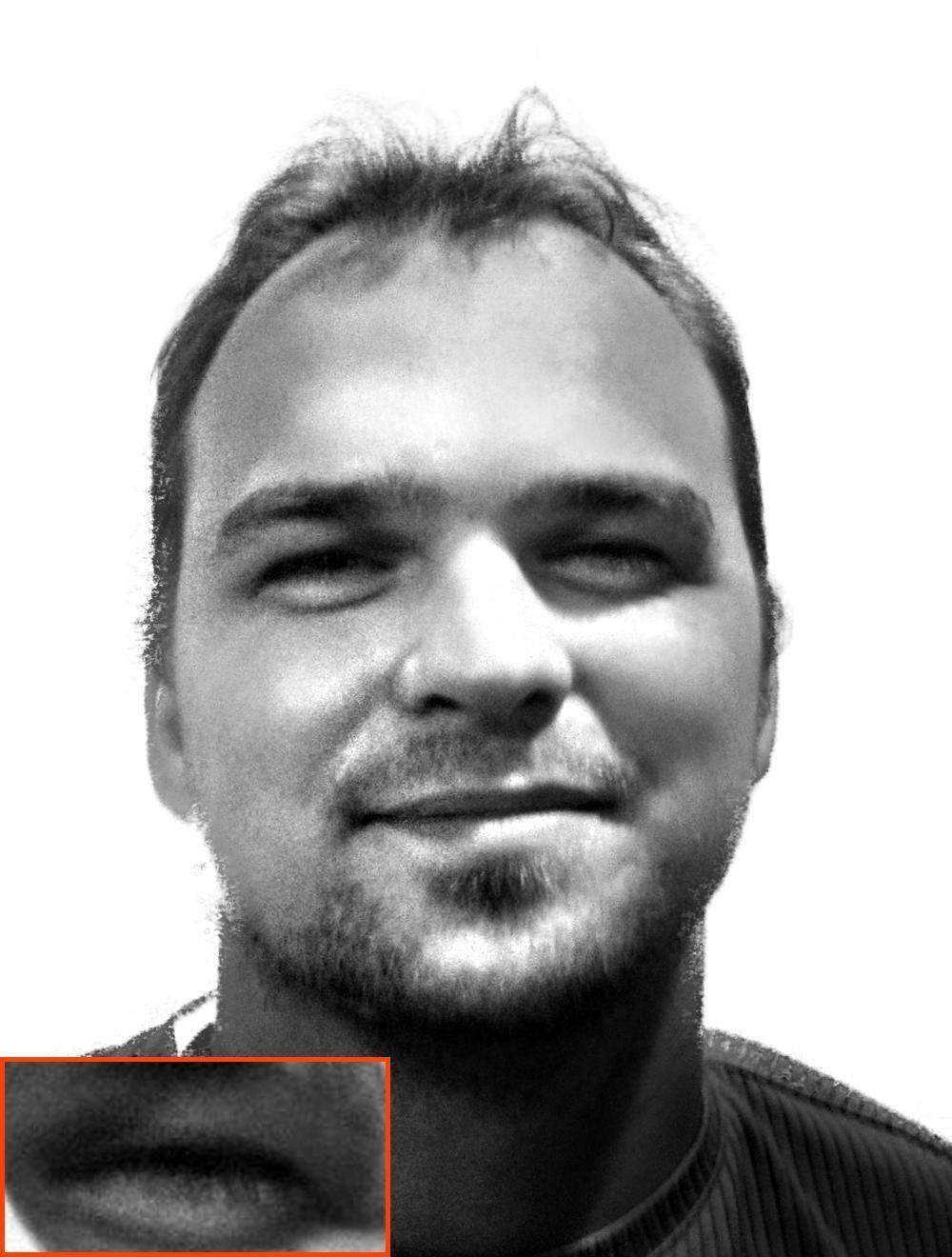}\\
\includegraphics[width=.245\linewidth]{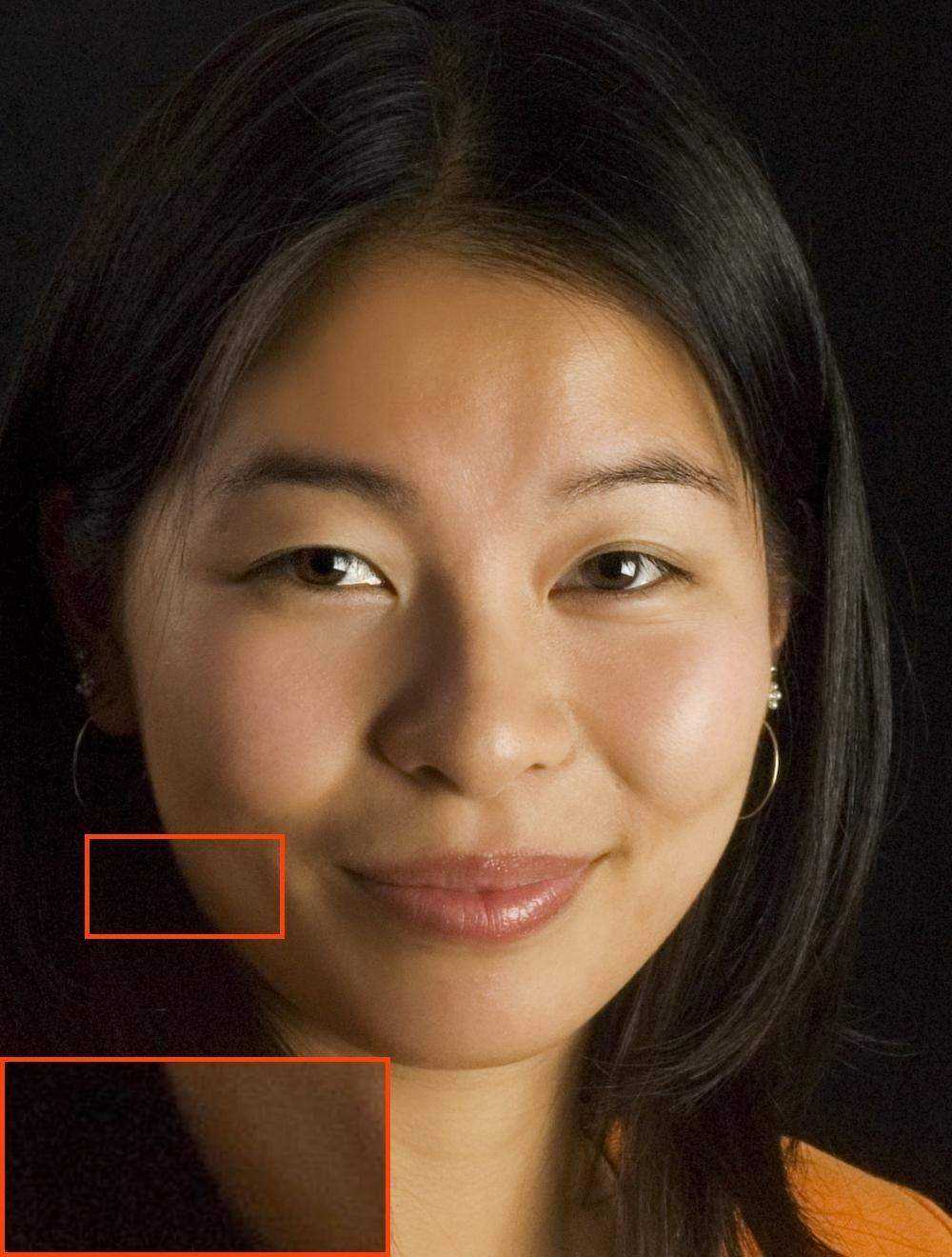}&
\includegraphics[width=.245\linewidth]{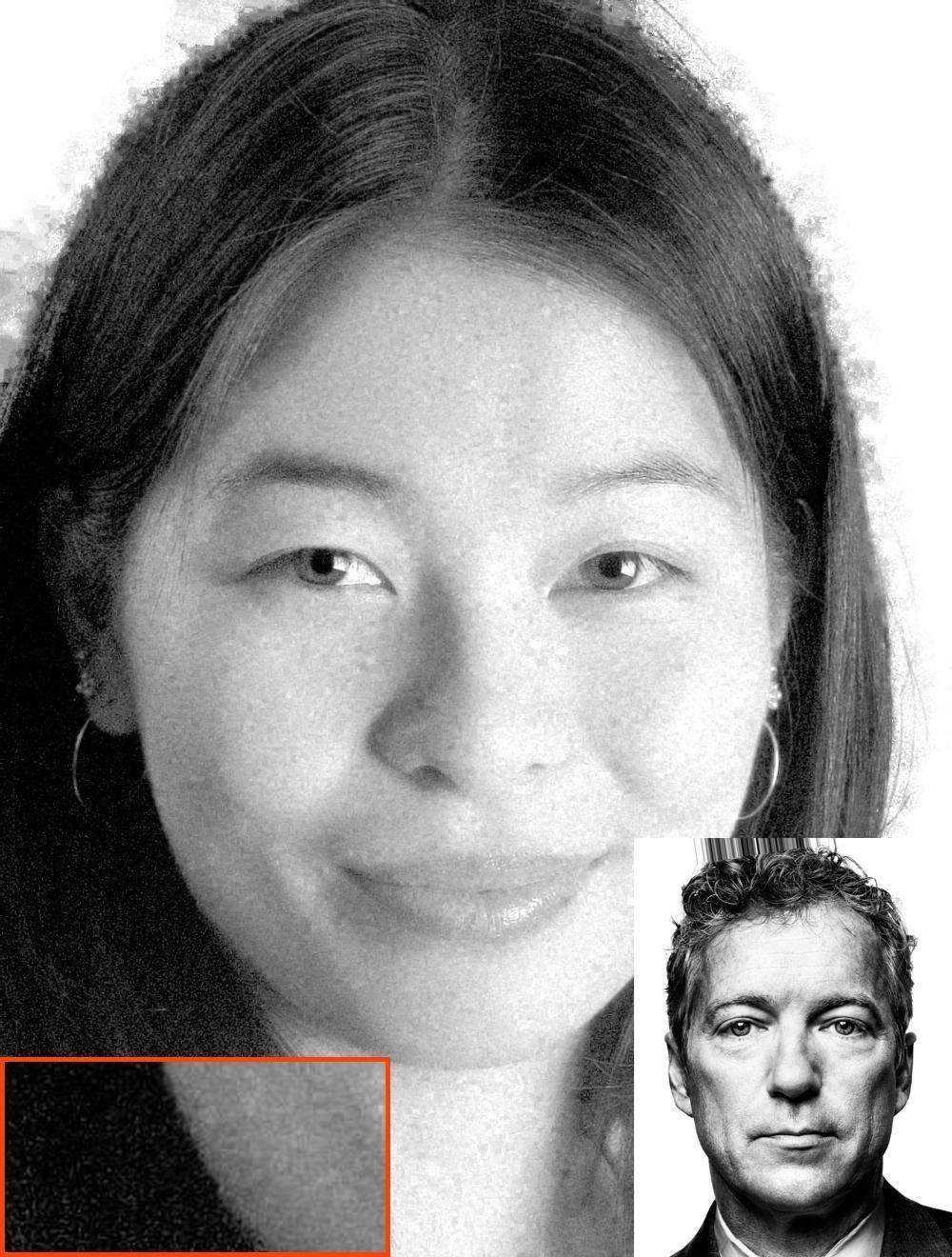}&
\includegraphics[width=.245\linewidth]{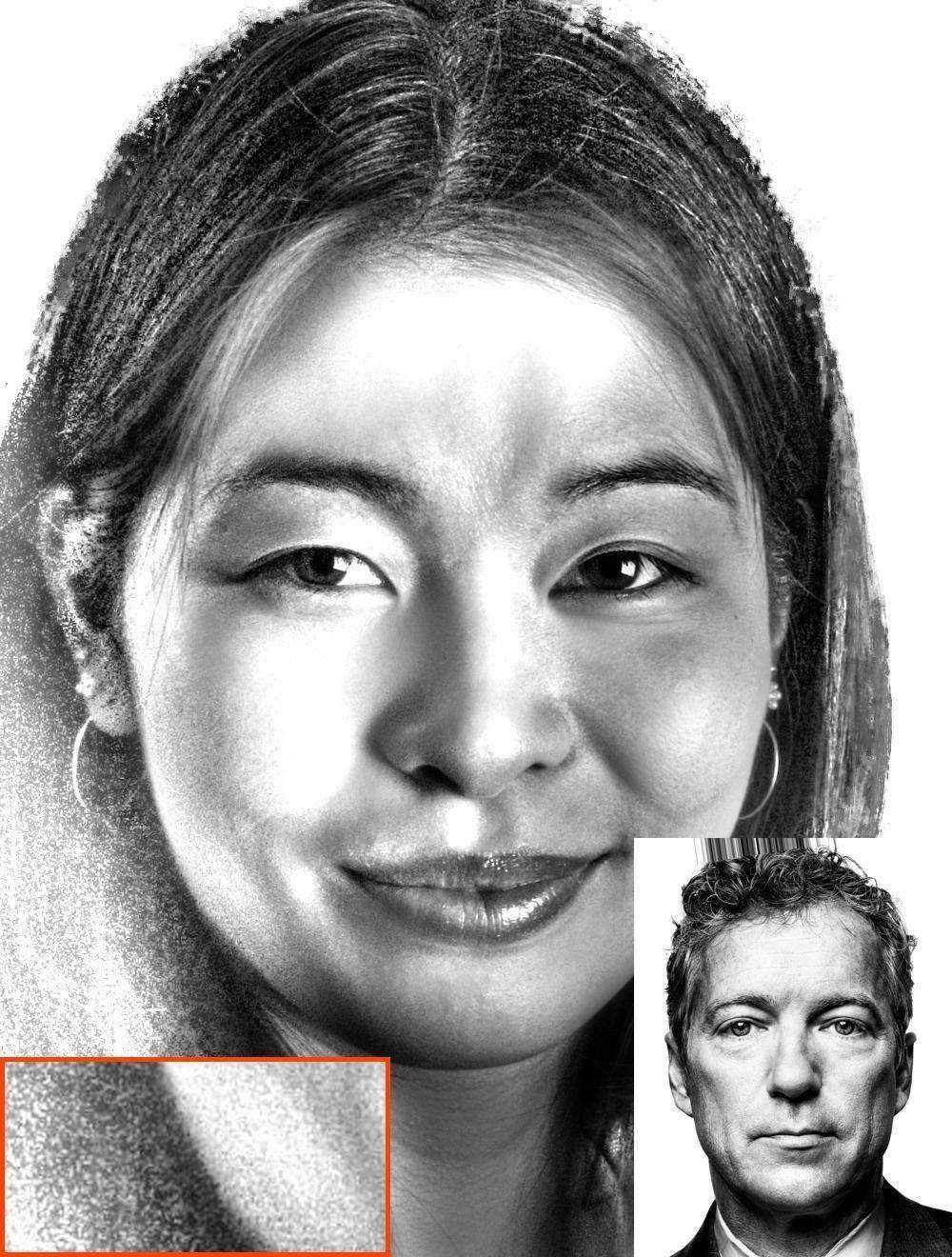}&
\includegraphics[width=.245\linewidth]{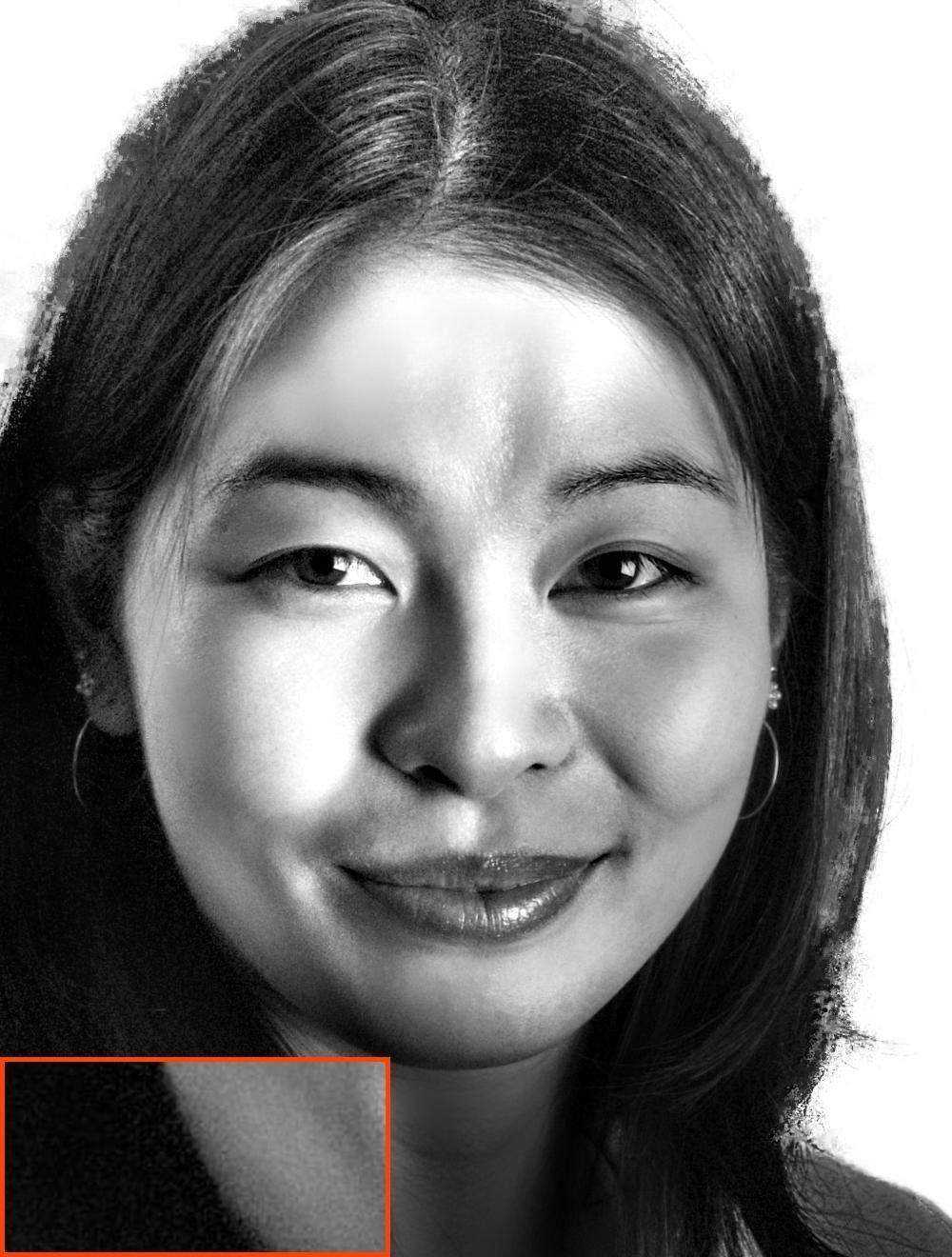}\\
\includegraphics[width=.245\linewidth]{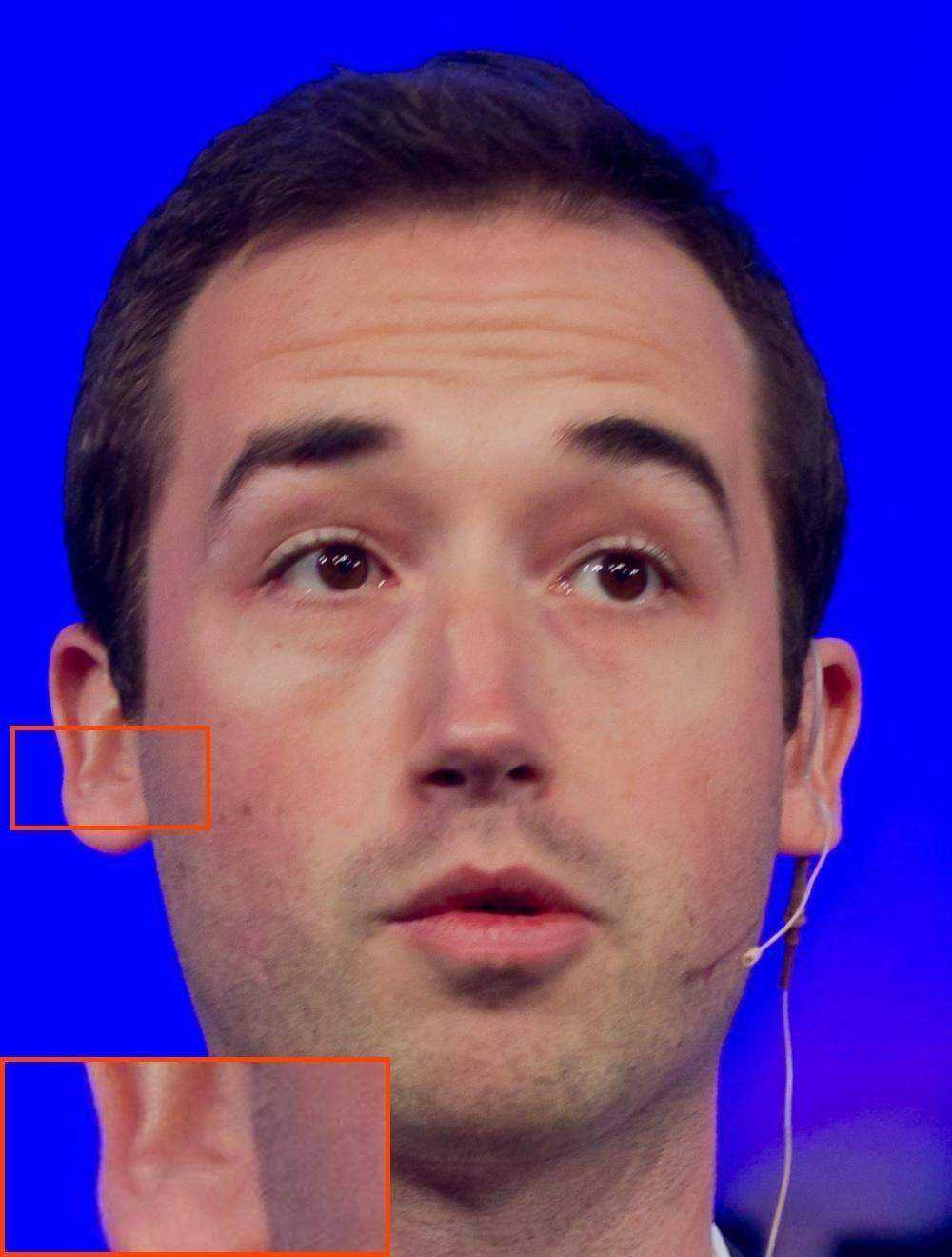}&
\includegraphics[width=.245\linewidth]{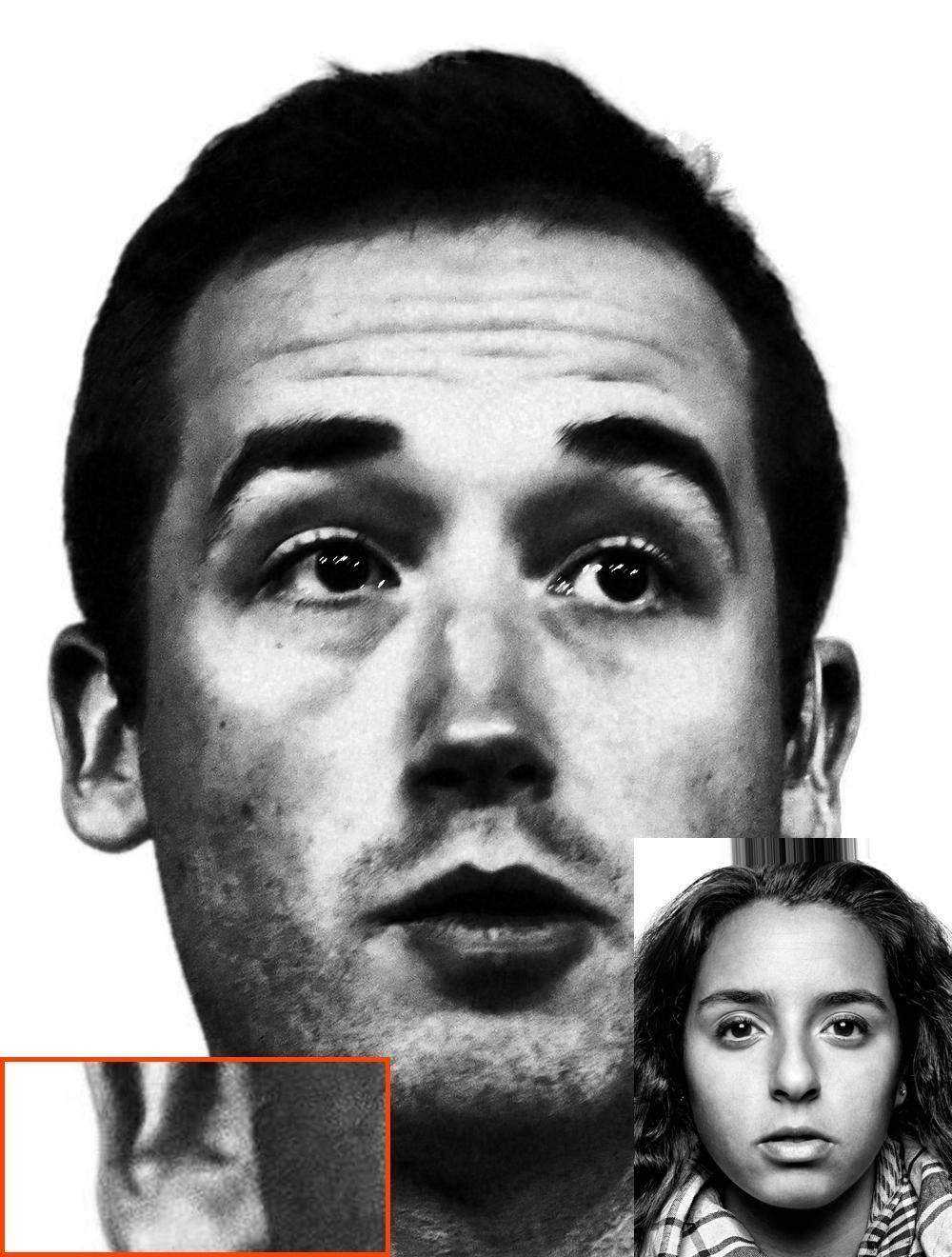}&
\includegraphics[width=.245\linewidth]{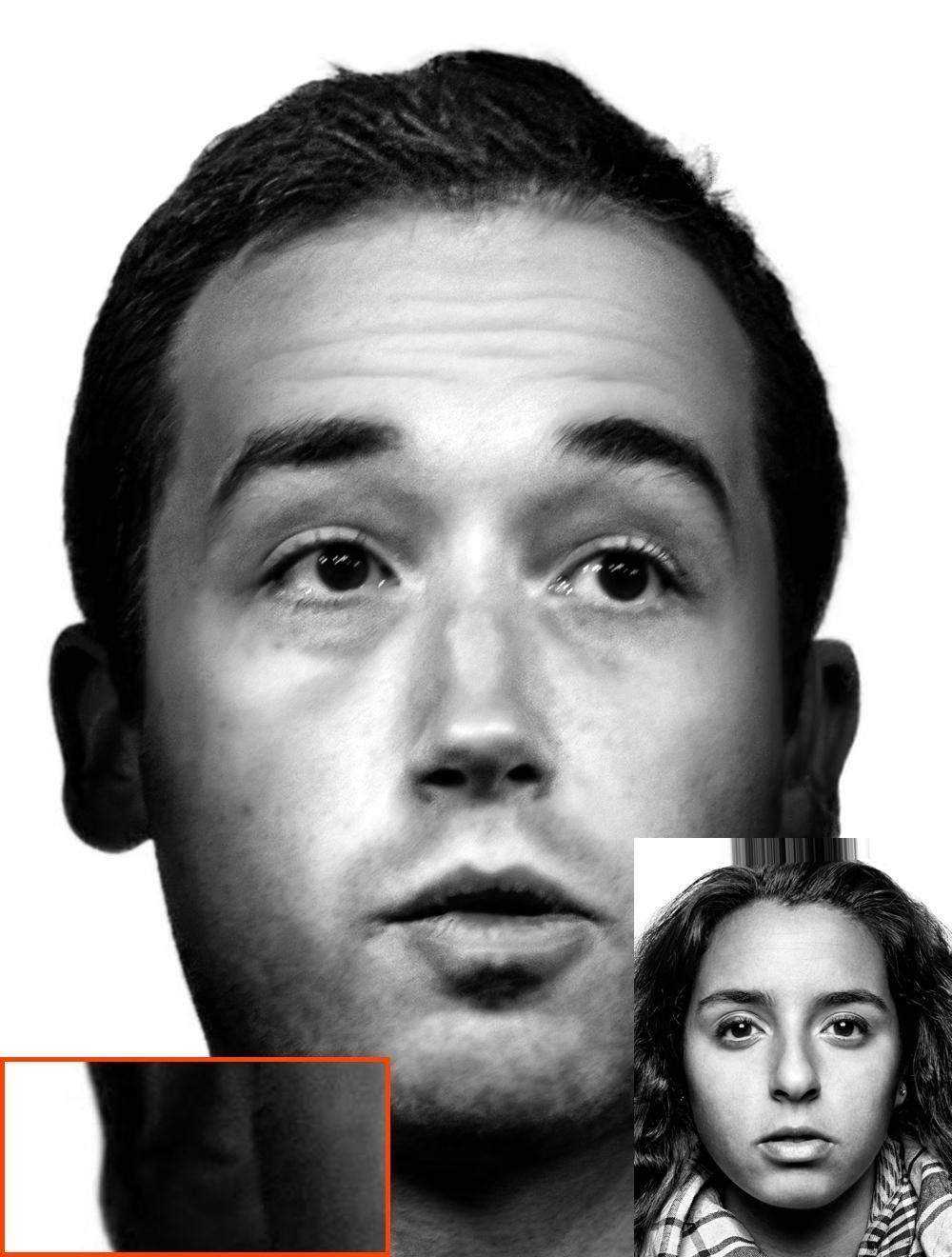}&
\includegraphics[width=.245\linewidth]{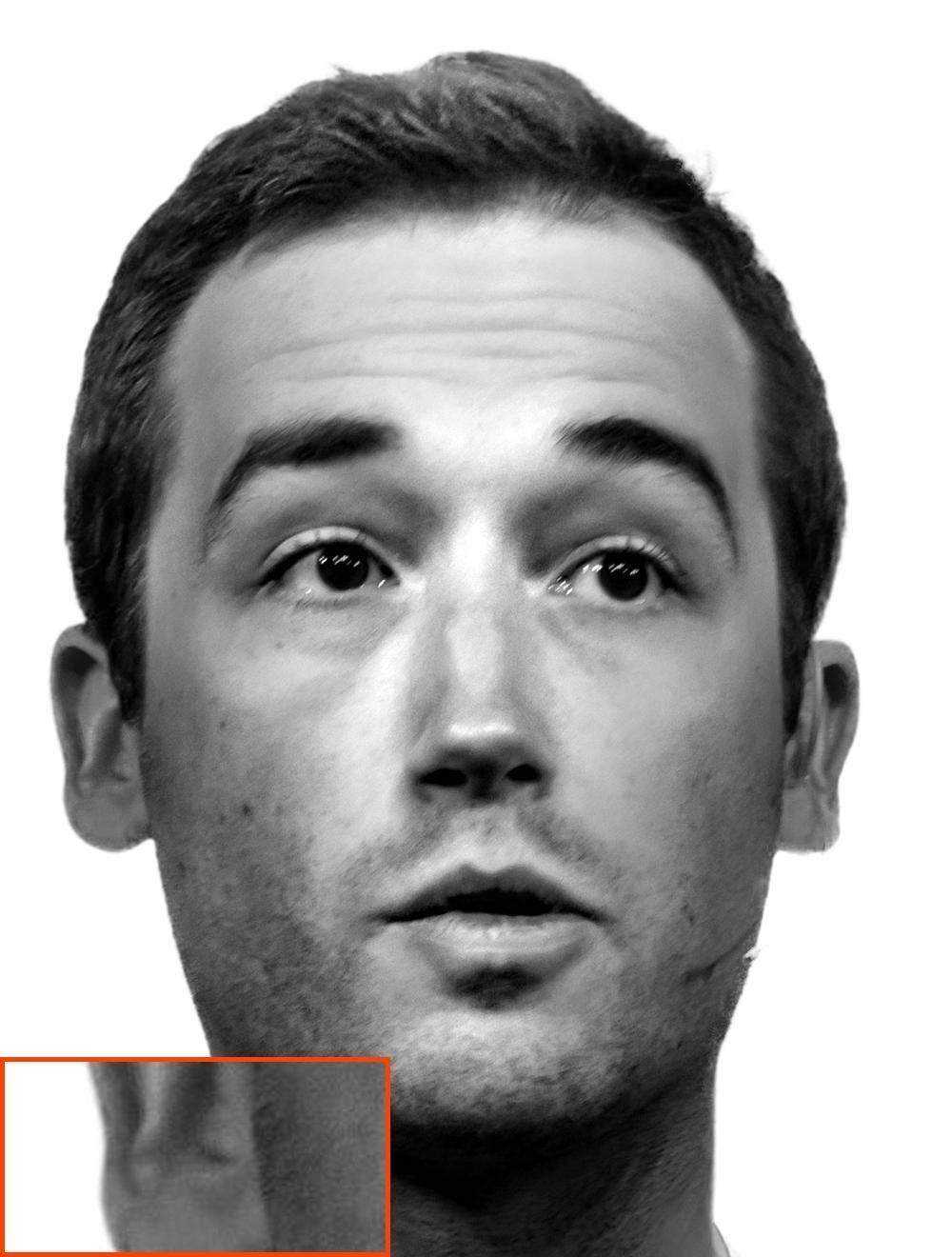}\\
(a) Input photo& (b) Holistic \cite{Sunkavalli-siggraph10-Harmonization}&
(c) Local \cite{Shih-siggraph14-StyleTransfer}& (d) Proposed\\
\end{tabular}
\end{center}
\caption{Qualitative evaluation on the Platon dataset. The proposed
  method performs favorably against holistic and local methods.
These images can be better visualized with zoom-in to analyze the details.}
\label{fig:exp1}
\end{figure*}

\begin{figure*}[!ht]
\begin{center}
\begin{tabular}{cccc}
\includegraphics[width=.245\linewidth]{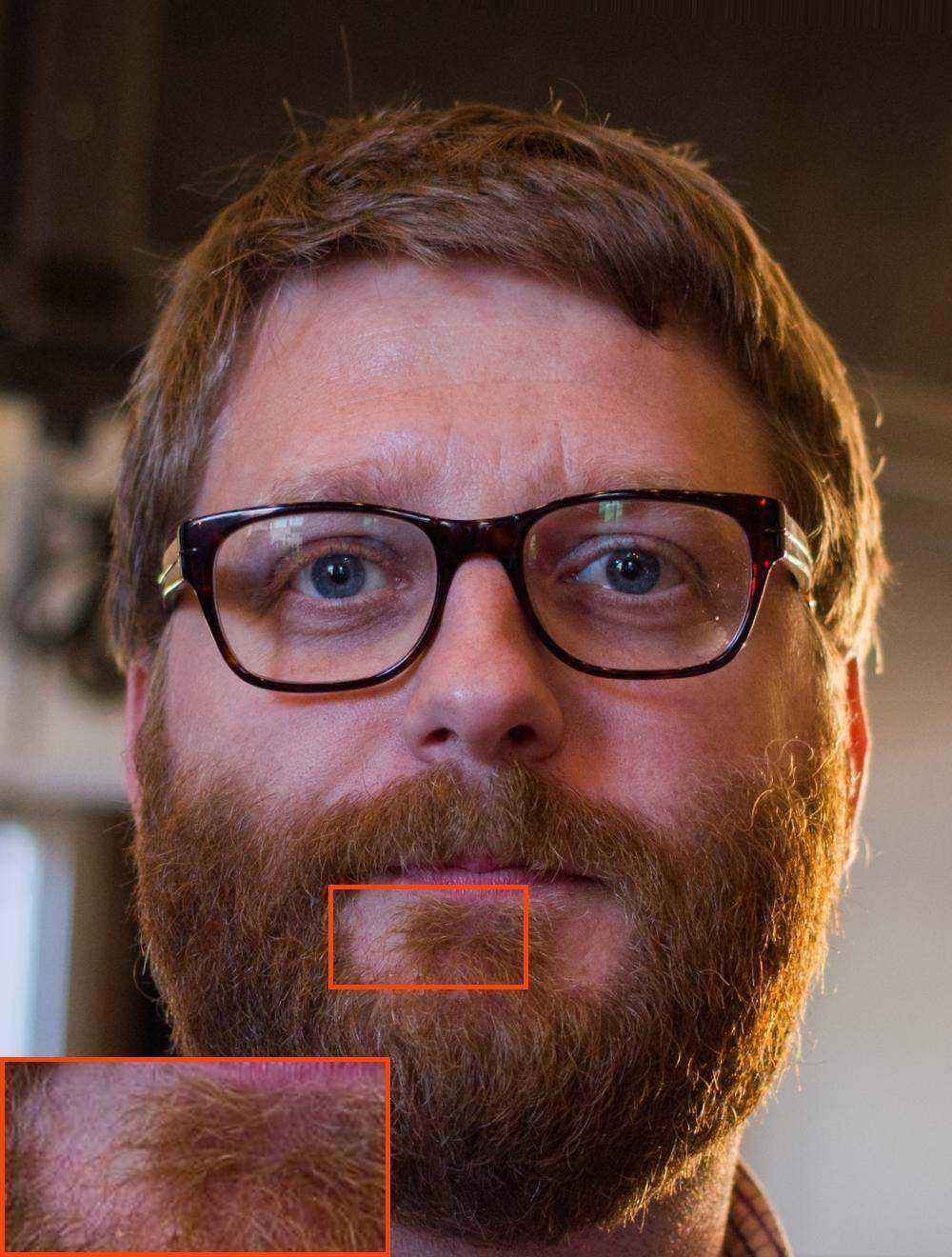}&
\includegraphics[width=.245\linewidth]{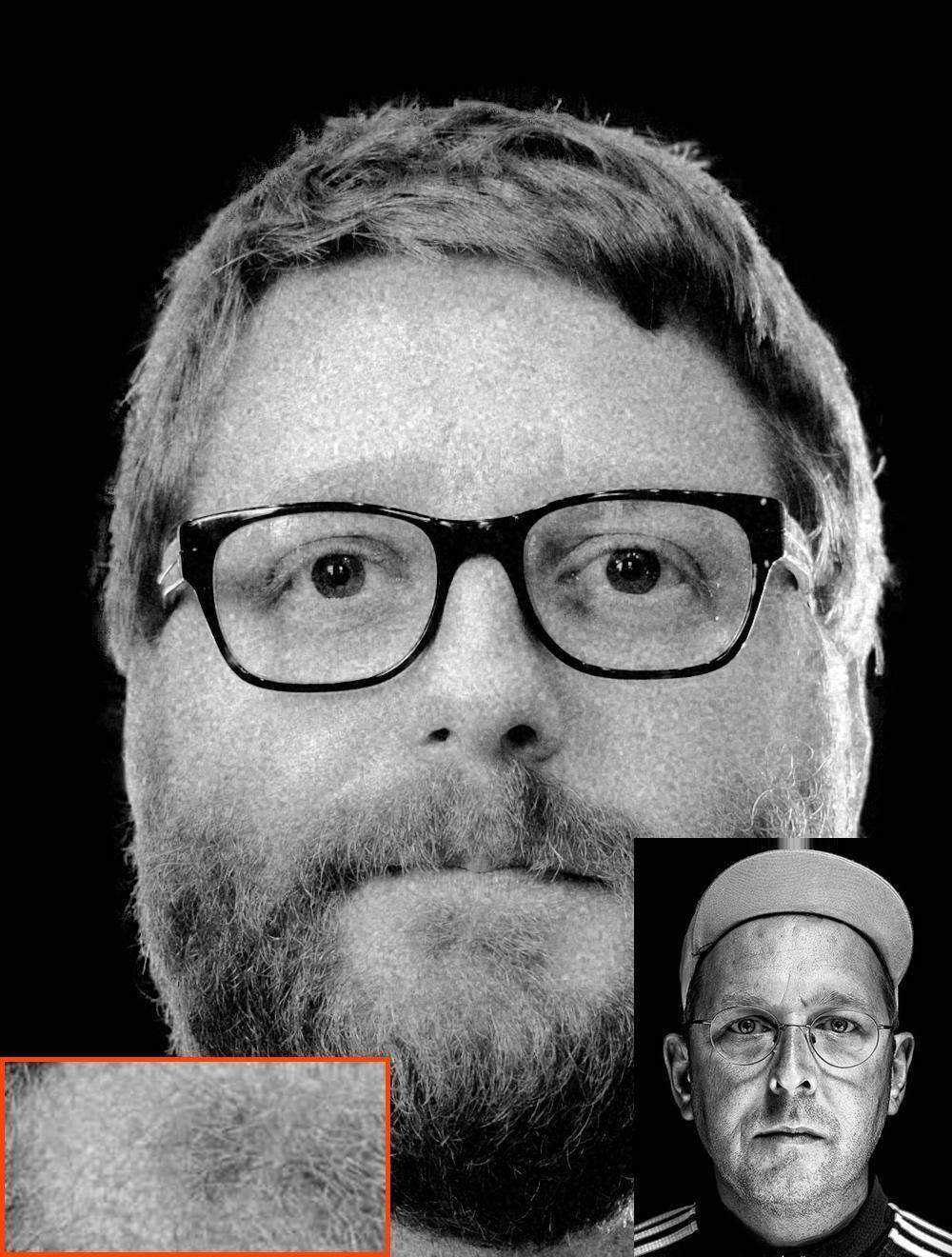}&
\includegraphics[width=.245\linewidth]{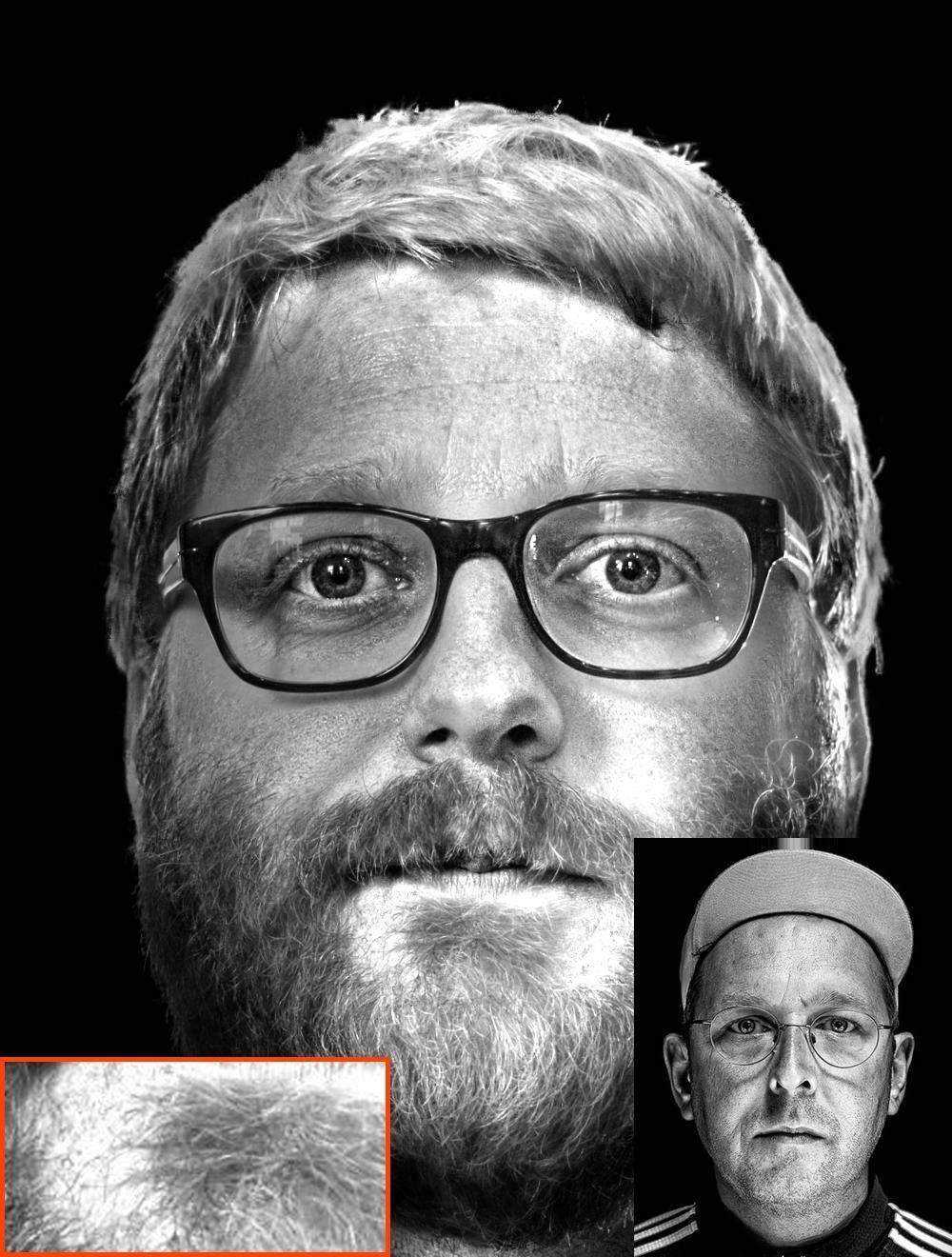}&
\includegraphics[width=.245\linewidth]{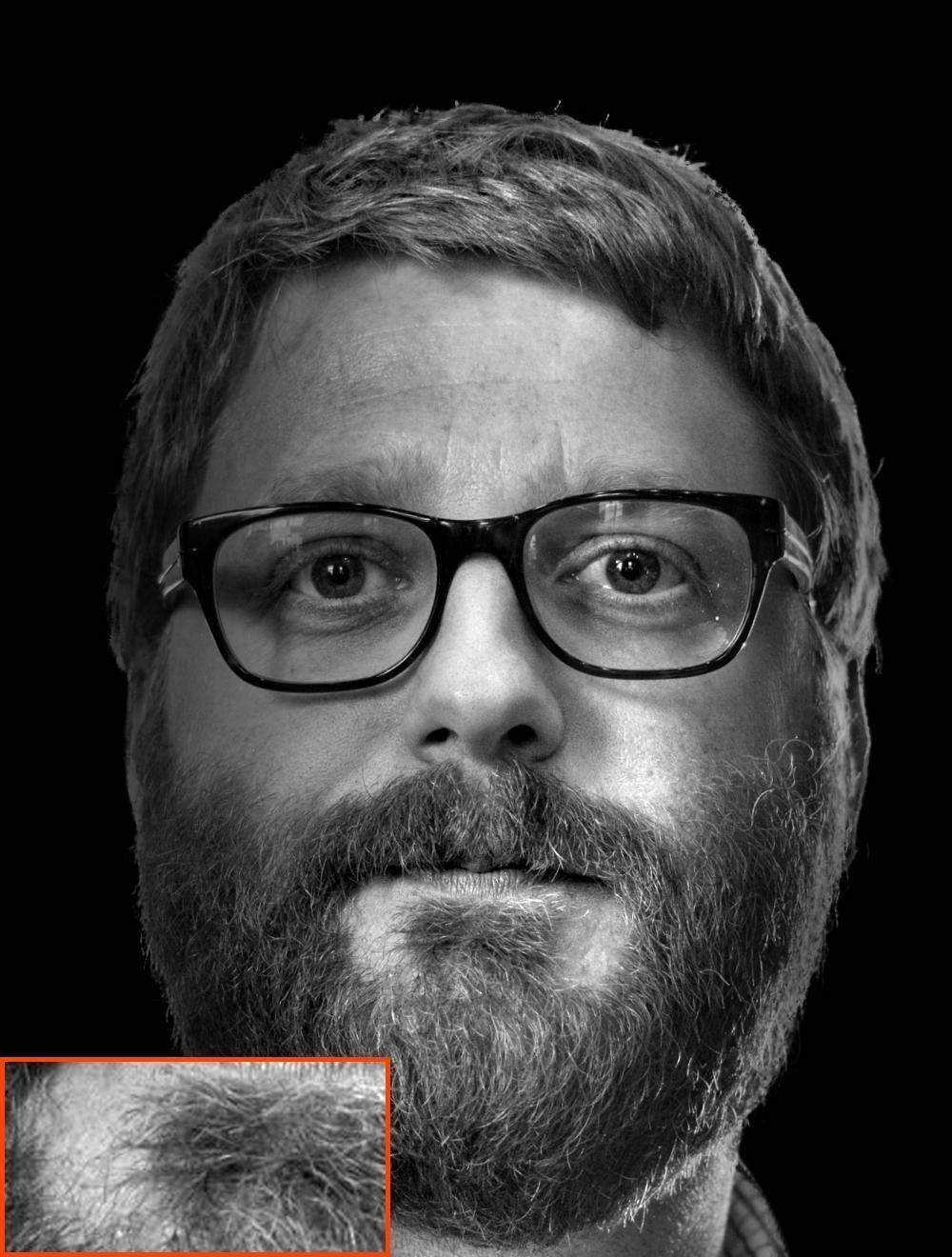}\\
\includegraphics[width=.245\linewidth]{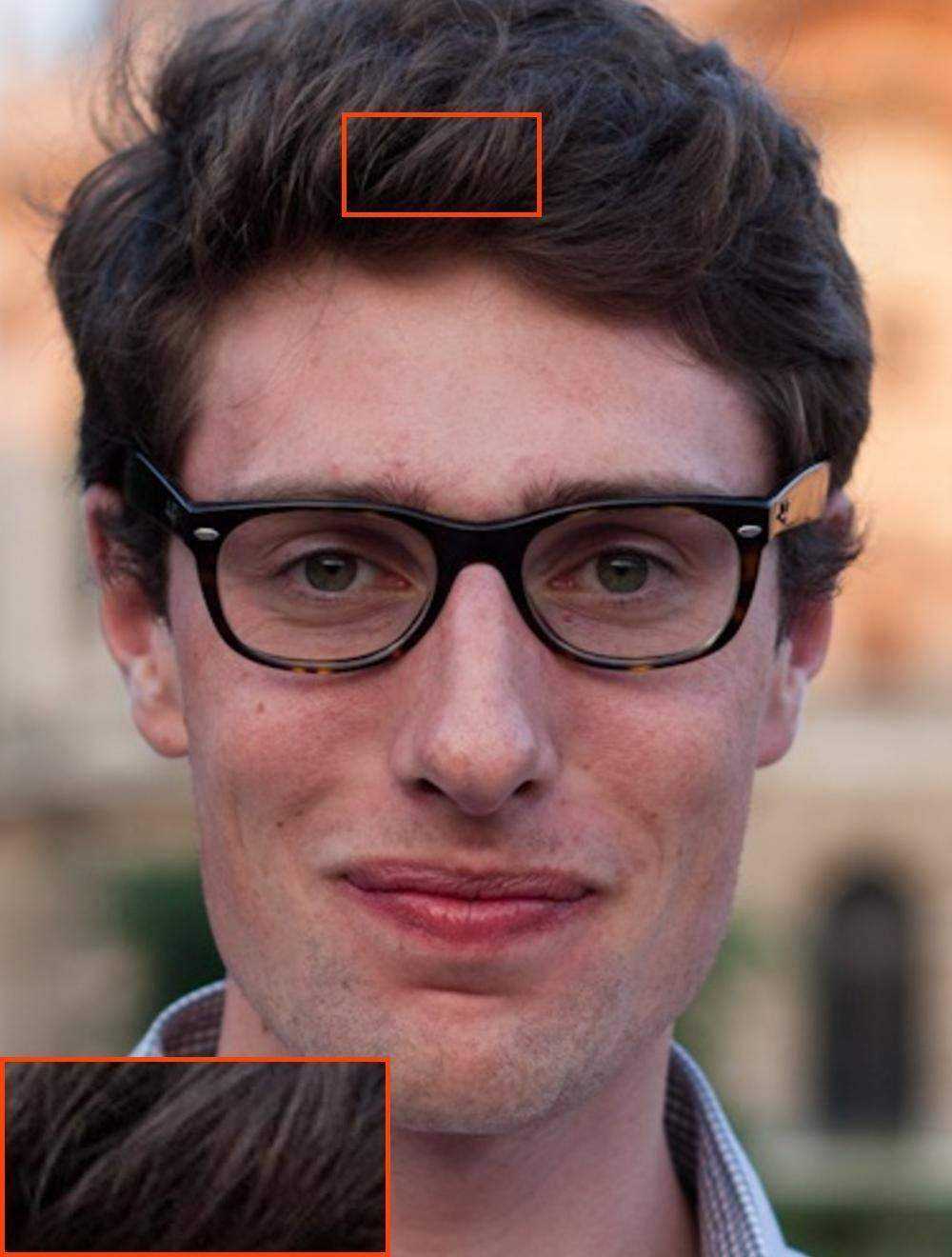}&
\includegraphics[width=.245\linewidth]{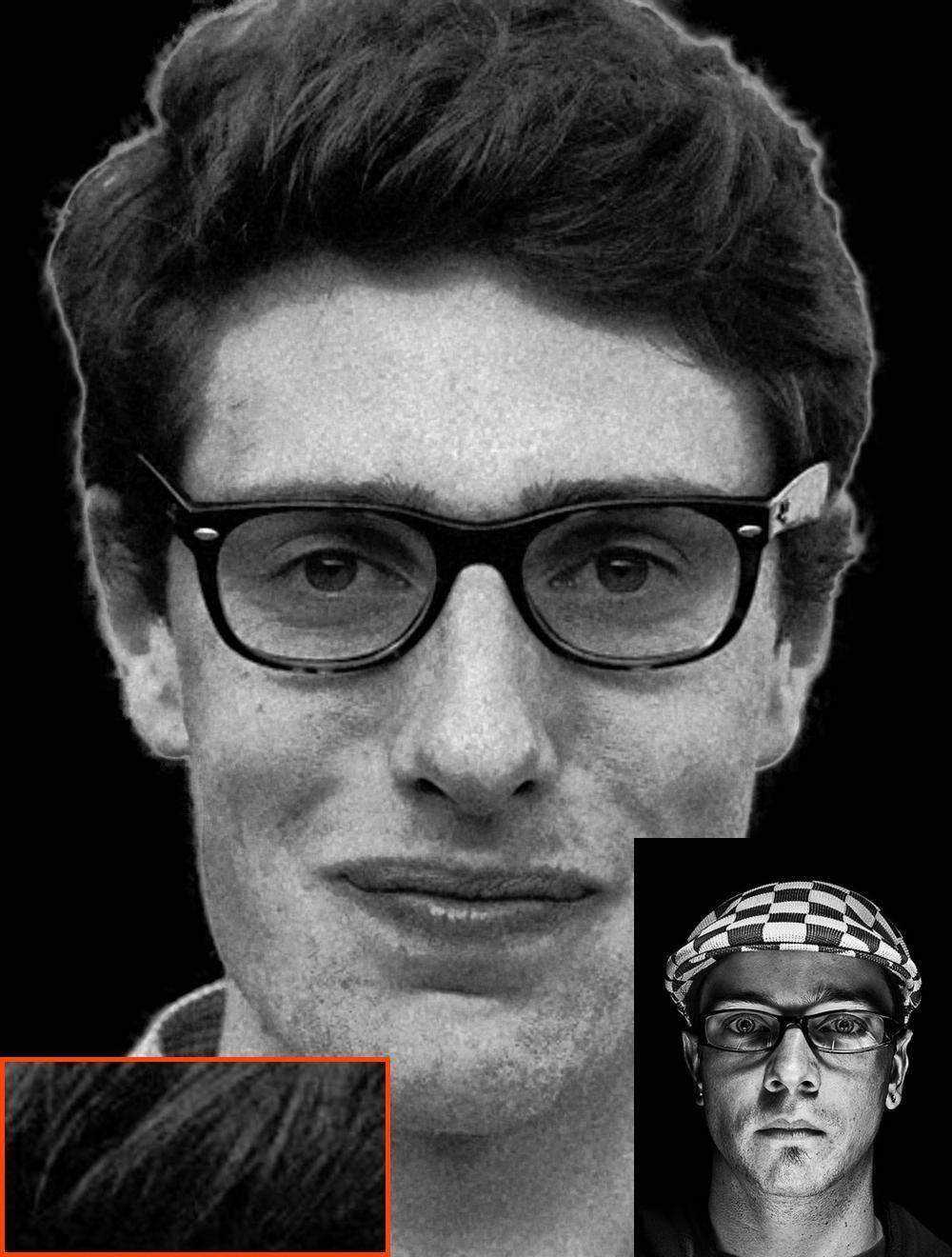}&
\includegraphics[width=.245\linewidth]{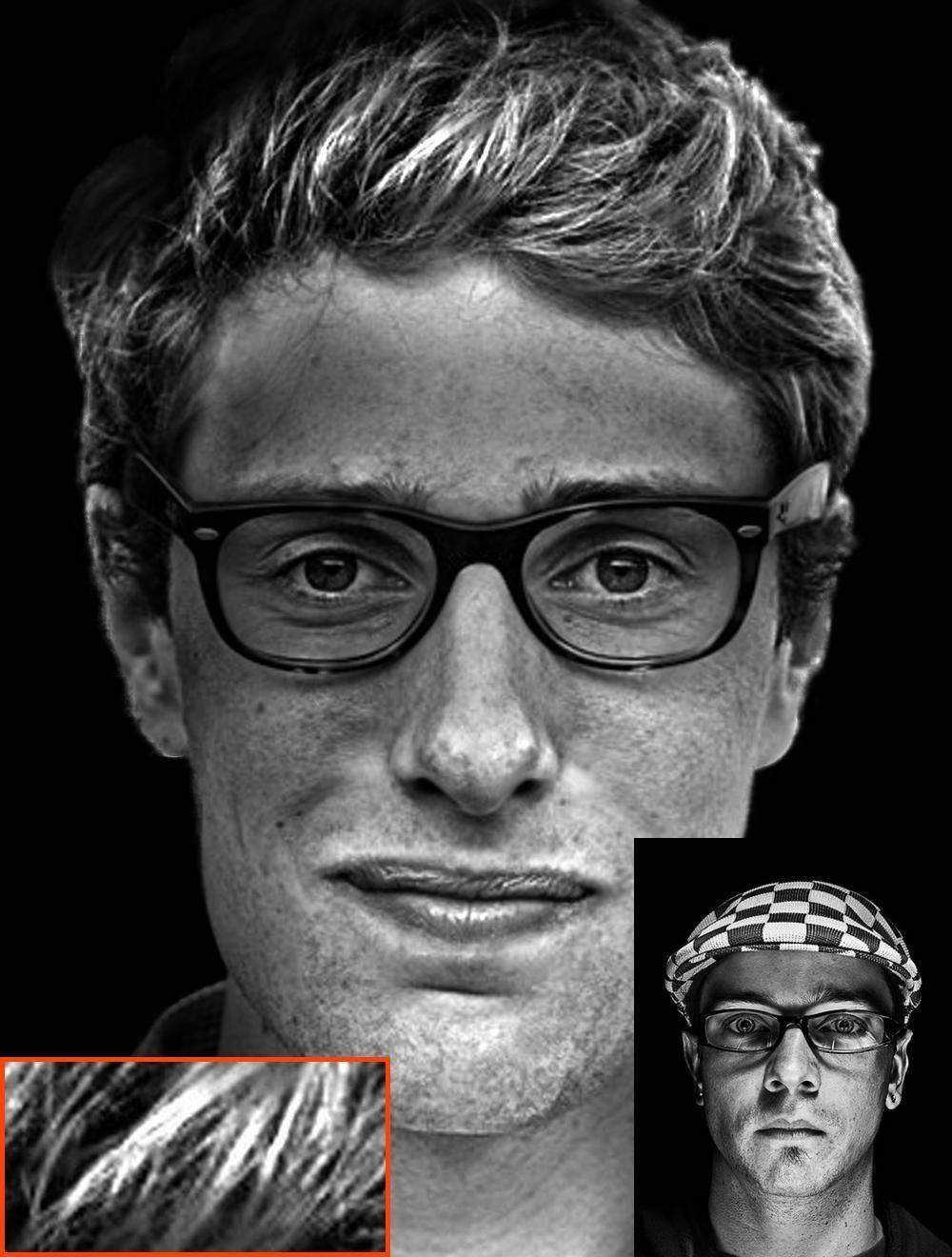}&
\includegraphics[width=.245\linewidth]{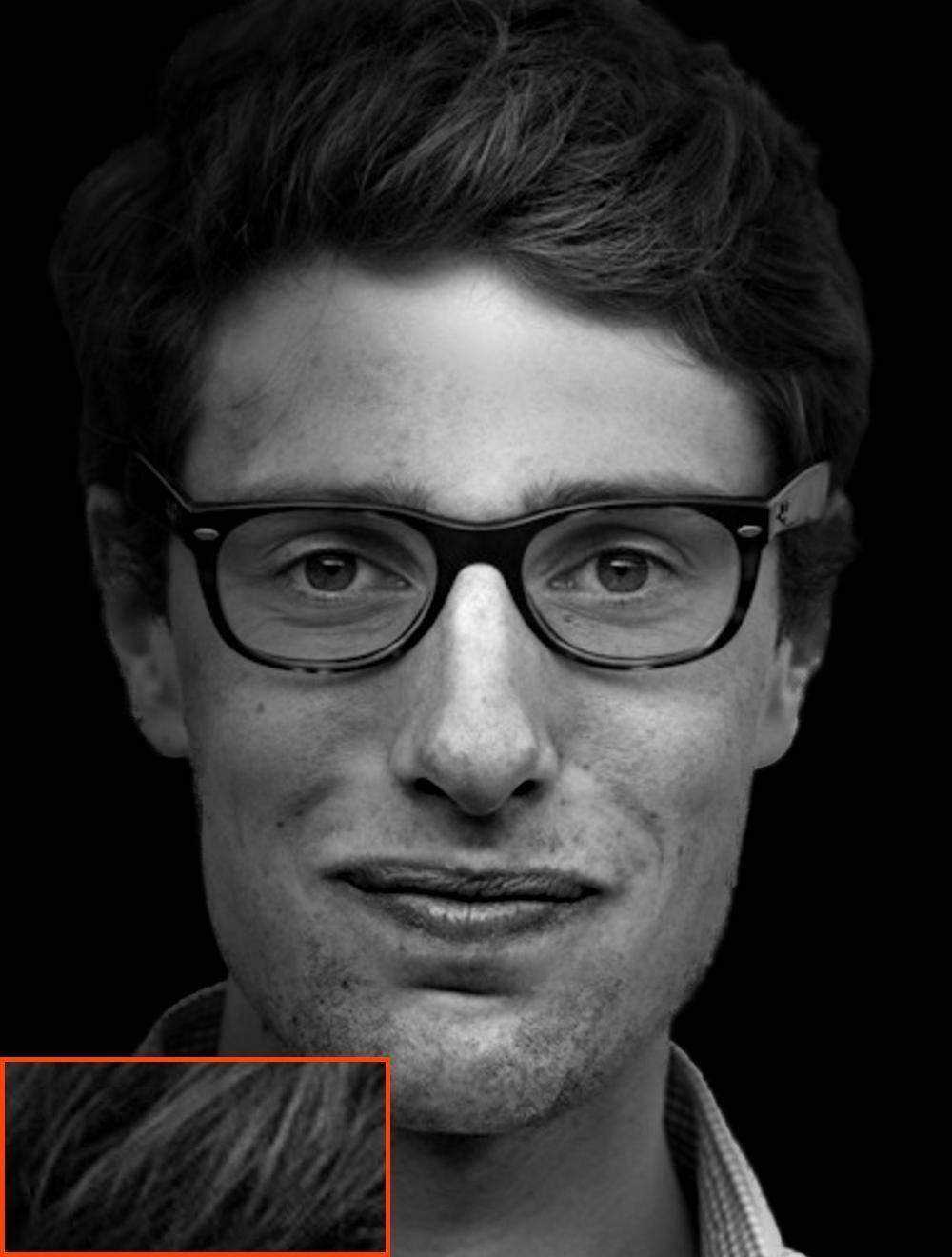}\\
\includegraphics[width=.245\linewidth]{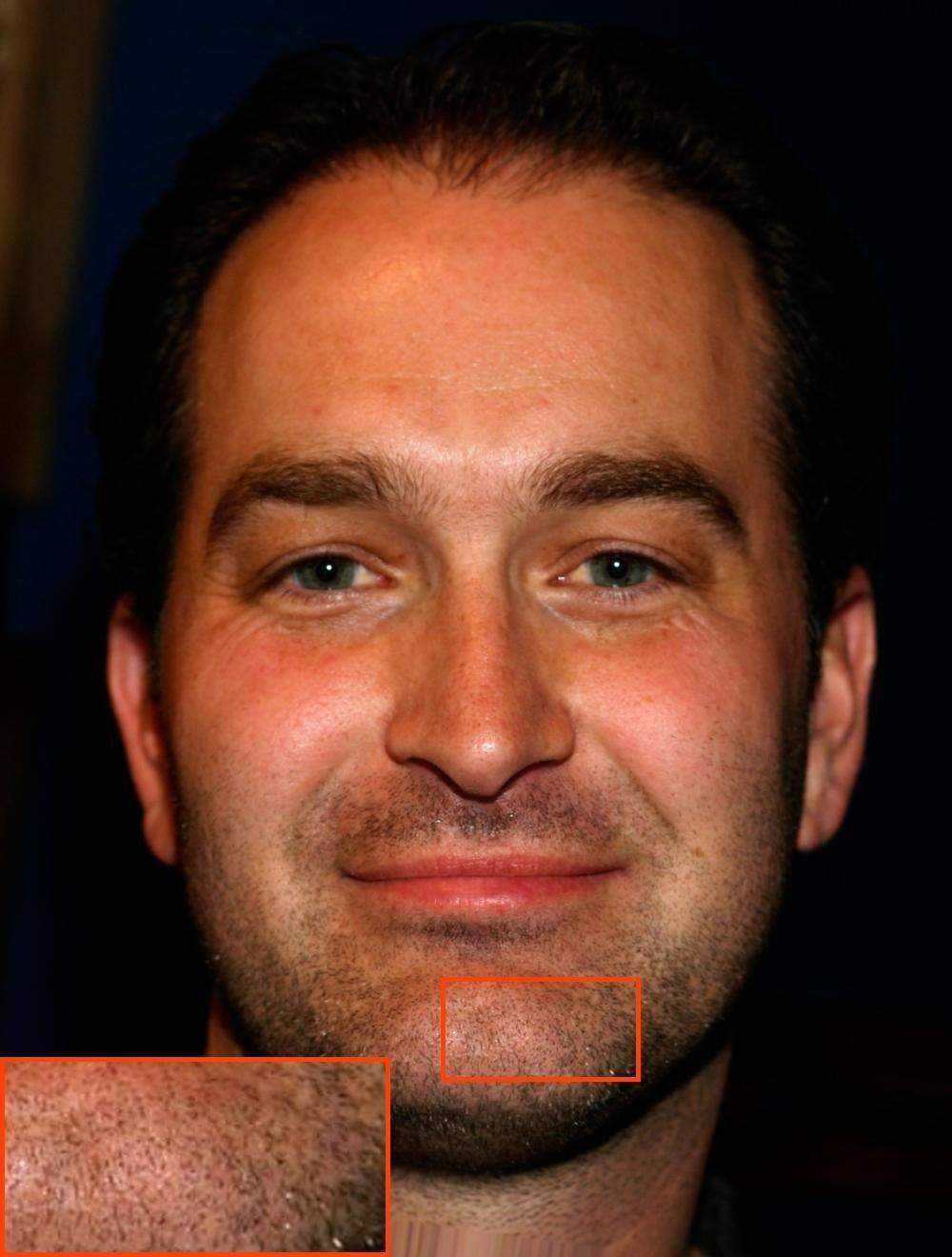}&
\includegraphics[width=.245\linewidth]{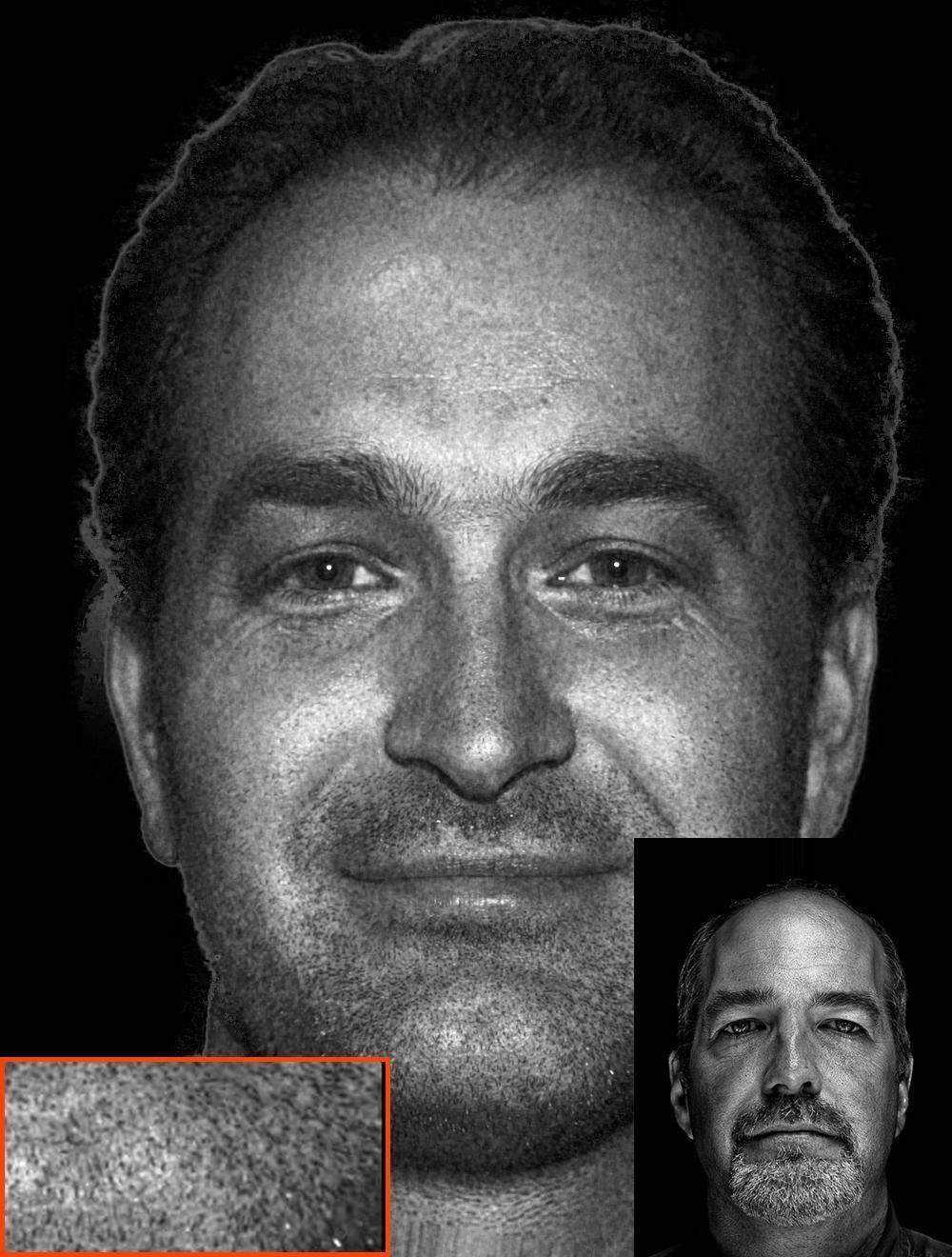}&
\includegraphics[width=.245\linewidth]{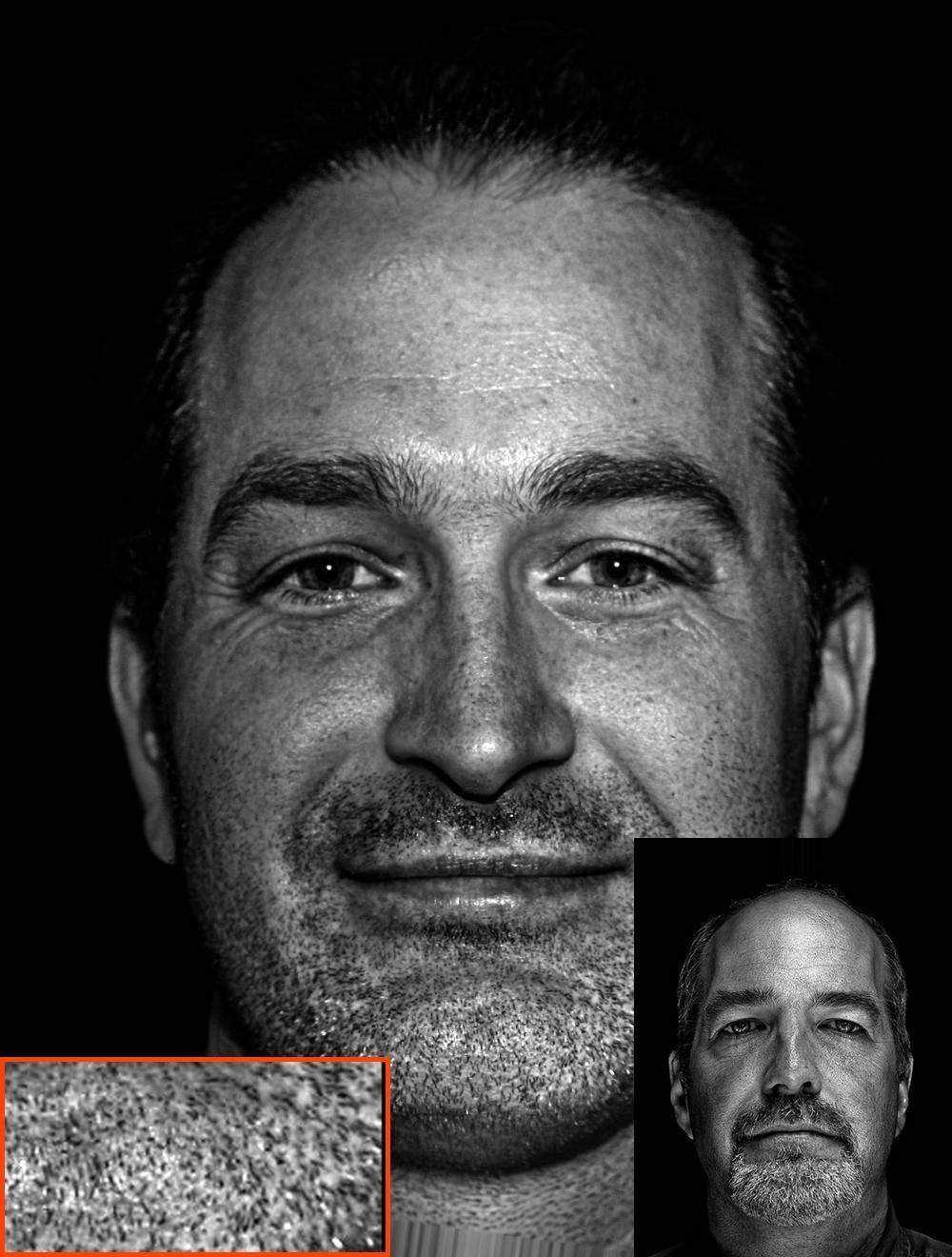}&
\includegraphics[width=.245\linewidth]{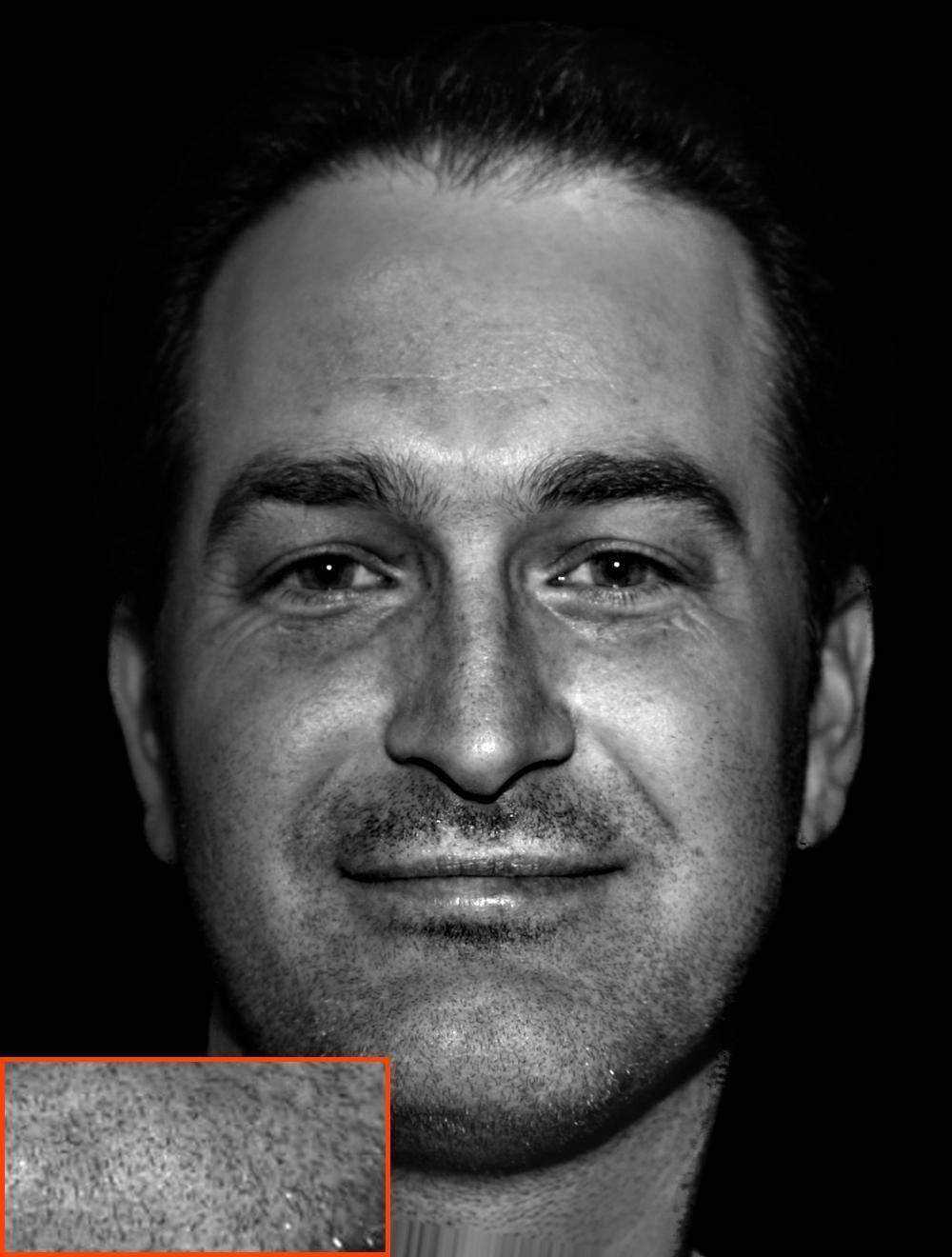}\\
(a) Input photo&
(b) Holistic \cite{Sunkavalli-siggraph10-Harmonization}&
(c) Local \cite{Shih-siggraph14-StyleTransfer}&
(d) Proposed
\end{tabular}
\end{center}
\caption{Qualitative evaluation on the Kelco dataset. The proposed
  method performs favorably against holistic and local methods.
These images can be better visualized with zoom-in to analyze the details.}
\label{fig:exp3}
\end{figure*}

\section{Experimental Results}
In all the experiments we set $\sigma_d$ in Equation \ref{eq:data} to be $0.5$ and
$\sigma_c$  of Equation \ref{eq:smooth1} in be $1$.
The Laplacian stack is set to be 5 (which is the same as \cite{Shih-siggraph14-StyleTransfer}).
The image resolution of both input photo and exemplars is $1320\times1000$ pixels.
\re{The resolution of the local patches used by the MRF is $40\times 40$ pixels}.
\re{When smaller patches are used, more artifacts may be introduced due to inconsistency among multiple exemplars}.
For the artifact removal process, the radius of the guided filter is $60$ pixels.
Note that we first generate the local energy map for each
layer in the Laplacian stack and warp this map using the dense correspondence field.
The evaluation is conducted on the benchmark dataset from \cite{Shih-siggraph14-StyleTransfer}.
The numbers of photos from the Platon, Martin and Kelco collections are 34, 54 and 77,
respectively.
As shown in Figure \ref{fig:style},
the photography style of each collection is drastically different from each other.
In addition, all the exemplars differ significantly from 98 input photos which are obtained from Flickr \cite{Flickr}.

We evaluate the proposed algorithm against the state-of-the-art
methods \cite{Sunkavalli-siggraph10-Harmonization,Shih-siggraph14-StyleTransfer}.
The results of these two methods are generated using the code provided
by authors.
For each photo we use the same exemplar from the collections for
these two methods which is selected by \cite{Shih-siggraph14-StyleTransfer}.
In the following, we present evaluation results on different
collections.
More experimental results can be found at the authors' website.

\subsection{Qualitative Evaluation}
\label{sec:visual_eval}
We evaluate all the comparing methods on the Platon dataset in
Figure \ref{fig:exp1} where
the input photos are acquired under varying lighting conditions.
The holistic method \cite{Sunkavalli-siggraph10-Harmonization}
does not perform well as there is strong contrast in the images.
It generates numerous artifacts on the regions with cast shadows
shown on the first row.
The local method \cite{Shih-siggraph14-StyleTransfer}
alleviates dark lighting effects on the
left cheek with the guidance of corresponding regions from the exemplar.
However, it is less effective to transfer details around the right eye region
mainly because the corresponding region of the exemplar is also dark.
The input photos and exemplars on the second and third rows of Figure \ref{fig:exp1} contain significant differences in facial components (e.g., long and short hair).
Neither of these two methods are able to transfer style naturally.
\re{In contrast, the proposed algorithm consistently selects similar
facial components from multiple exemplars, and
effectively transfers local contrast for stylization}.

Figure \ref{fig:exp2} shows the evaluation results using exemplar images
from the Martin dataset.
As a global transform is used in the holistic method, local details
are likely to be lost and the results are unnatural especially
around nose and mouth regions as shown on the first row of Figure \ref{fig:exp2}(b).
%
%
The local method can successfully transfer local contrast when
the input photo and exemplar have similar facial components.
However, it also transfers the high frequency details
of one exemplar to the stylized result, thereby making the image unnatural
when the exemplar and input photo have distinct local contents.
As shown in Figure \ref{fig:exp2}(c),
the wrinkle, beard and hair of the exemplar are transferred to the stylized image.
By using multiple exemplars the proposed algorithm can effectively
transfer lighting and low frequency components
of exemplars without obvious artifacts.
Compared to holistic and local methods, the proposed algorithm is more
effective in transferring local contrast and preserving nature \re{appearances} of
the input photos.

In Kelco dataset shown in Figure \ref{fig:exp3}, the local method is less
effective in transferring details around the dissimilar regions (e.g., hair).
For holistic method, the difference in the luminance distribution
results in unnatural stylized image.
Although a portrait may be acquired under various lighting conditions
with different facial components that are not well described or matched by one single exemplar,
with a collection of exemplars the proposed algorithm can accurately
identify corresponding patches for each photo patch to transfer
local details effectively.

\subsection{Quantitative Evaluation}
In quantitative evaluations we first compare the
results generated by different methods with one reference
image edited by an artist.
A human subject study is then conducted to evaluate
the local method \cite{Shih-siggraph14-StyleTransfer}
and the proposed algorithm.

\subsubsection{Evaluation with Reference Image}
\label{sec:obj}

We evaluate the results generated by three methods.
The exemplar is manually selected for holistic and local methods.
Instead of relying on automatic exemplar selection as
carried out in \cite{Shih-siggraph14-StyleTransfer},
this manually selected exemplar is used as the most similar one to the input photo.
We use the PSNR and FSIM \cite{lin-tip2011-fsim} metrics
to measure the tone and feature similarities with the reference images.

\begin{figure}[t]
\begin{center}
\begin{tabular}{ccc}
\includegraphics[width=0.33\linewidth]{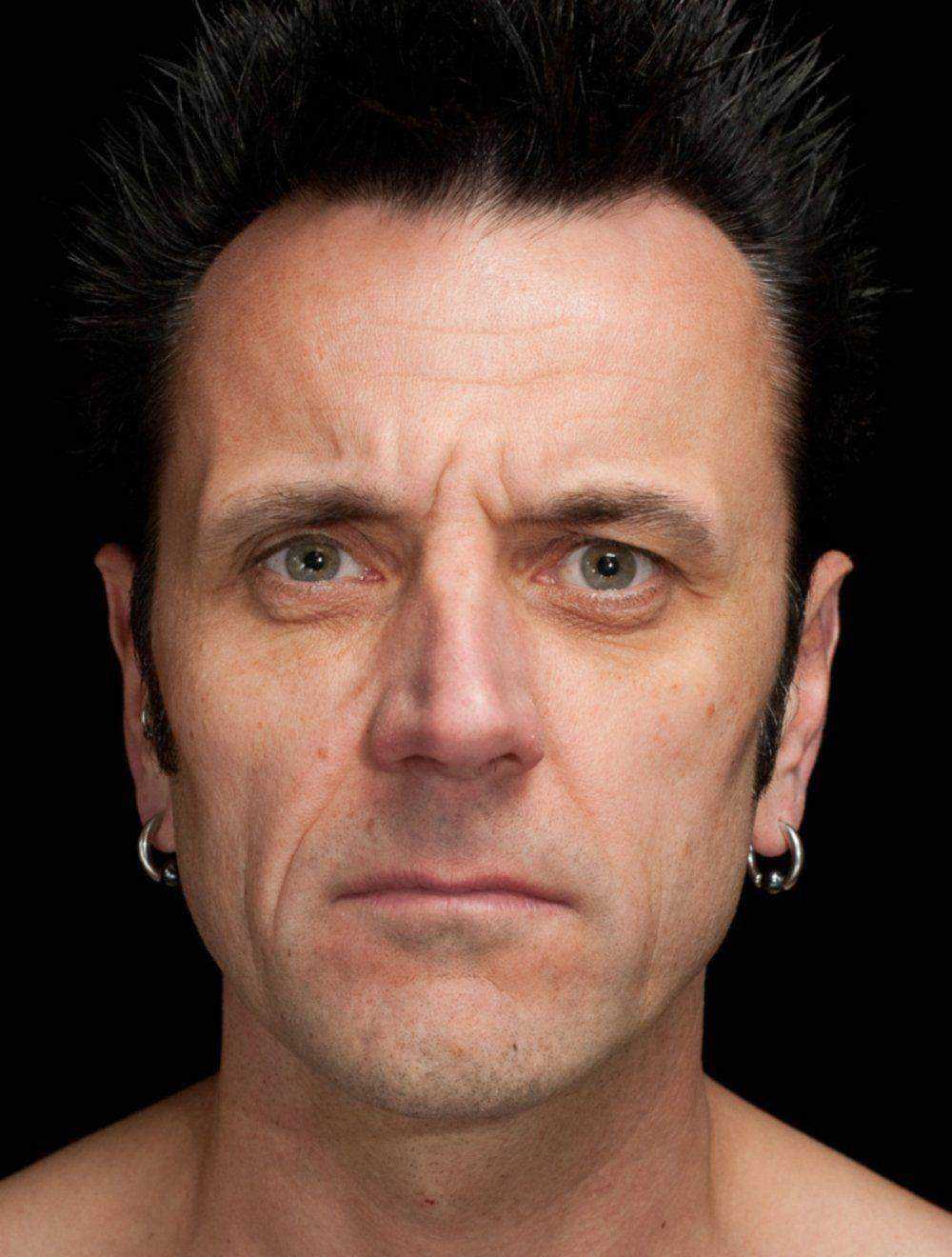}&
\includegraphics[width=0.33\linewidth]{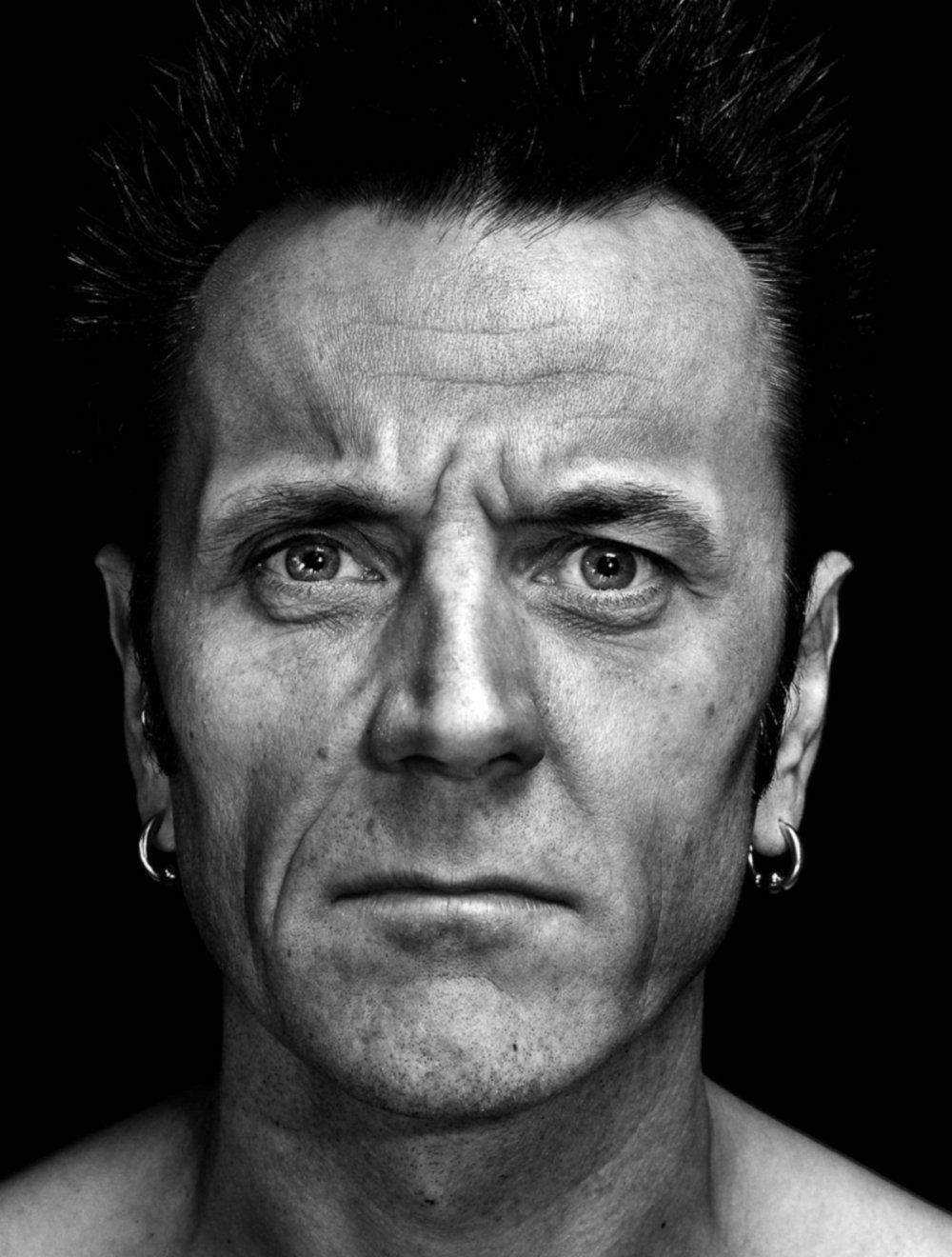}&
\includegraphics[width=0.33\linewidth]{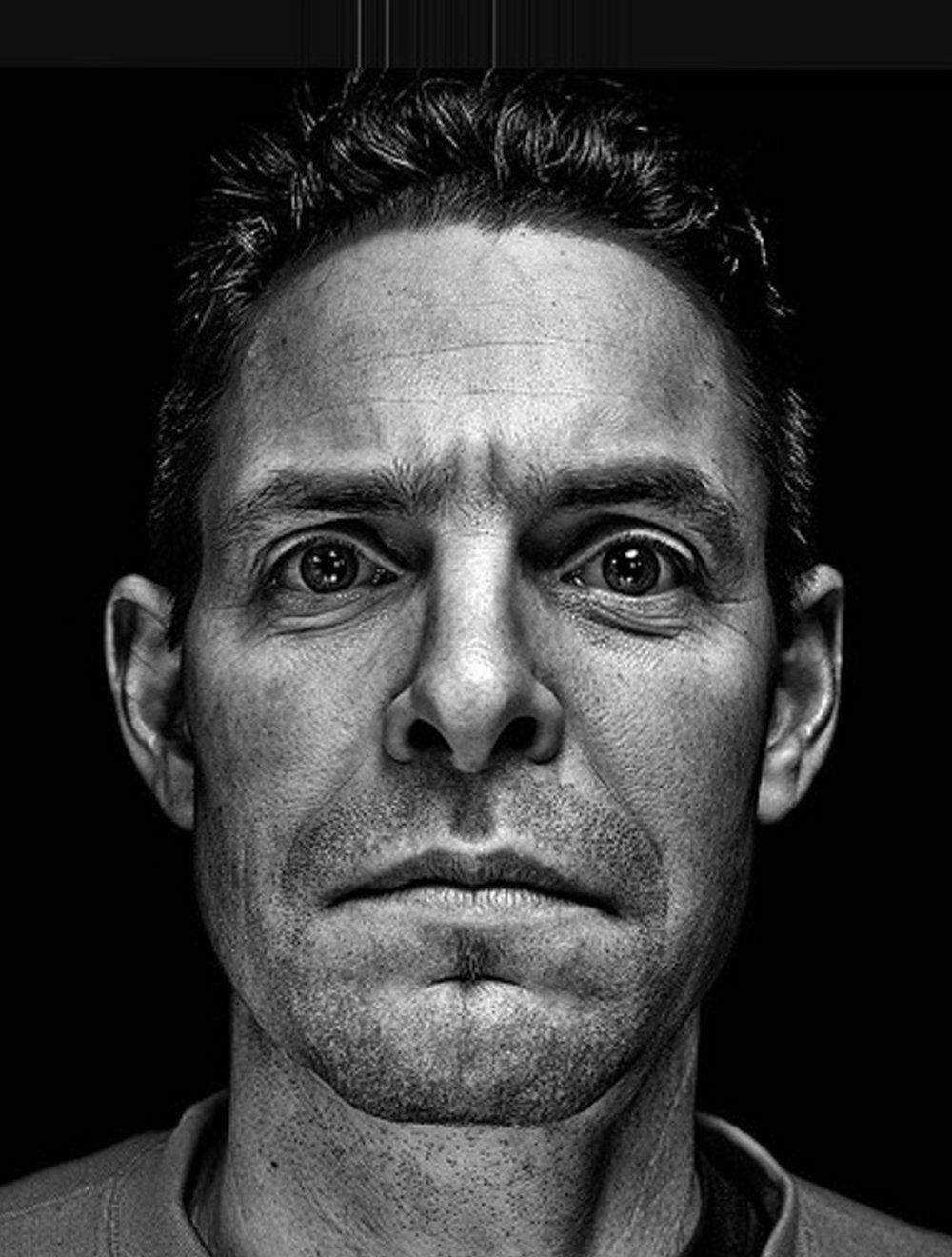}\\
(a) Input photo&(b) Reference&(c) Exemplar\\
\includegraphics[width=0.33\linewidth]{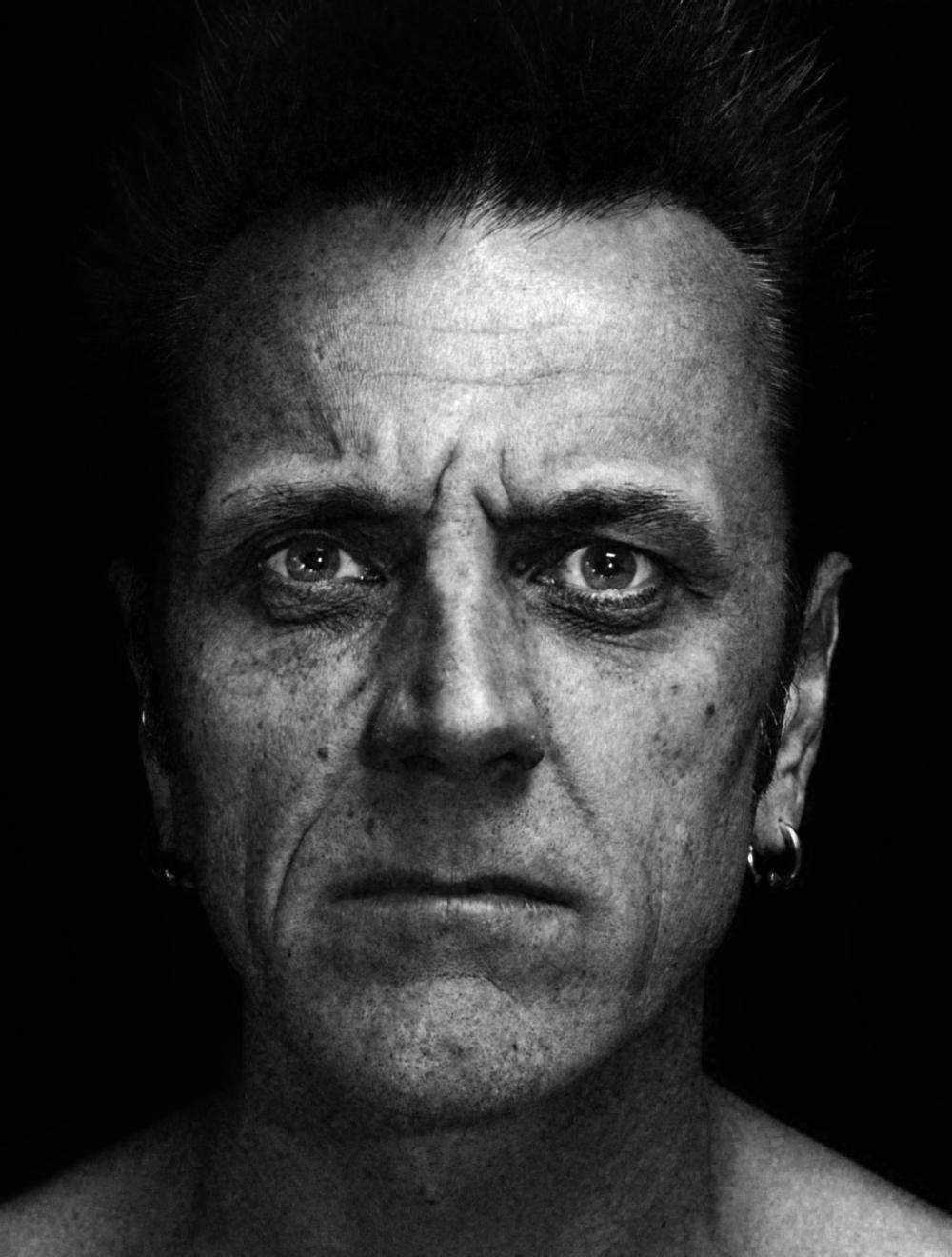}&
\includegraphics[width=0.33\linewidth]{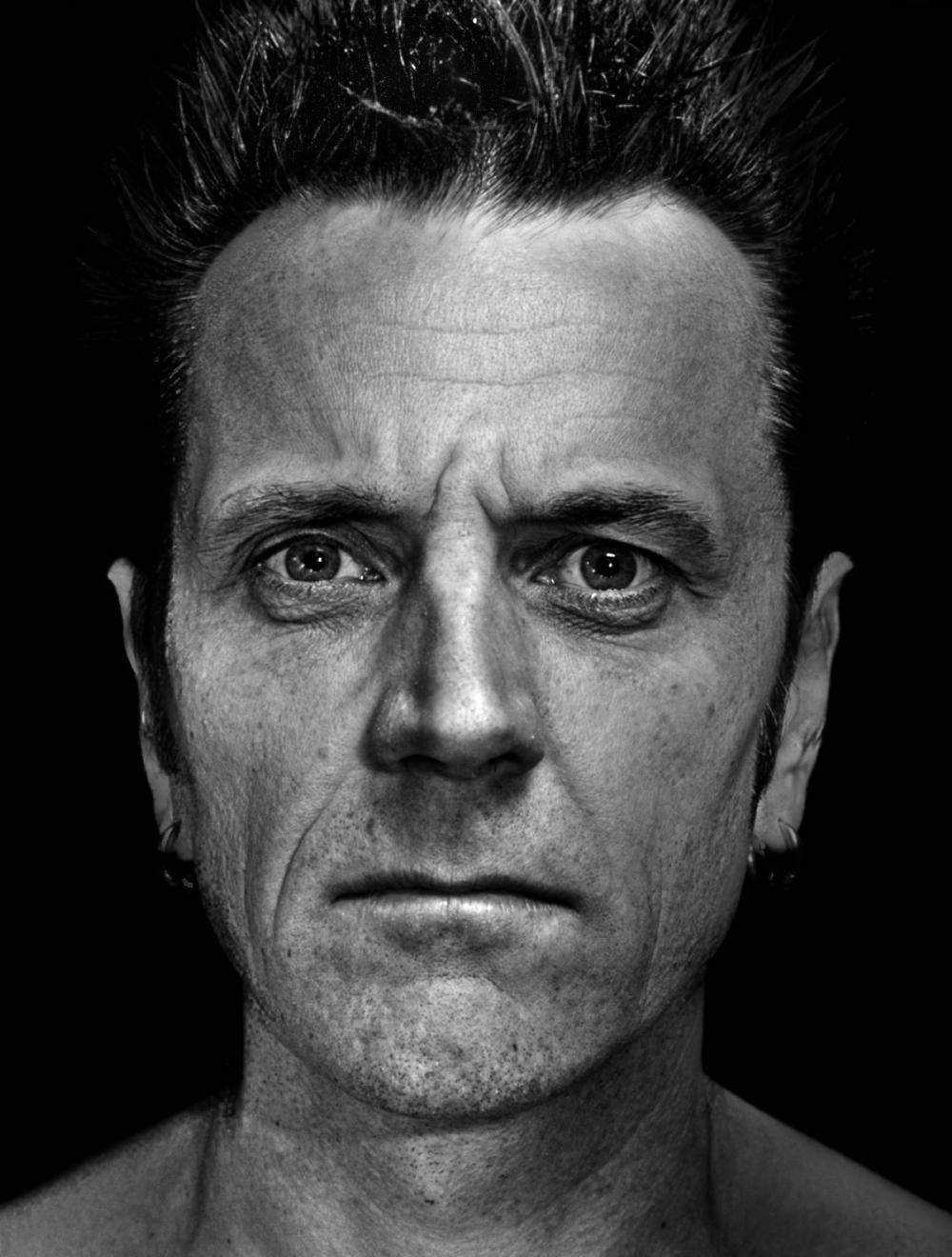}&
\includegraphics[width=0.33\linewidth]{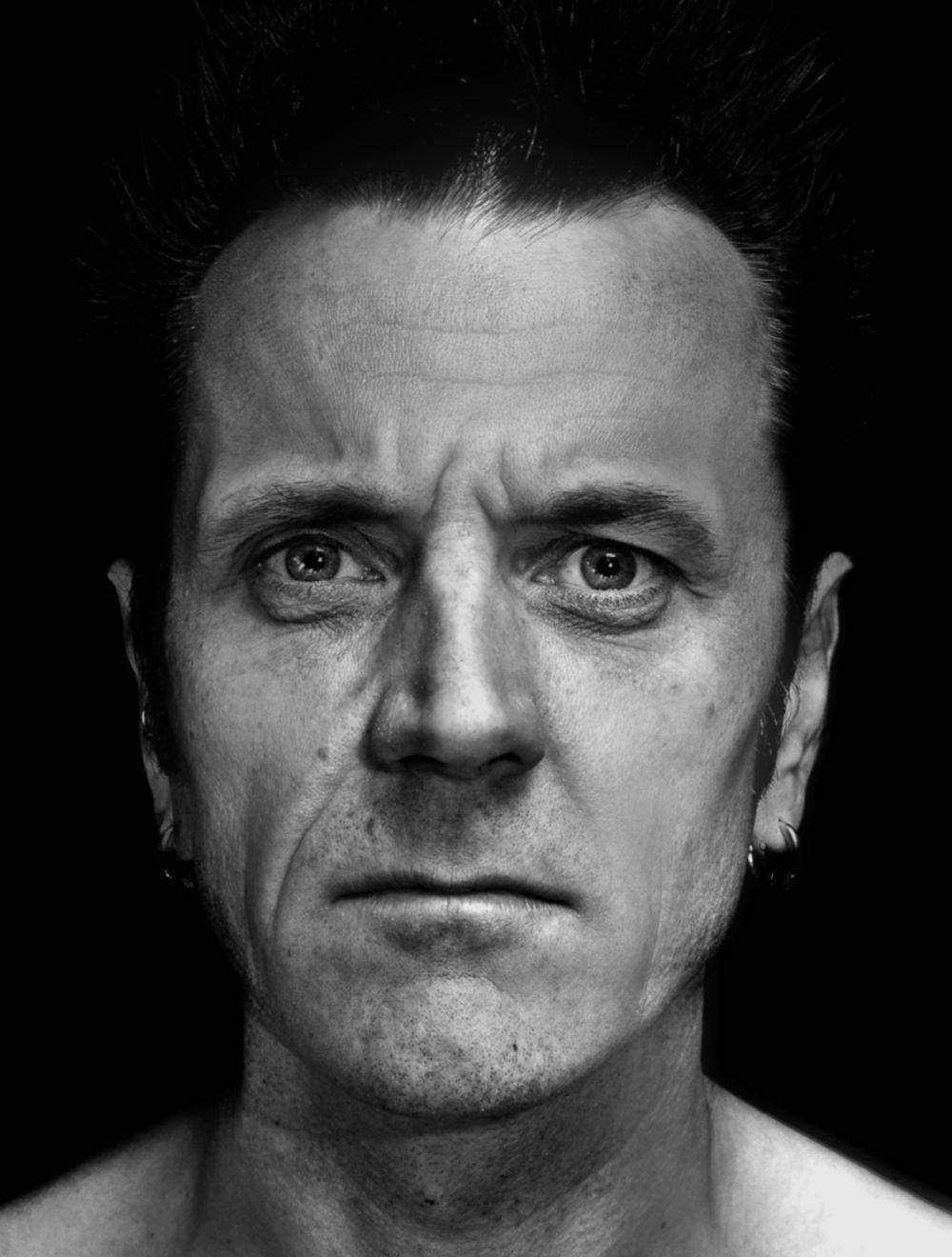}\\
(d) Holistic \cite{Sunkavalli-siggraph10-Harmonization}&(e) Local \cite{Shih-siggraph14-StyleTransfer}&(f) Proposed\\
\small{PSNR: 17.8563}&\small{17.9212}&\small{\textbf{20.4450}}\\
\small{FSIM: 0.9665}&\small{0.9676}&\small{\textbf{0.9759}}
\end{tabular}
\end{center}
\caption{Quantitative evaluation using one reference image.
(a) input photo.
(b) reference photo manually edited by an artist.
(c) exemplar manually selected from collections.
(d)-(f) evaluated methods where (c) is adopted in (d) and (e) during
style transfer.
PSNR and FSIM \cite{lin-tip2011-fsim} are used for evaluations.}
\label{fig:metric}
\end{figure}

\begin{figure}[!ht]
\begin{center}
\begin{tabular}{c}
\includegraphics[width=.95\linewidth]{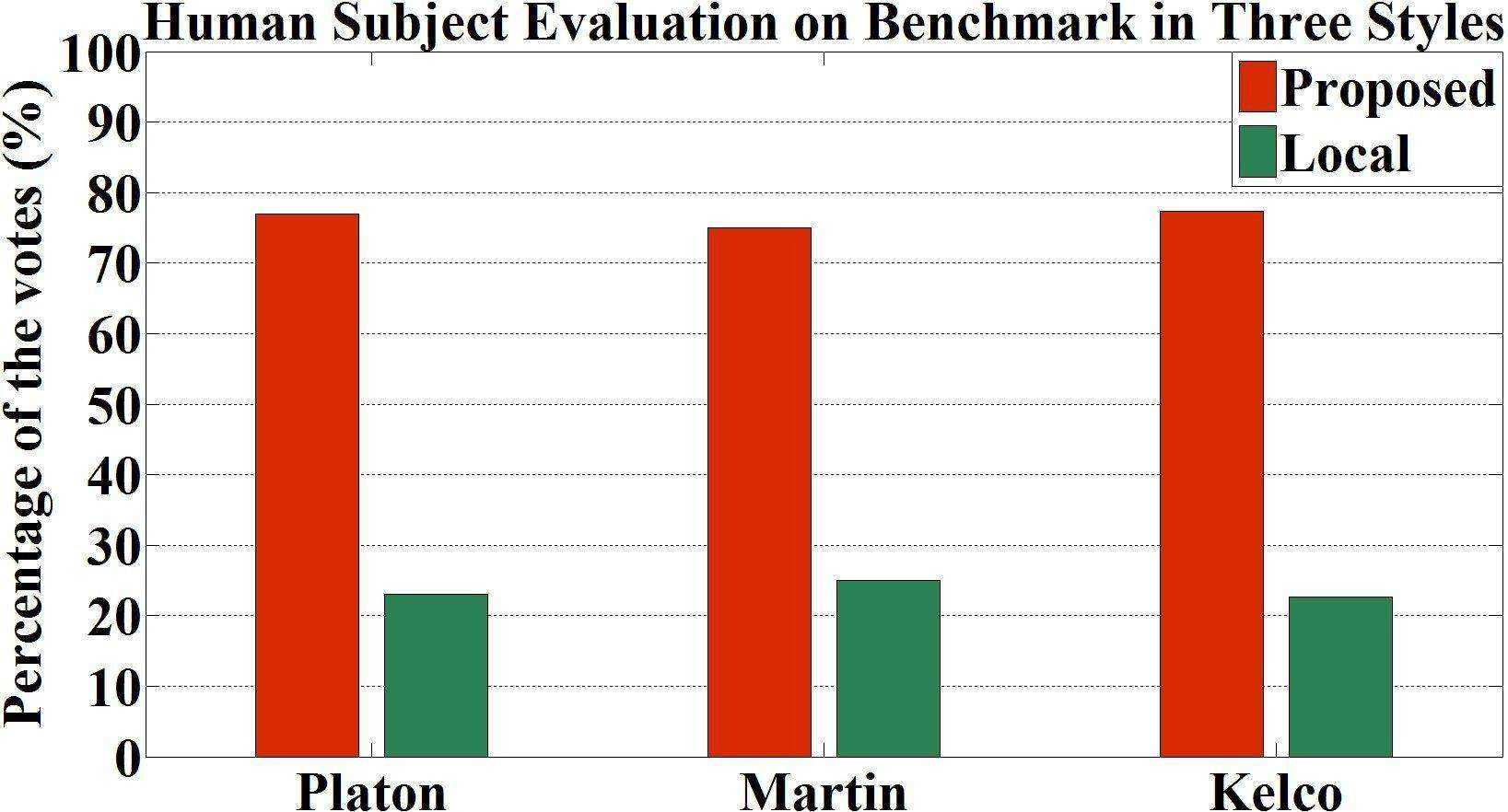}\\
\end{tabular}
\end{center}
\caption{Human subject evaluation on the input photos.
For each category proposed method is compared with local method among
45 subjects inside the university.
Each participant is asked to select the result containing less
artifacts, thus choosing the image in which local contrast is transferred most effectively.}
\label{fig:vote45}
\begin{center}
\begin{tabular}{c}
\includegraphics[width=.95\linewidth]{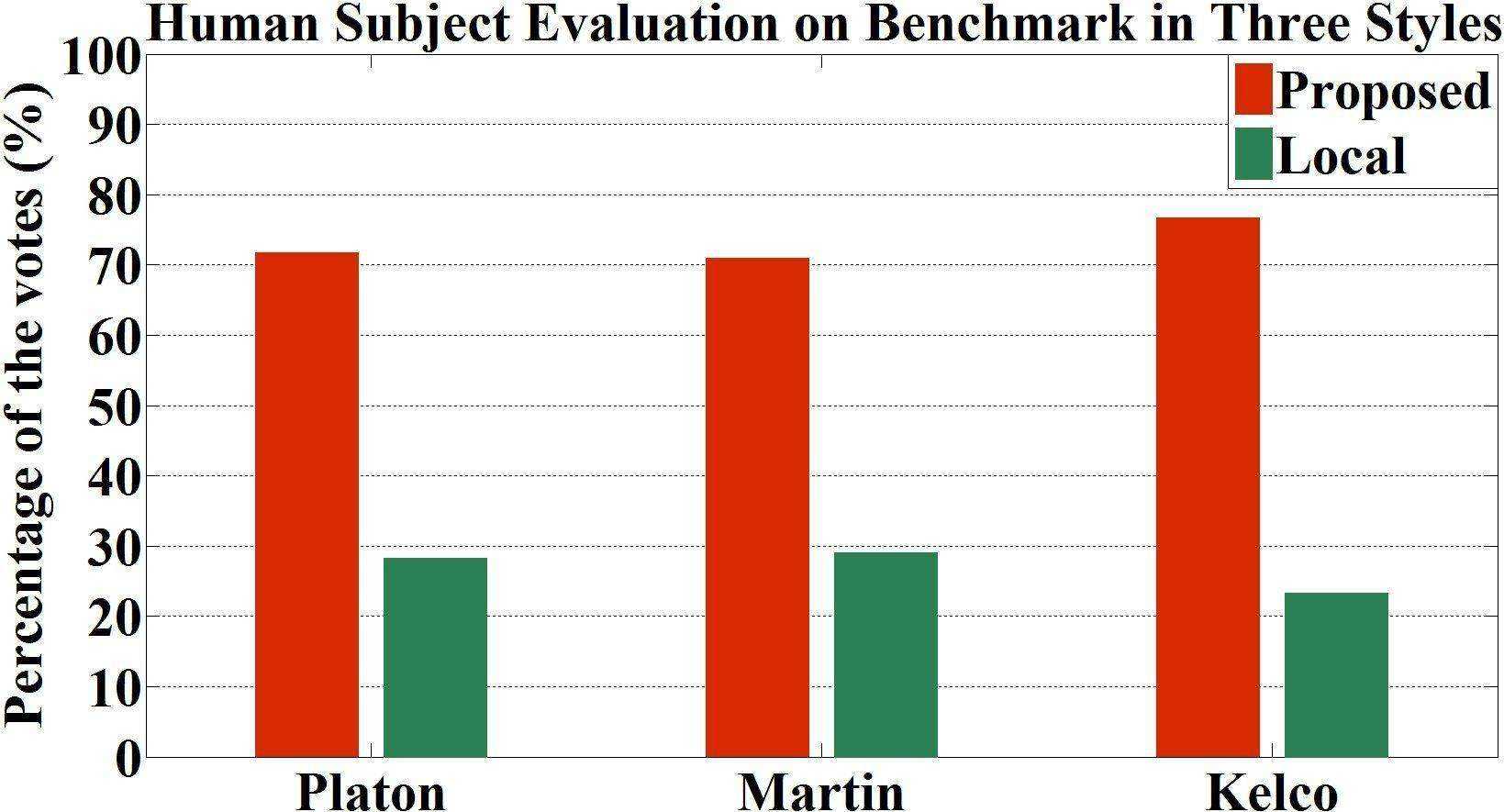}\\
\end{tabular}
\end{center}
\caption{Human subject evaluation on the input photos.
For each category proposed method is compared with local method among 20 subjects outside the university.
For each input image, the subjects are asked to select the result which
is more effective to transfer each style from the general feeling.
}
\label{fig:vote20}
\end{figure}

Figure \ref{fig:metric} shows the evaluation result where
the reference image shown in (b) is manually edited by an artist.
The proposed algorithm performs favorably
against the other methods in terms of PSNR and FSIM.
The exemplar shown in (c) shares many similarities to the input photo
in facial components (i.e, eyes, nose, mouth and ears).
However, it still contains differences around the hair and shoulder regions.
The hair region of the exemplar is bright while it is dark in
the input photo.
On the other hand, the shoulder of the exemplar is not as bright as that in the
input photo.
Despite significant similarities, these differences affect
how the holistic and local methods
generate stylized face images
based on one exemplar as shown in Figure \ref{fig:metric}(d) and (e).
The stylized image by the holistic method contains artifacts on the face region,
and the result by the local scheme consists of regions with unnatural lighting (e.g.,
bright hair and dark shoulders) when compared with the reference photo.
%
%
In other words, minor differences are likely to affect
existing methods based on a single exemplar holistically or locally.
Furthermore,
we note in practice it is challenging to find a well suited exemplar for an input photo.
However, the proposed method alleviates this problem by establishing the identification
in a collection of exemplars for effective \re{stylization} of facial details.

\subsubsection{Human Subject Evaluation}

\renewcommand{\tabcolsep}{.8pt}
\begin{figure*}[t]
\begin{center}
\begin{tabular}{cc}
\begin{minipage}[b]{0.01\linewidth}
    \centering
    $\vcenter{\rotatebox{-90}{\normalsize Input}}$
\end{minipage}
\begin{minipage}[t]{0.99\linewidth}
    \centering
    \begin{tabular}{ccccc}
    \includegraphics[width=0.19\linewidth]{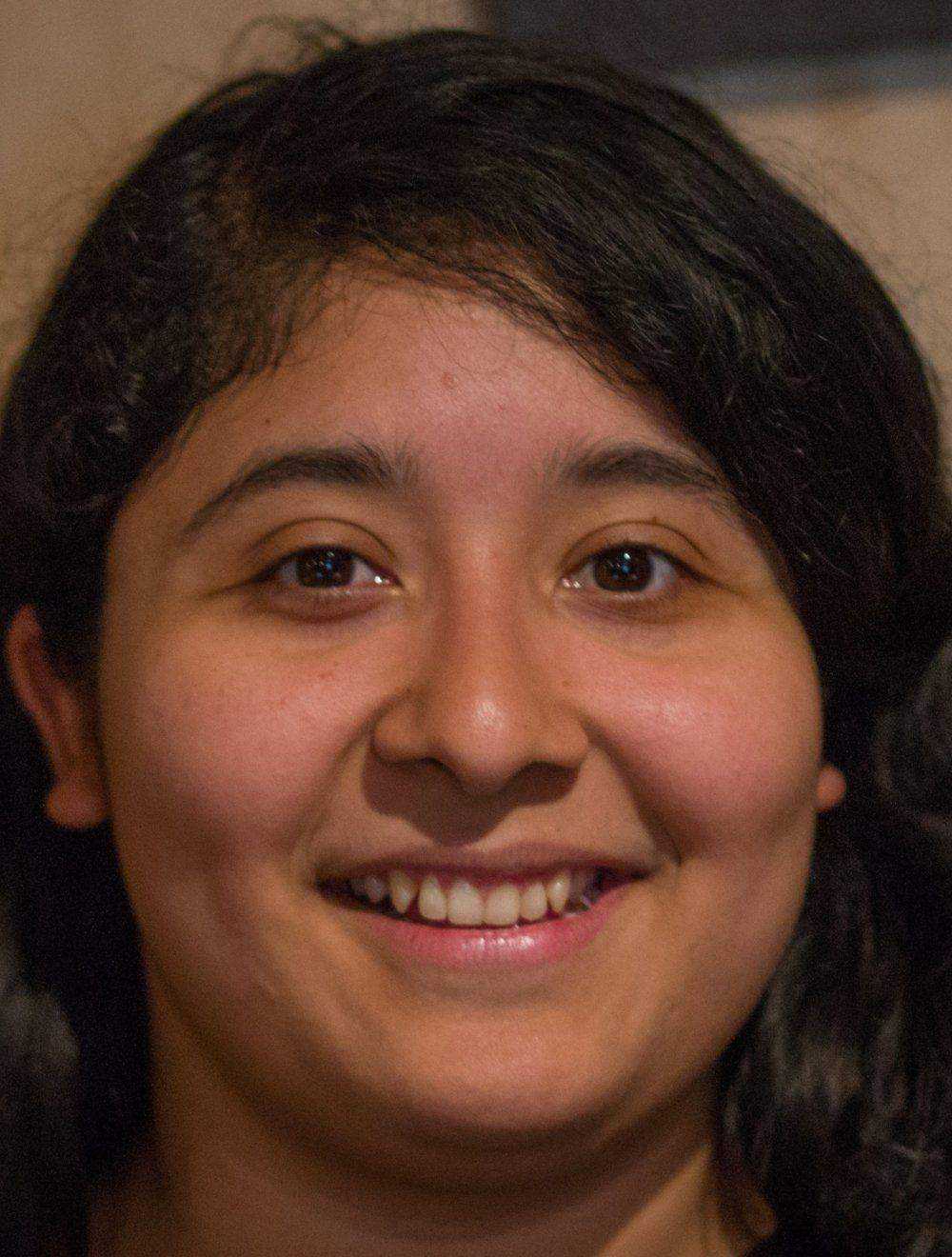}&
    \includegraphics[width=0.19\linewidth]{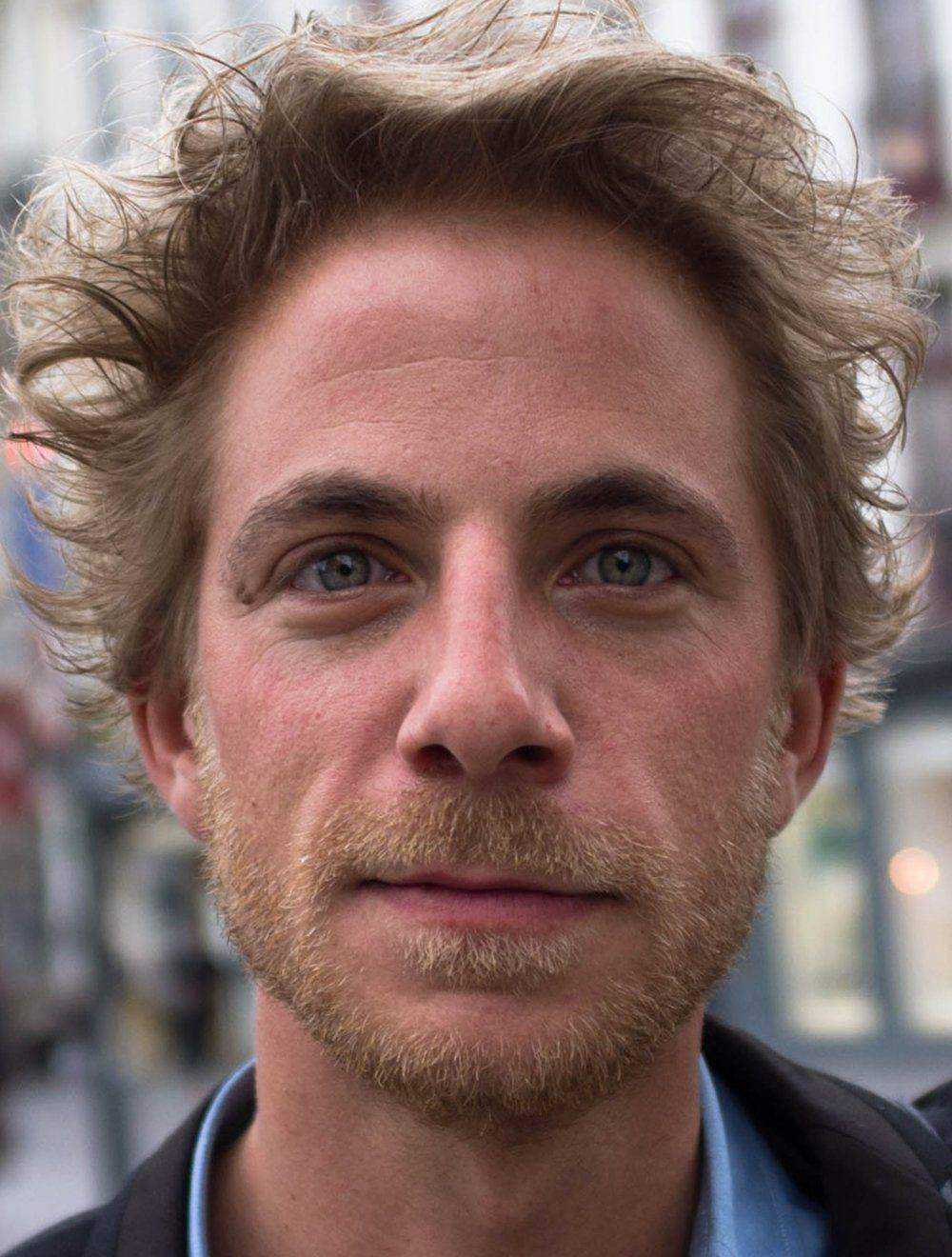}&
    \includegraphics[width=0.19\linewidth]{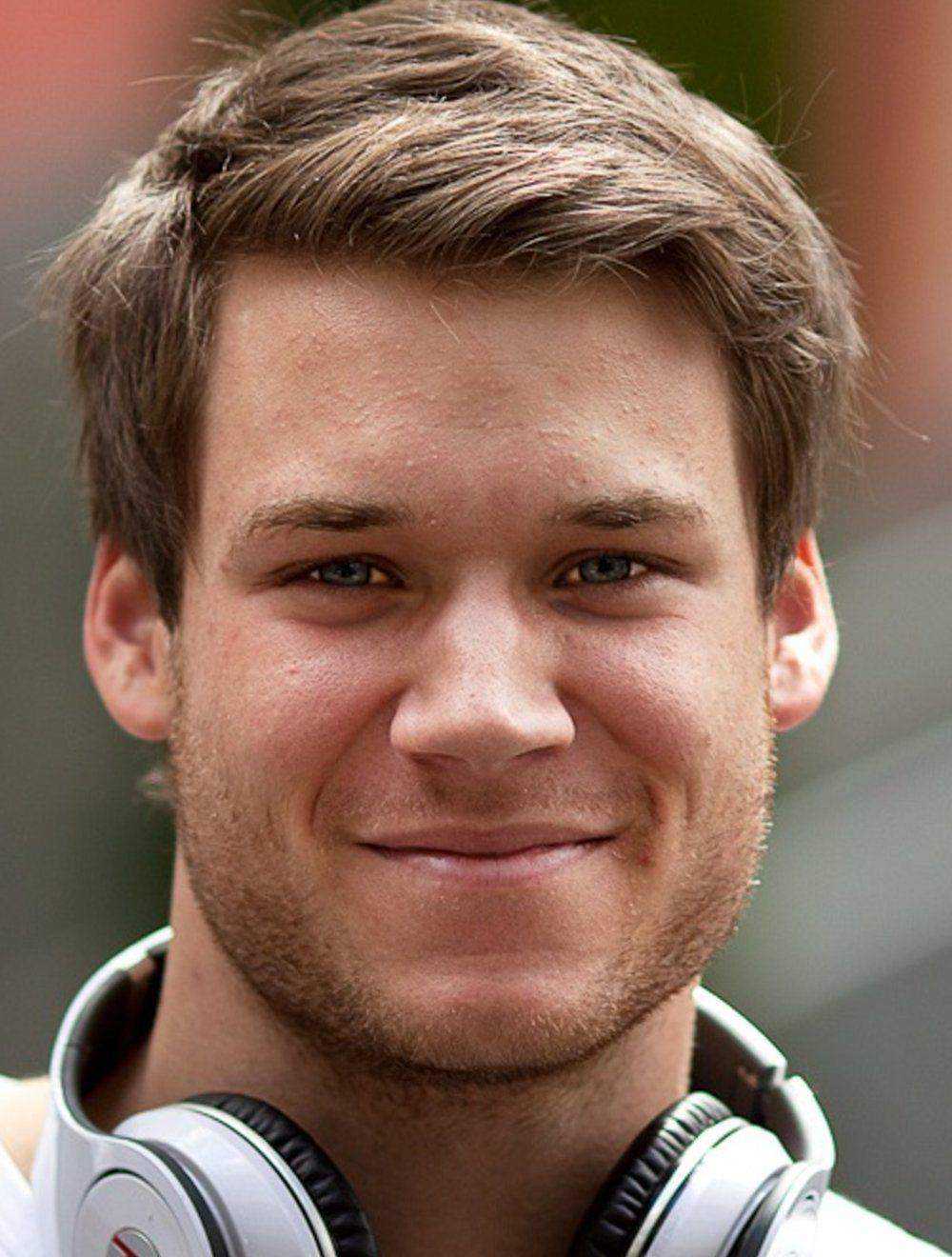}&
    \includegraphics[width=0.19\linewidth]{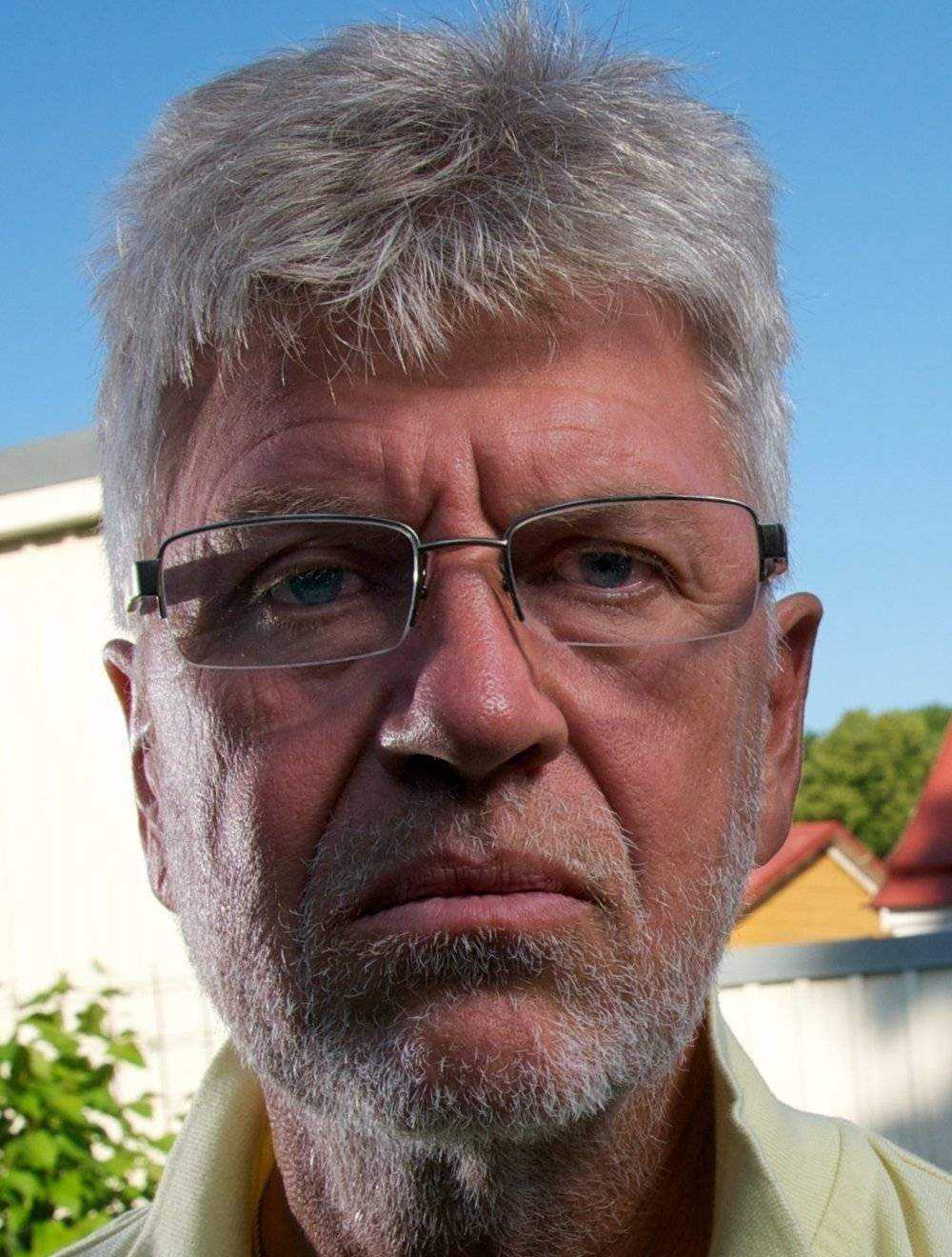}&
    \includegraphics[width=0.19\linewidth]{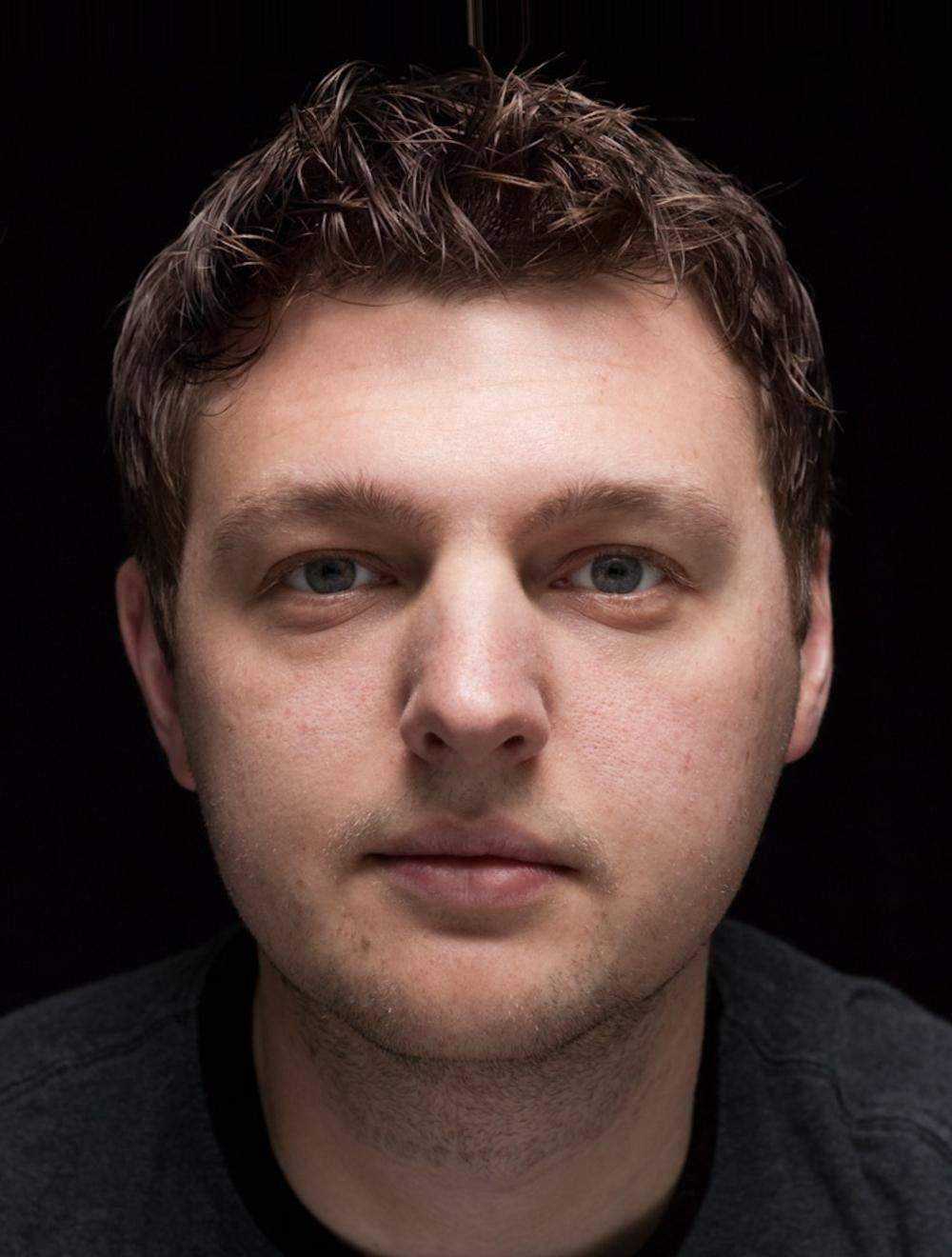}\\
    \end{tabular}
\end{minipage}\\
\begin{minipage}[b]{0.01\linewidth}
    \centering
    $\vcenter{\rotatebox{-90}{\normalsize Local}}$
\end{minipage}
\begin{minipage}[t]{0.99\linewidth}
    \centering
    \begin{tabular}{ccccc}
    \includegraphics[width=0.19\linewidth]{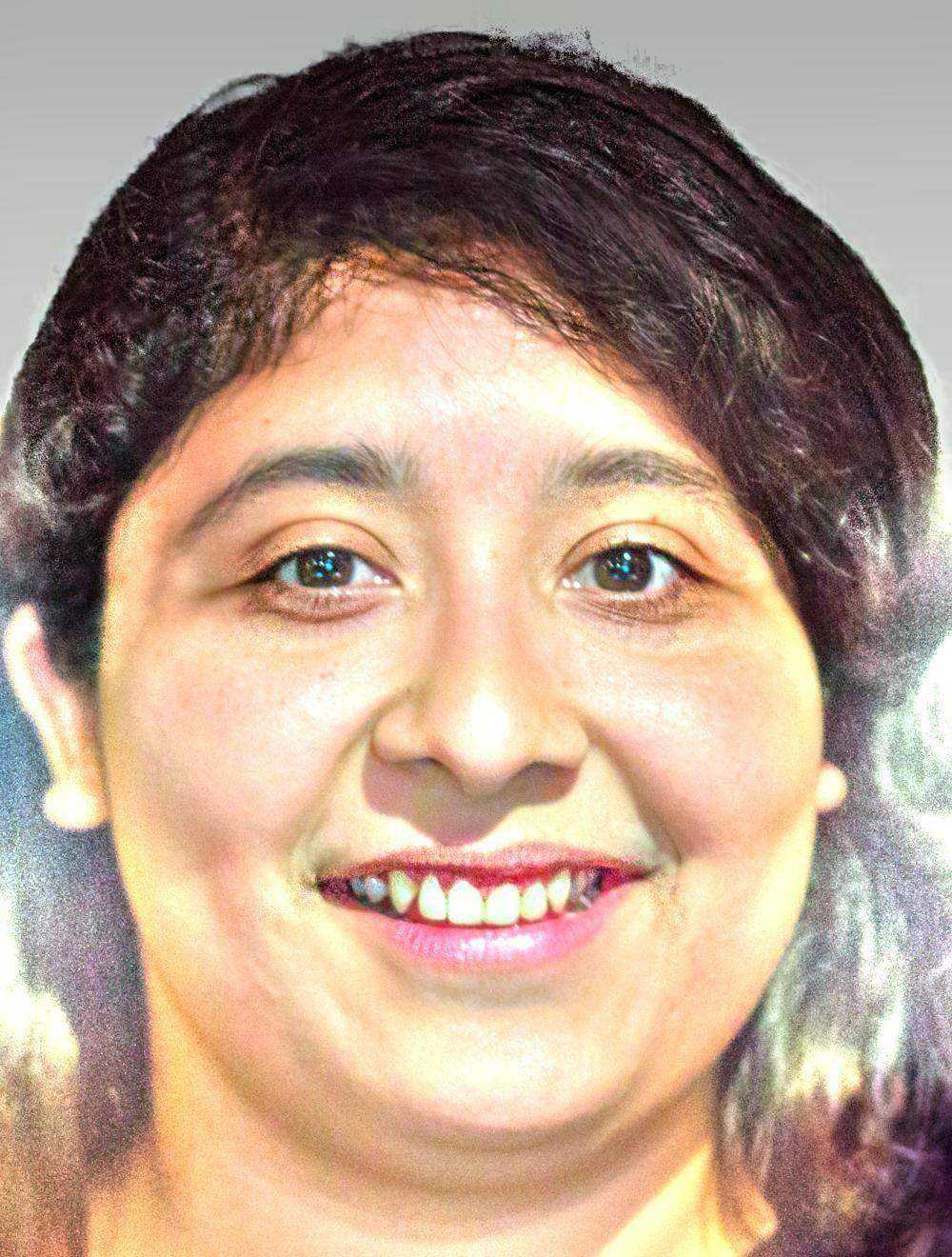}&
    \includegraphics[width=0.19\linewidth]{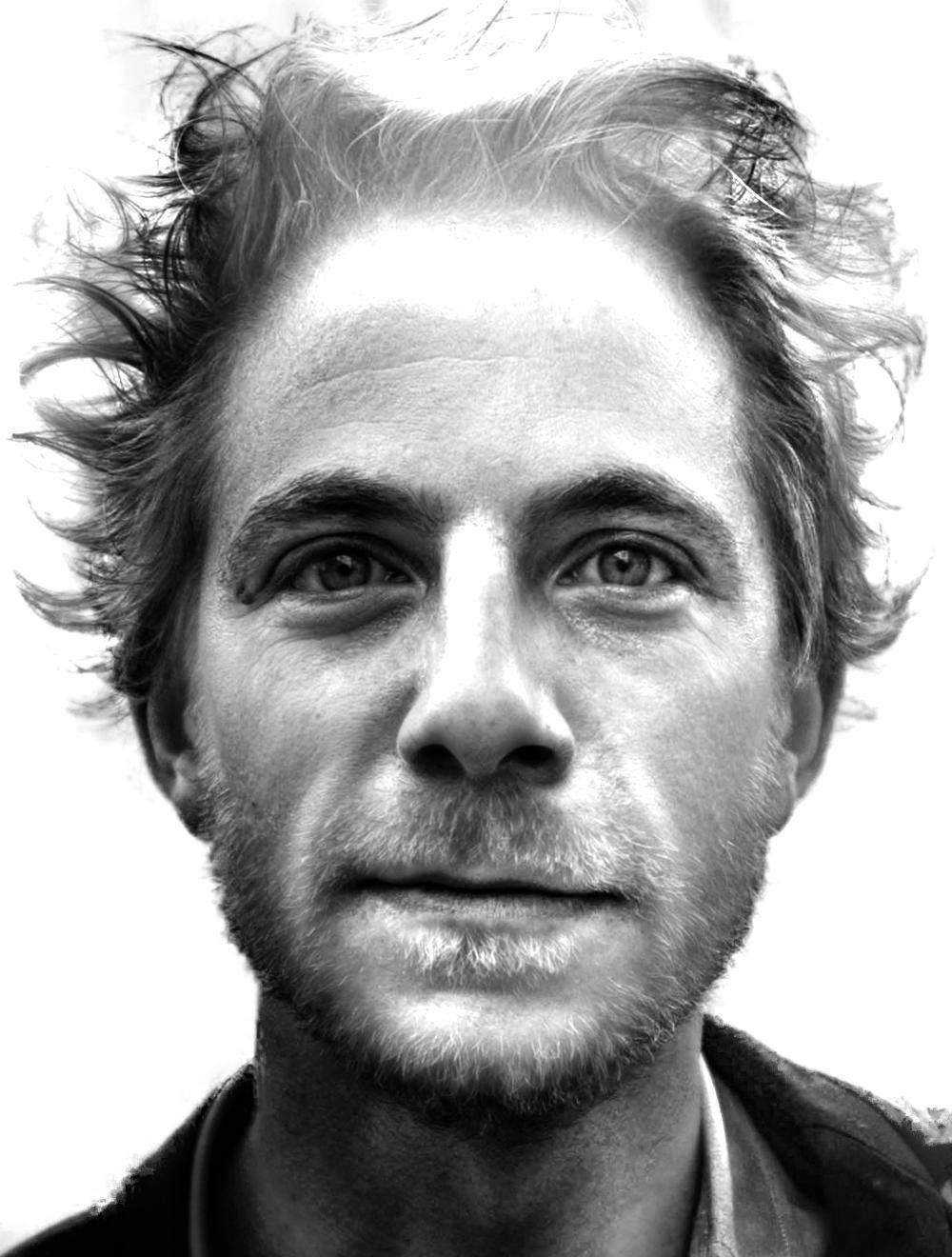}&
    \includegraphics[width=0.19\linewidth]{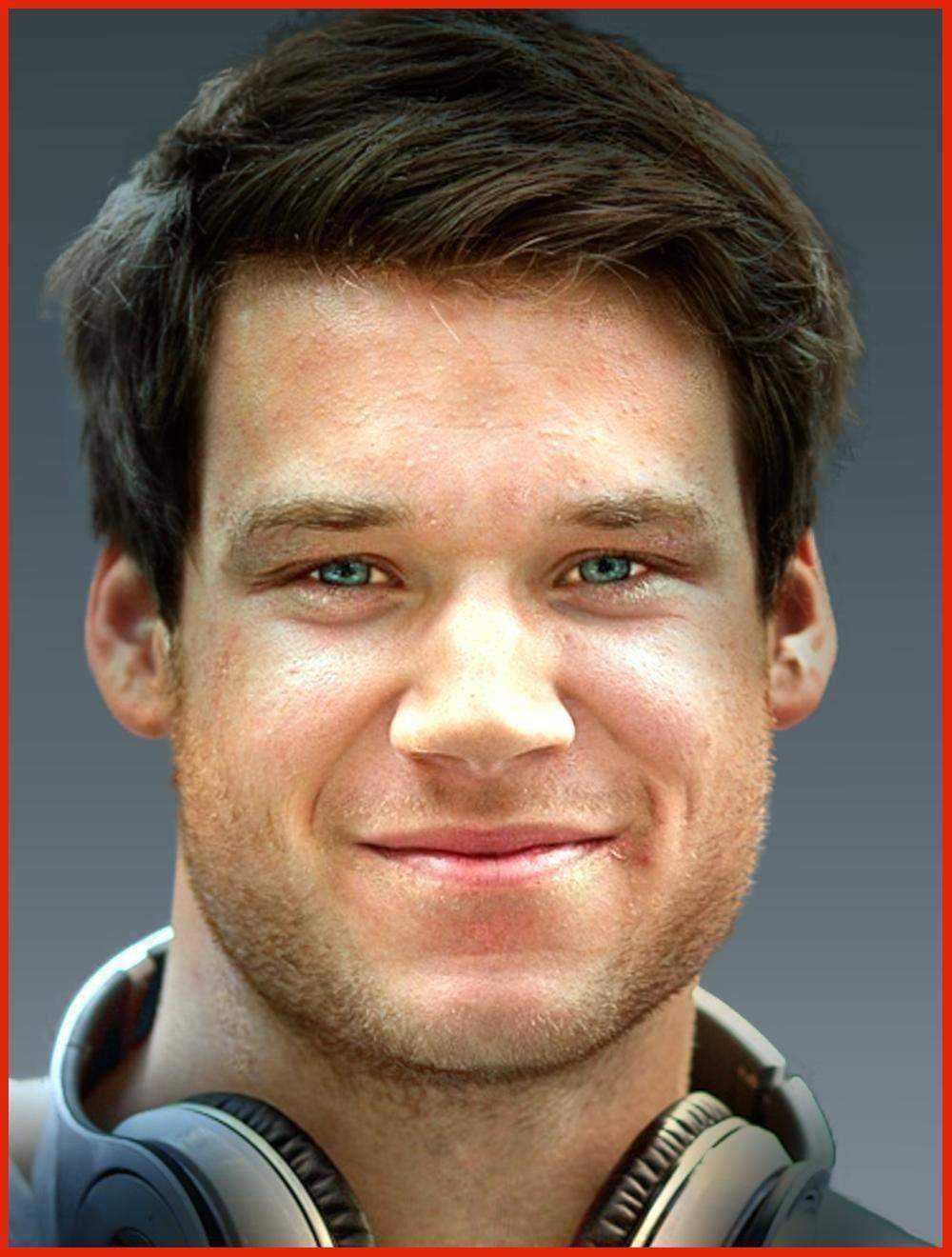}&
    \includegraphics[width=0.19\linewidth]{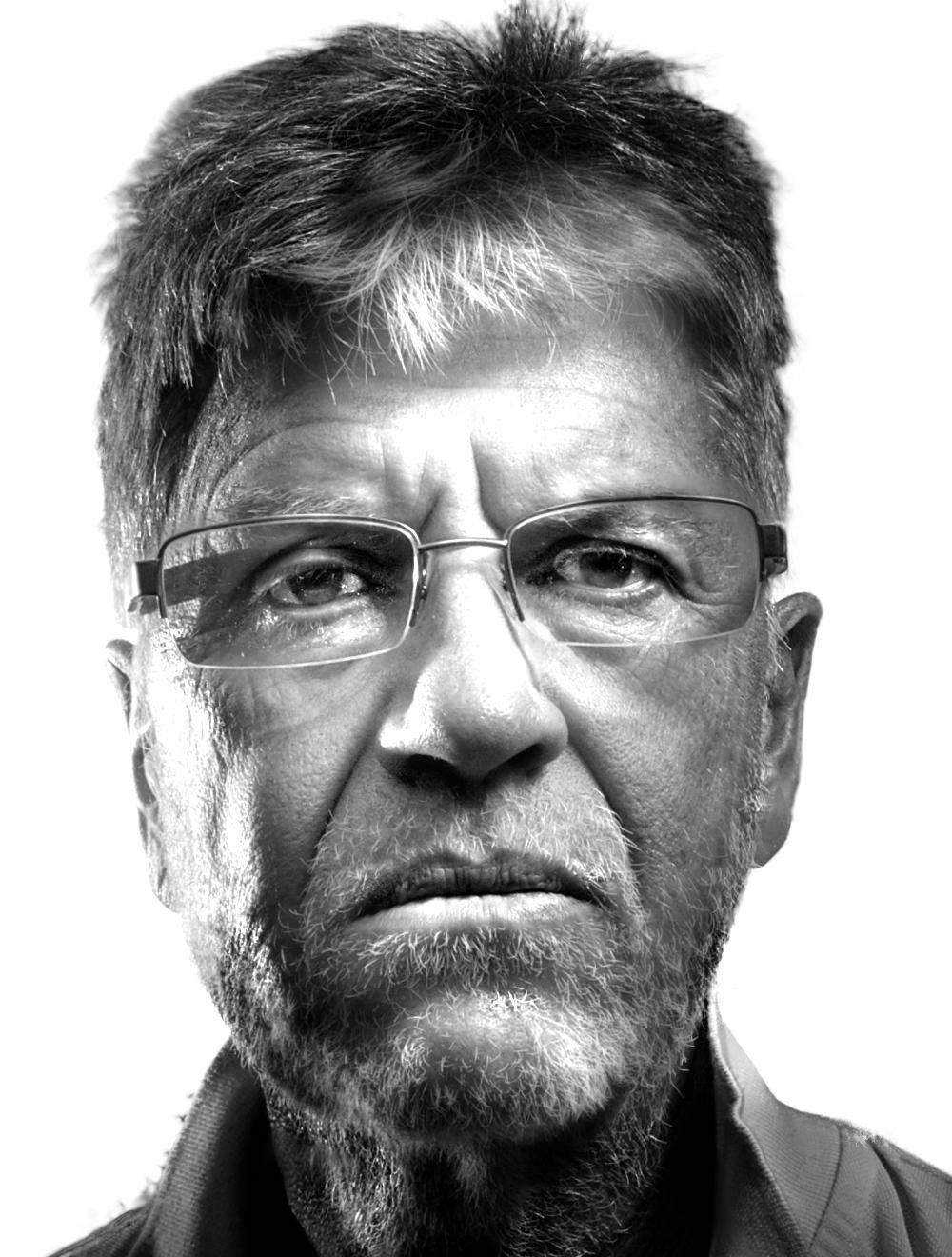}&
    \includegraphics[width=0.19\linewidth]{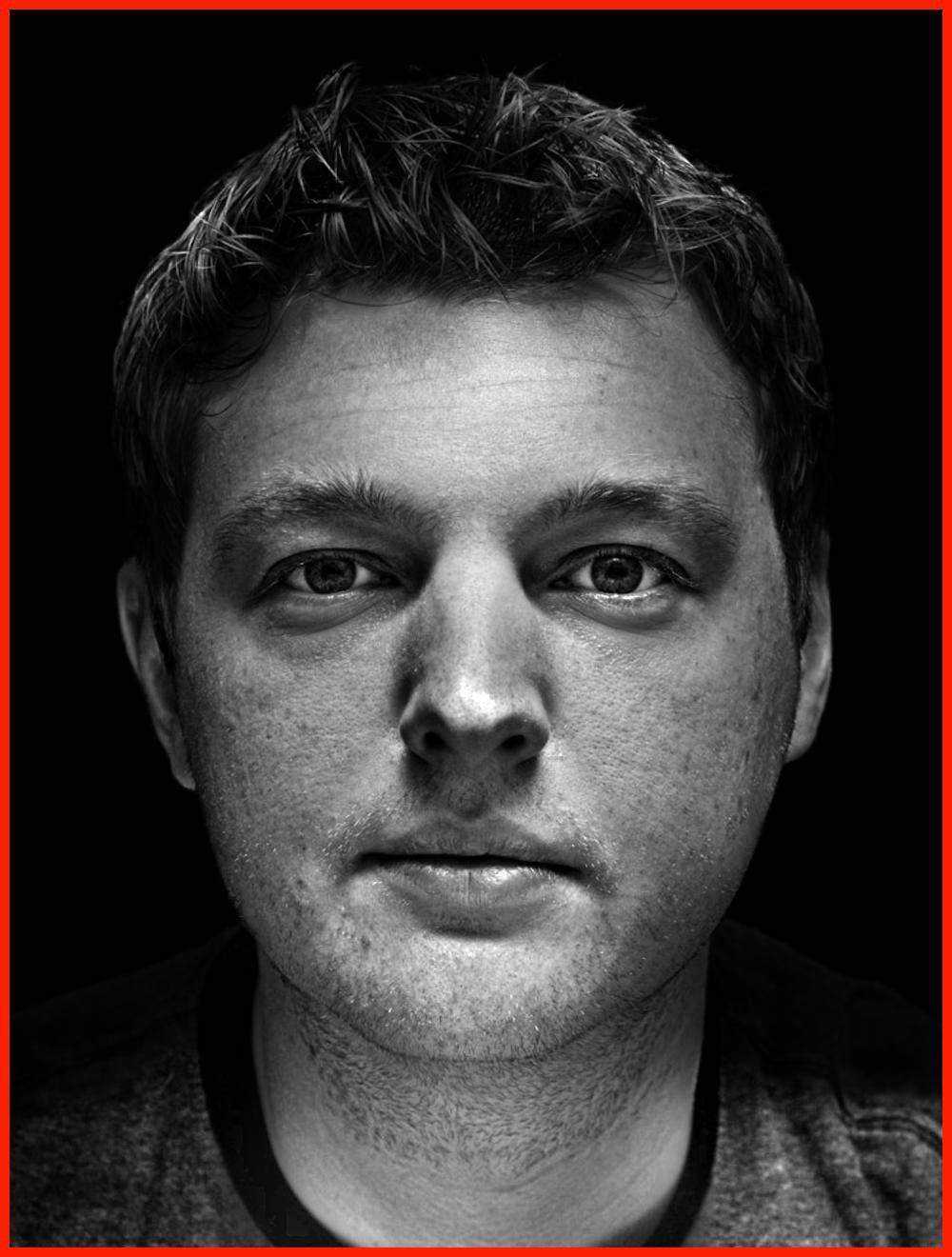}\\
    \end{tabular}
\end{minipage}\\
\begin{minipage}[b]{0.01\linewidth}
    $\vcenter{\rotatebox{-90}{\normalsize Proposed}}$
\end{minipage}
\begin{minipage}[t]{0.99\linewidth}
    \centering
    \begin{tabular}{ccccc}
    \includegraphics[width=0.19\linewidth]{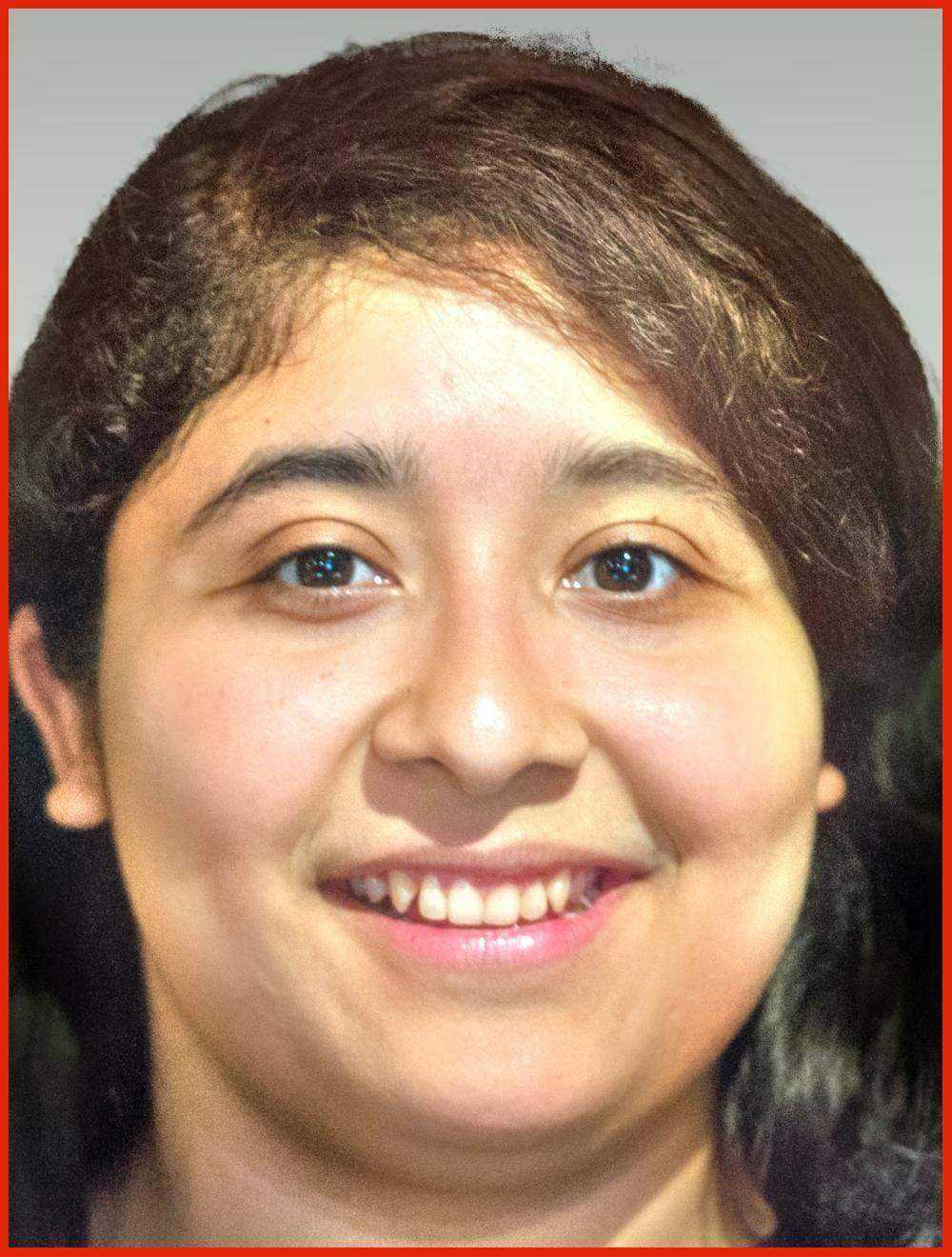}&
    \includegraphics[width=0.19\linewidth]{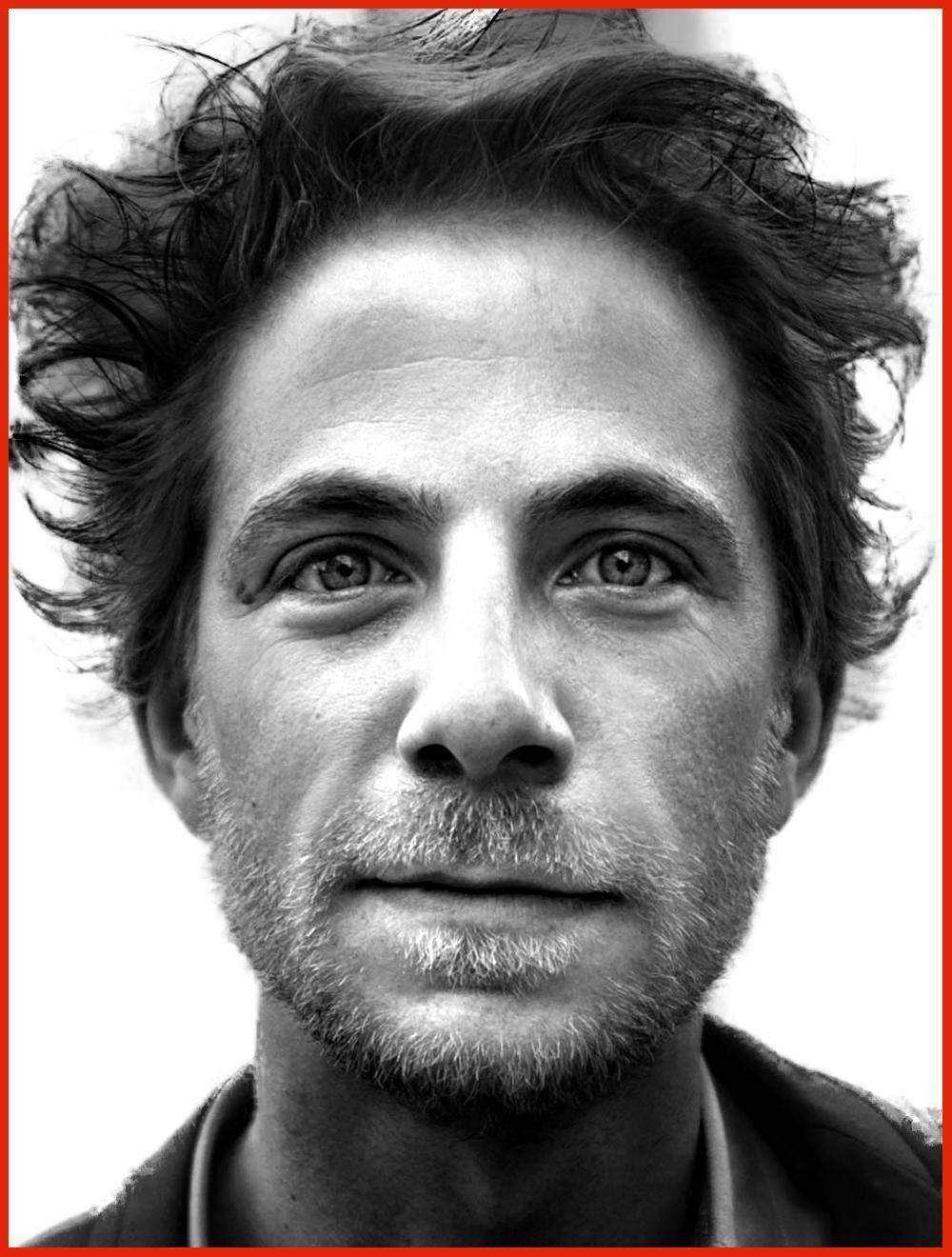}&
    \includegraphics[width=0.19\linewidth]{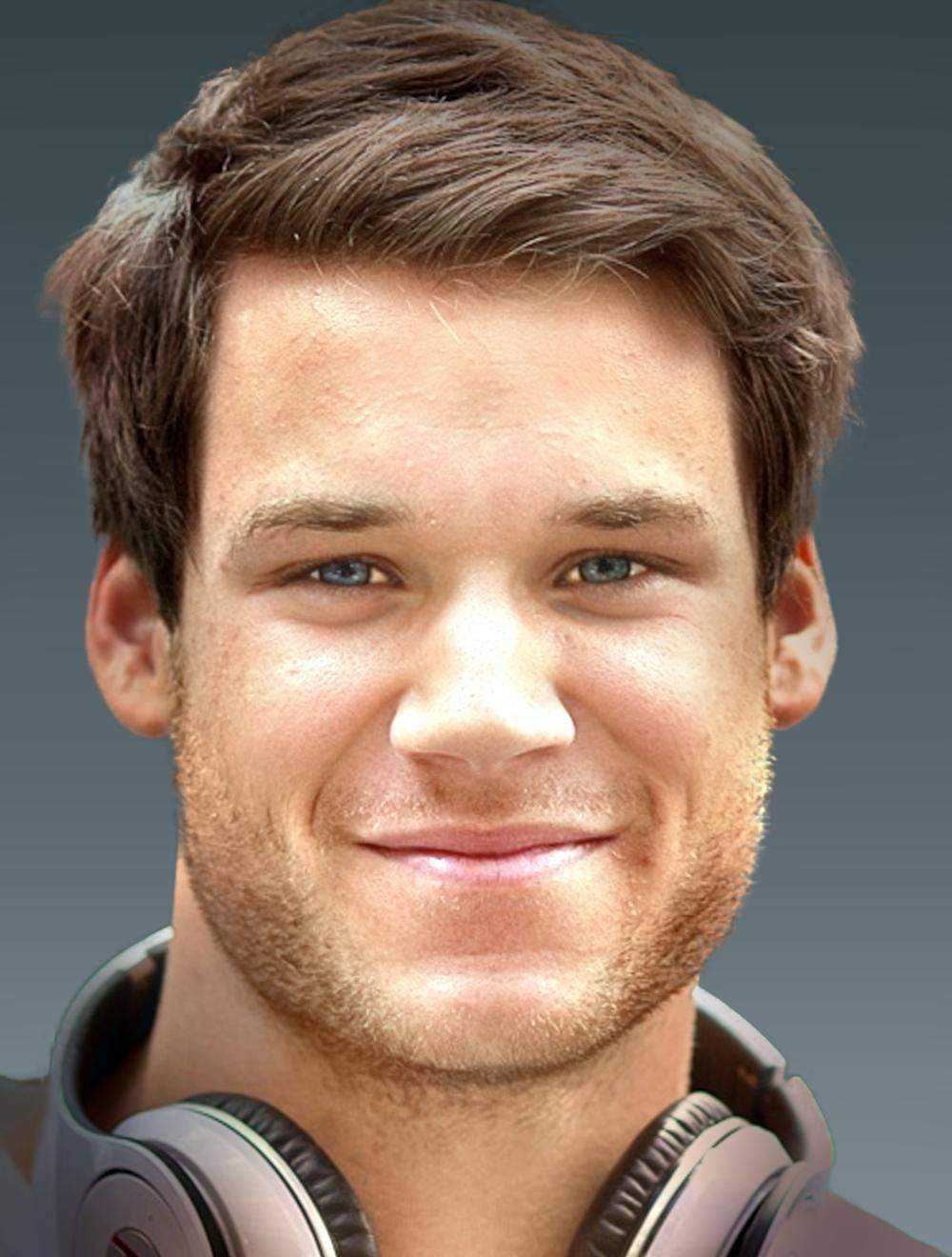}&
    \includegraphics[width=0.19\linewidth]{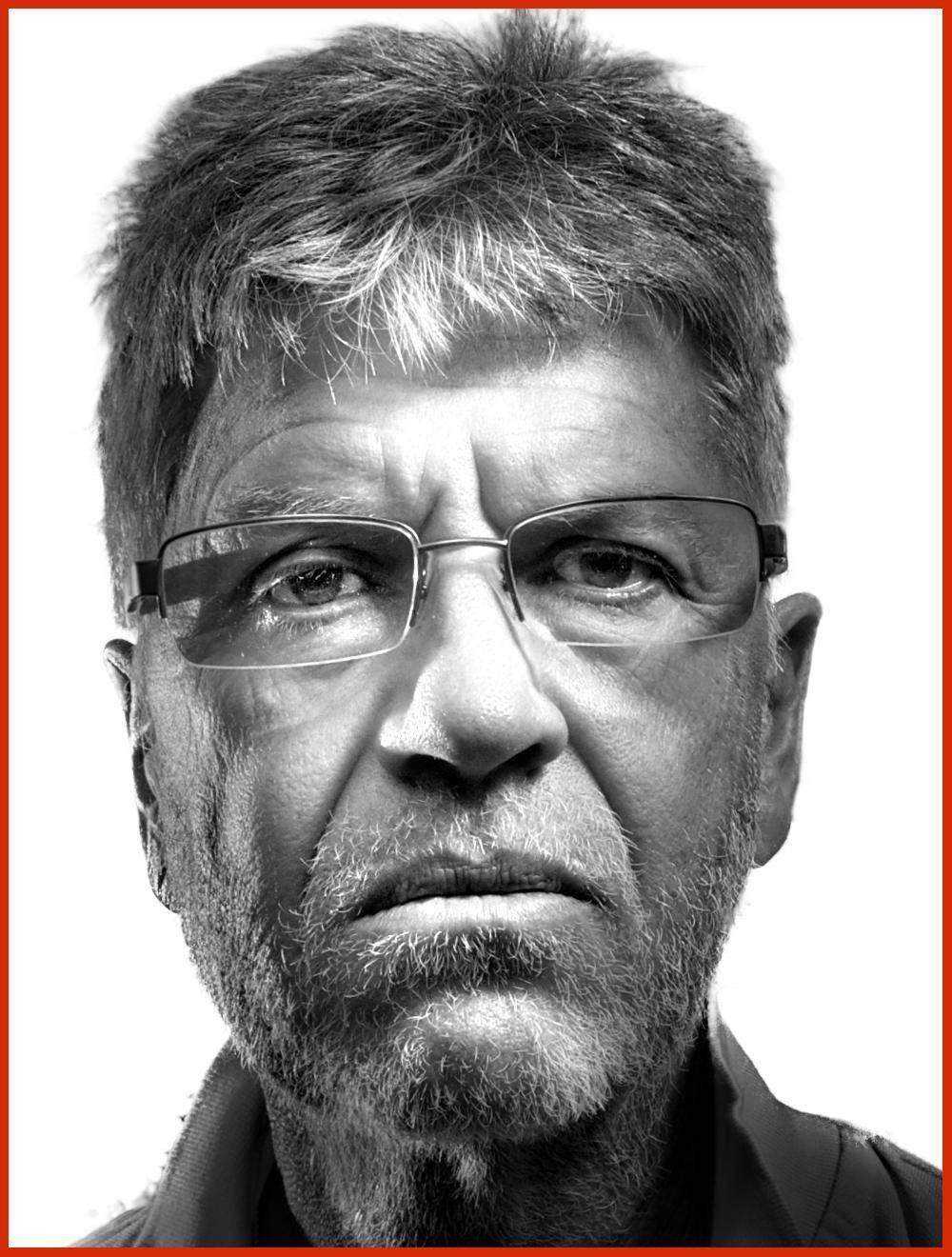}&
    \includegraphics[width=0.19\linewidth]{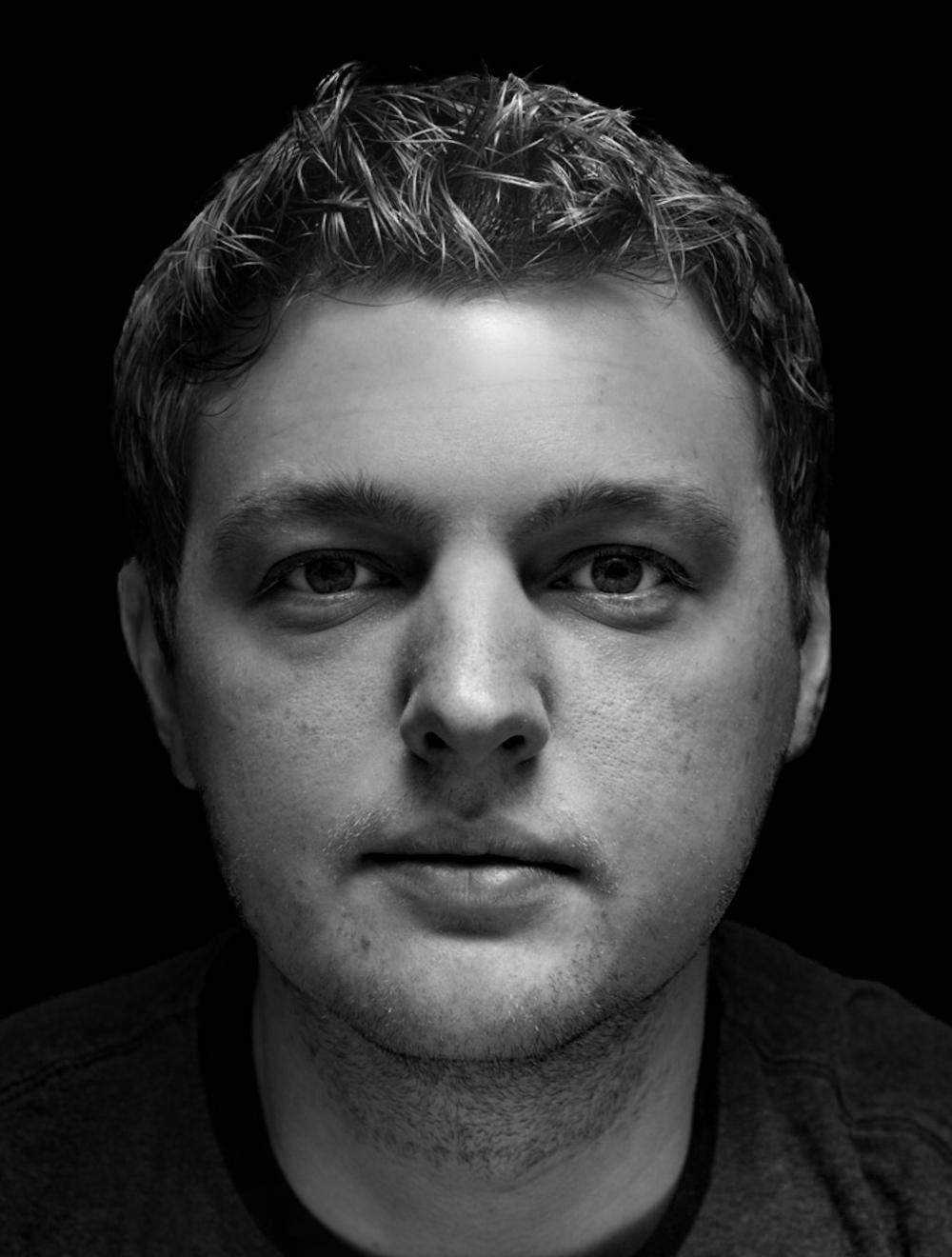}\\
    (a) Martin&(b) Platon&(c) Martin&(d) Platon&(e) Kelco\\
    \end{tabular}
\end{minipage}\\
\end{tabular}
\end{center}
\caption{Qualitative evaluation of human subject evaluation.
Input photos are in the first row. The results generated by local and proposed method are in the
second and third row, respectively. The transferred results in different styles are shown from (a) to (f). Photos marked by red rectangles indicate the preferred result by subjects.}
\label{fig:vote1}
\end{figure*}

The human subject evaluation
on the stylized face images
is carried out under three datasets.
As shown in Section \ref{sec:obj}, the holistic method is not effective
for transferring local contrast and thus the evaluation focuses on
the comparison between the proposed and local approaches.

There are 65 participants in the experiments
(45 are graduate students or faculty members).
For each participant, we randomly select 60 photos and split them
into three subsets.
%
%
We assign three styles to the three subsets randomly and generate
transferred results using two evaluated methods.
For visual comparison the input photo is positioned in the middle and
two results are shown on each side randomly on a high resolution
display.
we show some photo samples in each style
to one subject before experiments.
For the participants affiliated with the university,
the subjects are asked to select the result with the
fewest artifacts (i.e., in order to choose
the image in which local contrast is transferred most effectively).
The other participants are asked to subjectively select the result
in which the style represented by local contrast is well transferred.
We use different criteria as most participants affiliated with the university
have research background and are experienced to pick up
minor artifacts of the transferred images.
Meanwhile, the other participants tend to select images based on personal preference.
We tally the votes and show the voting results
of each method in Figure \ref{fig:vote45} and \ref{fig:vote20} respectively.
The evaluation results indicate that the performance
is similar between two groups of participants.
In other words, the quality of a \re{stylized face image} is mainly affected by artifacts.
%
%
Overall, human subjects consider that
the proposed method performs favorably against the local method on the three styles.

Figure \ref{fig:vote1} shows some stylized images in this evaluation.
The input images are on the first row.
The results by the local and proposed algorithms are on the second and third rows, respectively.
The photos marked by red rectangles indicate the
preferred results by subjects.
The stylized image generated by the local method shown in (a) contains
inconsistent local contrast around the hair and ear region.
In (b) the result generated by the local method lacks contrast in the hair region.
In addition, this stylized image contains artifacts in the forehead region.
%
In contrast, the proposed algorithm is able to effectively transfer the local contrast
without generating the artifacts.
In (c)-(e) both methods are able to effectively transfer local contrast without
introducing artifacts.
The user preference for these two images is somewhat random and
two methods receive almost the same number of votes.
As in practice different subjects appear in the exemplar and input photo,
it is challenging to find similar facial components from only one
exemplar.
%
%
The proposed algorithm alleviates this problem by using a collection
of exemplars, and performs favorably against the local method on
average across three styles as shown in Figures \ref{fig:vote45}, \ref{fig:vote20}, and \ref{fig:vote1}.
%

\section{Concluding Remarks}

In this work, we propose a face stylization algorithm using multiple exemplars.
As single exemplar-based methods are less effective to find similar facial
components for effective style transfer, we propose an algorithm using a collection of
exemplars and perform local patch identification via a Markov Random Field model.
\re{The facial components of an input photo can be properly selected from multiple exemplars through the MRF regularization}.
\re{It enables effective local energy transfer in the Laplacian stacks to construct the stylized output}.
\re{However, the stylized image is likely to contain artifacts due to inconsistency among multiple exemplars}.
\re{An effective artifact removal method based on an edge-preserving filter is used to refine the stylized output without losing local details}.
Experiments on the benchmark datasets containing three styles
demonstrate the effectiveness of the proposed algorithm against the state-of-the-art methods
in terms of qualitative and quantitative evaluations.





\section*{References}
{
\bibliographystyle{elsarticle-num}
\bibliography{ref}
}

\end{document}